\documentclass[10pt,twocolumn,letterpaper]{article}

\usepackage{cvpr}              % To produce the CAMERA-READY version
\usepackage{graphicx}
\usepackage{amsmath}
\usepackage{amssymb}
\usepackage{booktabs}
\usepackage[dvipsnames]{xcolor}
\usepackage{makecell}
\usepackage{amssymb}% http://ctan.org/pkg/amssymb
\usepackage{pifont}% http://ctan.org/pkg/pifont
\usepackage{diagbox}
\usepackage{multirow}
\usepackage{amsmath}
\usepackage{bbm}
\usepackage{fancyhdr}
\usepackage[accsupp]{axessibility}  % Improves PDF readability for those with disabilities.

\definecolor{citecolor}{HTML}{0071bc}
\definecolor{abbrcolor}{HTML}{990000}
\usepackage[breaklinks,colorlinks,citecolor=citecolor,urlcolor=citecolor,bookmarks=false]{hyperref}

\newcommand{\ourmodel}{ZOOM\xspace}

\DeclareMathOperator*{\argmin}{argmin}

\usepackage[capitalize]{cleveref}
\crefname{section}{Sec.}{Secs.}
\Crefname{section}{Section}{Sections}
\Crefname{table}{Table}{Tables}
\crefname{table}{Tab.}{Tabs.}

\def\cvprPaperID{3083} % *** Enter the CVPR Paper ID here
\def\confName{CVPR}
\def\confYear{2023}

\begin{document}

\title{Zero-shot Model Diagnosis}

\author{Jinqi Luo\footnotemark[1] \qquad Zhaoning Wang\footnotemark[1] \qquad Chen Henry Wu \qquad Dong Huang \qquad Fernando De la Torre\\
Carnegie Mellon University\\
{\tt\small \{jinqil, zhaoning, chenwu2, dghuang, ftorre\}@cs.cmu.edu}
}

\maketitle

{
  \renewcommand{\thefootnote}%
    {\fnsymbol{footnote}}
  \footnotetext[1]{Equal contribution.}
  \footnotetext[2]{The code is publicly available at the project page: \href{https://zero-shot-model-diagnosis.github.io/}{https://zero-shot-model-diagnosis.github.io/}. }
}

%%%%%%%%% ABSTRACT
\begin{abstract}
\vspace{-3mm}
When it comes to deploying deep vision models, the behavior of these systems must be explicable to ensure confidence in their reliability and fairness. A common approach to evaluate deep learning models is to build a labeled test set with attributes of interest and assess how well it performs. However, creating a balanced test set (i.e., one that is uniformly sampled over all the important traits) is often time-consuming, expensive, and prone to mistakes. The question we try to address is: can we evaluate the sensitivity of deep learning models to arbitrary visual attributes \textbf{without an annotated test set}? 

This paper argues the case that \textcolor{abbrcolor}{\textbf{Z}}er\textcolor{abbrcolor}{\textbf{o}}-sh\textcolor{abbrcolor}{\textbf{o}}t \textcolor{abbrcolor}{\textbf{M}}odel Diagnosis (ZOOM) is possible without the need for a test set nor labeling. To avoid the need for test sets, 
our system relies on a generative model and CLIP. The key idea is enabling the user to select a set of prompts (relevant to the problem) and our system will automatically search for semantic counterfactual images (i.e., synthesized images that flip the prediction in the case of a binary classifier) using the generative model.  We evaluate several visual tasks (classification, key-point detection, and segmentation) in multiple visual domains to demonstrate the viability of our methodology. Extensive experiments demonstrate that our method is capable of producing counterfactual images and offering sensitivity analysis for model diagnosis without the need for a test set. 
\end{abstract}

%%%%%%%%% BODY TEXT
% \jinqi{I think we have two points that need discussion across the whole paper: \\
% 1. we are using a bit much (e.g., ...), (i.e., ...). Maybe re-paraphrase them is a good choice.\\
% 2. we sometimes repeatedly saying the same thing in Figure, caption of Figure, and the description text. For example, Fig.~\ref{sec:single_attr_diagnosis} "multi-attr is powerful". The redundancy reduces the information density.\\
% 3. Sec 4.3 is about single-attribute validation, need minor change about description.\\
% 4. The Title of our work and the abbreviation needs thorough discussion.\\
% }

\begin{figure}[!t]
\centering
  \includegraphics[width=\linewidth]{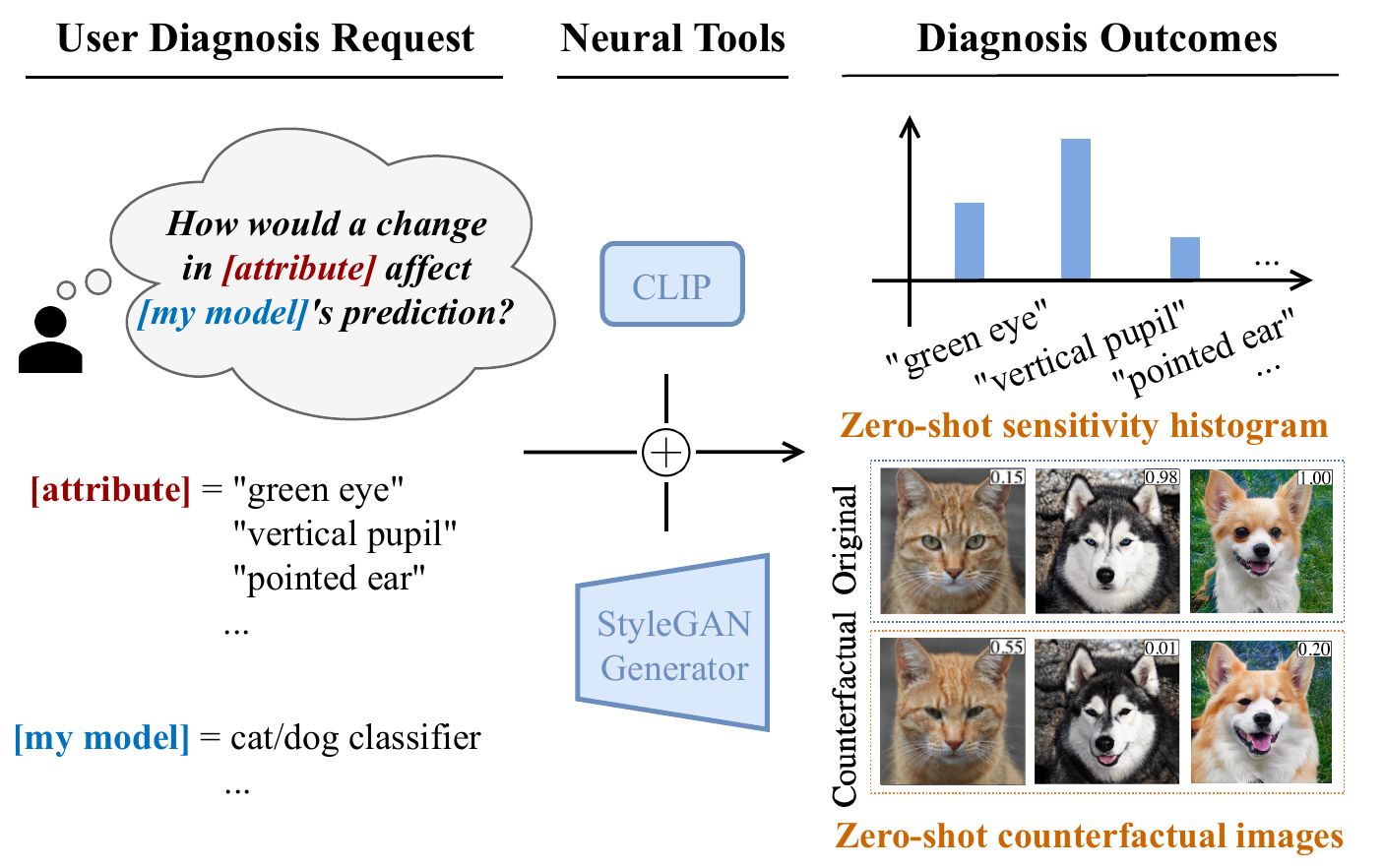}
\caption{Given a differentiable deep learning model (e.g., a cat/dog classifier) and user-defined text attributes, how can we determine the model's sensitivity to specific attributes without using labeled test data? 
Our system generates counterfactual images (bottom right) based on the textual directions provided by the user, while also computing the sensitivity histogram (top right).
%Our system creates counterfactual images (see right bottom) in directions specified in texts by the user, as well as computing the sensitivity histogram (see right top).
}
\label{fig:first_fig}
\vspace{-4mm}
\end{figure}

\vspace{-.15in}

\section{Introduction}
\label{sec:intro}

Deep learning models inherit data biases, which can be accentuated or downplayed depending on the model's architecture and optimization strategy. 
Deploying a computer vision deep learning model requires extensive testing and evaluation, with a particular focus on features with potentially dire social consequences (e.g., non-uniform behavior across gender or ethnicity). Given the importance of the problem, it is common to collect and label large-scale datasets to evaluate the behavior of these models across attributes of interest. Unfortunately, collecting these test datasets is extremely time-consuming, error-prone, and expensive. 
Moreover, a balanced dataset, that is uniformly distributed across all attributes of interest, is also typically impractical to acquire due to its combinatorial nature. Even with careful metric analysis in this test set, no robustness nor fairness can be guaranteed since there can be a mismatch between the real and test distributions~\cite{Ramaswamy_2021_CVPR}. This research will explore model diagnosis without relying on a test set in an effort to {\em democratize} model diagnosis and lower the associated cost. 

Counterfactual explainability as a means of model diagnosis is drawing the community's attention \cite{Mothilal_2020, pmlr-v97-goyal19a}.
Counterfactual images visualize the sensitive factors of an input image that can influence the model's outputs. In other words, counterfactuals answer the question: \textit{``How can we modify the input image $\mathbf{x}$ (while fixing the ground truth) so that the model prediction would diverge from $\mathbf{y}$ to $\hat{\mathbf{y}}$?''}. The parameterization of such counterfactuals will provide insights into identifying key factors of where the model fails. Unlike existing image-space adversary techniques \cite{goodfellow2014fgsm, madry2018towards}, counterfactuals provide semantic perturbations that are interpretable by humans. However, existing counterfactual studies require the user to either collect uniform test sets~\cite{karkkainenfairface}, annotate discovered bias~\cite{li-2021-discover}, or train a model-specific explanation every time the user wants to diagnose a new model \cite{Lang_2021_ICCV}.
%How to design a plug-and-play model diagnosis toolkit that can democratize to greater community has become a challenging question.

On the other hand, recent advances in Contrastive Language-Image Pretraining (CLIP)~\cite{CLIP} can help to overcome the above challenges. CLIP enables text-driven applications that map user text representations to visual manifolds for downstream tasks such as avatar generation \cite{hong2022avatarclip}, motion generation \cite{Zhang2022MotionDiffuseTH} or neural rendering \cite{Poole2022DreamFusionTU,2022CLIPNeRF}. In the domain of image synthesis, StyleCLIP~\cite{2021StyleCLIP} reveals that text-conditioned optimization in the StyleGAN \cite{2019stylegan} latent space can decompose latent directions for image editing, allowing for the mutation of a specific attribute without disturbing others. 
With such capability, users can freely edit semantic attributes conditioned on text inputs. This paper further explores its use in the scope of model diagnosis. 
% Freedom of users to edit semantic attributes conditioned on text inputs inspires us to further explore its potential in the scope of model diagnosis. 

The central concept of the paper is depicted in Fig.~\ref{fig:first_fig}. Consider a user interested in evaluating which factors contribute to the lack of robustness in a cat/dog classifier (target model).  By selecting a list of keyword attributes, the user is able to (1) see counterfactual images where semantic variations flip the target model predictions (see the classifier score in the top-right corner of the counterfactual images) and (2) quantify the sensitivity of each attribute for the target model (see sensitivity histogram on the top).  Instead of using a test set, we propose using a StyleGAN generator as the picture engine for sampling counterfactual images. CLIP transforms user's text input, and enables model diagnosis in an open-vocabulary setting. This is a major advantage since there is no need for collecting and annotating images and minimal user expert knowledge. In addition, we are not tied to a particular annotation from datasets (e.g., specific attributes in CelebA~\cite{liu2015celeba}). 

To summarize, our proposed work offers three major improvements over earlier efforts:
\vspace{-1mm}
\begin{itemize}

 \item The user requires neither a labeled, balanced test dataset, and minimal expert knowledge in order to evaluate where a model fails (i.e., model diagnosis). In addition, the method provides a sensitivity histogram across the attributes of interest. 

 \item When a different target model or a new user-defined attribute space is introduced, it is not necessary to re-train our system, allowing for practical use. 

 \item The target model fine-tuned with counterfactual images not only slightly improves the classification performance, but also greatly increases the distributional robustness against counterfactual images. 

 \end{itemize}
 
\begin{figure*}[!t]
    \centering
    \vspace{-2mm}
    \includegraphics[width=0.95\linewidth]{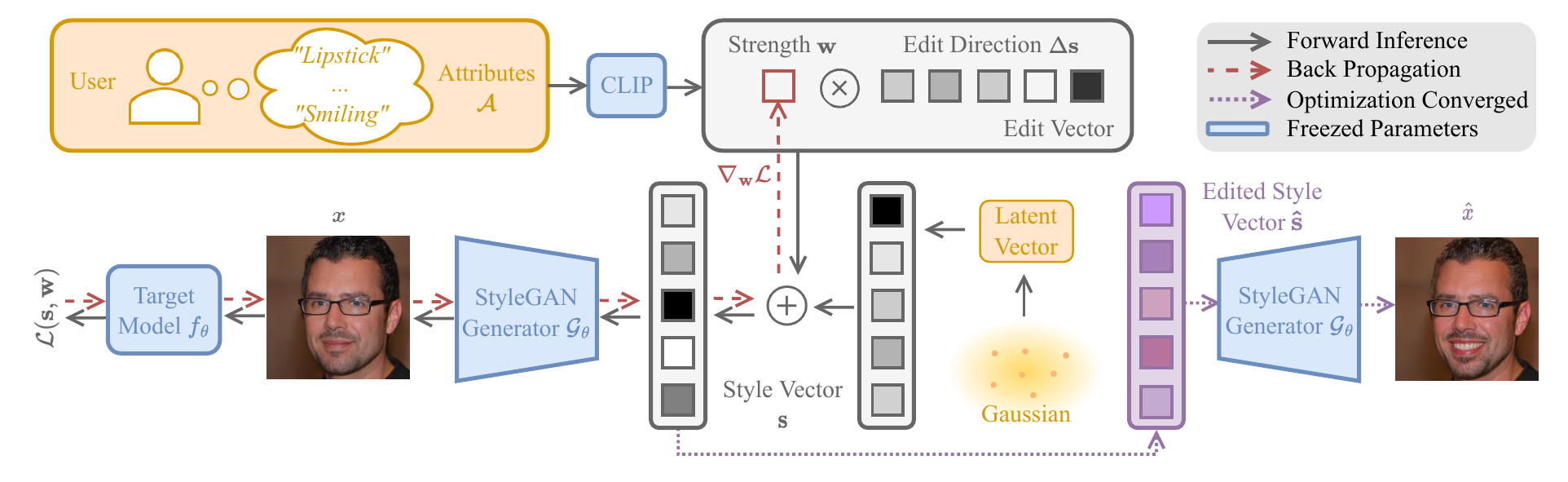}
    \vspace{-1mm}
    \caption{The \ourmodel framework. Black solid lines stand for forward passes, red dashed lines stand for backpropagation, and purple dashed lines stand for inference after the optimization converges. The user inputs single or multiple attributes, and we map them into edit directions with the method in Sec.~\ref{sec:global_direction_extraction}. Then we assign to each edit direction (attribute) a weight, which represents how much we are adding/removing this attribute. We iteratively perform adversarial learning on the attribute space to maximize the counterfactual effectiveness.
    }
    \label{fig:model}
    \vspace{-2mm}
\end{figure*}
\vspace{-1mm}
\section{Related Work}
\label{sec:related_work}

This section reviews prior work on attribute editing with generative models and recent efforts on model diagnosis. 

\subsection{Attribute Editing with Generative Models} 
\vspace{-1mm}
With recent progress in generative models, GANs supports high-quality image synthesis, as well as semantic attributes editing \cite{Xia2022GANIA}. \cite{attGAN, choi2020starganv2} edit the images by perturbing the intermediate latent space encoded from the original images. These methods rely on images to be encoded to latent vectors to perform attribute editing. On the contrary, StyleGAN~\cite{2019stylegan} can produce images by sampling the latent space. Many works have explored ways to edit attributes in the latent space of StyleGAN, either by relying on image annotations~\cite{interfacegan} or in an unsupervised manner~\cite{sefa, ganspace}. StyleSpace~\cite{stylespace} further disentangles the latent space of StyleGAN and can perform specific attribute edits by disentangled style vectors. Based upon StyleSpace, StyleCLIP~\cite{2021StyleCLIP} builds the connection between the CLIP language space and StyleGAN latent space to enable arbitrary edits specified by the text. Our work adopts this concept for fine-grained attribute editing.

\subsection{Model Diagnosis}

To the best of our knowledge, model diagnosis without a test set is a relatively unexplored problem. In the adversarial learning literature, it is common to find methods that show how image-space perturbations \cite{goodfellow2014fgsm,madry2018towards} flip the model prediction; however, such perturbations lack visual interpretability. \cite{2018GANAE} pioneers in synthesizing adversaries by GANs. More recently, \cite{2019SemanticAE,CounterfactualGAN,2020semanticadv} propose generative methods to synthesize semantically perturbed images to visualize where the target model fails. However, their attribute editing is limited within the dataset's annotated labels. Instead, our framework allows users to easily customize their own attribute space, in which we visualize and quantify the biased factors that affect the model prediction. On the bias detection track, \cite{Lang_2021_ICCV} co-trains a model-specific StyleGAN with each target model, and requires human annotators to name attribute coordinates in the Stylespace. \cite{li-2021-discover,denton2019image,attractive} synthesize counterfactual images by either optimally traversing the latent space or learning an attribute hyperplane, after which the user will inspect the represented bias. Unlike previous work, we propose to diagnose a deep learning model without any model-specific re-training, new test sets, or manual annotations/inspections.
 
\section{Method}
\label{sec:our_method}
\vspace{-1mm}
This section firstly describes our method to generate counterfactual images guided by CLIP in a zero-shot manner. We then introduce how we perform the sensitivity analysis across attributes of interest. Fig.~\ref{fig:model} shows the overview of our framework.

\subsection{Notation and Problem Definition}

Let $f_\theta$, parameterized by $\theta$, be the target model that we want to diagnose. In this paper, $f_\theta$ denotes two types of deep nets: binary attribute classifiers and face keypoint detectors. Note that our approach is extendable to any end-to-end differentiable target deep models. Let $\mathcal{G}_\phi$, parameterized by $\phi$, be the style generator that synthesizes images by $\mathbf{x} = \mathcal{G}_\phi(\mathbf{s})$ where $\mathbf{s}$ is the style vector in Style Space $\mathcal{S}$ \cite{stylespace}.
%Our goal is to search and visualize the $\hat{u}$ following the user's flexible prompt, which can inspire future application of model debiasing. 
We denote a counterfactual image as $\mathbf{\hat{x}}$, which is a synthesized image that misleads the target model $f_\theta$, and denote the 
original reference image as $\mathbf{x}$. $a$ is defined as a single user input text-based attribute, with its domain $\mathcal{A}  = \{a_i\}_{i=1}^N$ for $N$ input attributes.
$\mathbf{\hat{x}}$ and $\mathbf{x}$ differs only along attribute directions $\mathcal{A}$. Given a set of $\{f_\theta, \mathcal{G}_\phi, \mathcal{A}\}$, our goal is to perform counterfactual-based diagnosis to interpret where the model fails without manually collecting nor labeling any test set. 
% The counterfactual images will be created with adversarial learning; however, 
Unlike traditional approaches of image-space noises which lack explainability to users, our method adversarially searches the counterfactual in the user-designed semantic space. To this end, our diagnosis will have three outputs, namely counterfactual images (Sec.~\ref{sec:Counterfactual_Synthesis}), sensitivity histograms (Sec.~\ref{sec:Attribute_Sensitivity_Analysis}), and distributionally robust models (Sec.~\ref{sec:ct}).

\subsection{Extracting Edit Directions}\label{sec:global_direction_extraction}
\vspace{-1mm}
%In order to make the paper self-contained, 
This section examines the terminologies, method, and modification we adopt in \ourmodel to extract suitable global directions for attribute editing. Since CLIP has shown strong capability in disentangling visual representation \cite{2022Disentangling}, we incorporate 
style channel relevance from StyleCLIP~\cite{2021StyleCLIP} to find edit directions for each attribute. 

 Given the user's input strings of attributes, we want to find an image manipulation direction $\Delta \mathbf{\mathbf{s}}$ for any $\mathbf{s} \sim \mathcal{S}$, such that the generated image $\mathcal{G}_\phi(\mathbf{s} + \Delta{ \mathbf{s}) }$ {\em only} varies in the input attributes. Recall that CLIP maps strings into a text embedding $\mathbf{t} \in \mathcal{T}$, the text embedding space. For a string attribute description $a$ and a neutral prefix $p$, we obtain the CLIP text embedding difference $\Delta\mathbf{t}$ by:
\begin{align}
        \Delta\mathbf{t} = \operatorname{CLIP_{text}}(p \oplus a) - \operatorname{CLIP_{text}}(p)
                % \Delta\mathbf{t} = \{&\operatorname{CLIP}_{text}(p \oplus a_i) -\operatorname{CLIP}_{text}(p)\}_{i=1}^{N}
\end{align}
where $\oplus$ is the string concatenation operator. To take `Eyeglasses' as an example, we can get $\Delta\mathbf{t}= \operatorname{CLIP_{text}}(\textnormal{`a face with Eyeglasses'}) -\operatorname{CLIP_{text}}(\textnormal{`a face'}) $. 

To get the edit direction, $\Delta \mathbf{s}$, we need to utilize a style relevance mapper $\mathbf{M} \in \mathbbm{R}^{c_\mathcal{S} \times c_\mathcal{T}}$ to map between the CLIP text embedding vectors of length $c_\mathcal{T}$
and the Style space vector of length $c_\mathcal{S}$. 
StyleCLIP optimizes $\mathbf{M}$ by iteratively searching meaningful style channels: mutating each channel in $\mathcal{S}$ and encoding the mutated images by CLIP to assess whether there is a significant change in $\mathcal{T}$ space. To prevent undesired edits that are irrelevant to the user prompt, the edit direction $\Delta \mathbf{s}$ will filter out channels that the style value change is insignificant:
\begin{align}
    \label{eq:CRM}
    \Delta \mathbf{s} &= (\mathbf{M} \cdot \Delta\mathbf{t}) \odot \mathbbm{1}((\mathbf{M} \cdot \Delta\mathbf{t}) > \lambda),
\end{align}
 where $\lambda$ is the hyper-parameter for the threshold value. $\mathbbm{1}(\cdot)$ is the indicator function, and $\odot$ is the element-wise product operator. Since the success of attribute editing by the extracted edit directions will be the key to our approach, Appendix A will show the capability of CLIP by visualizing the global edit direction on multiple sampled images, conducting the user study, and analyzing the effect of $\lambda$.

\subsection{Style Counterfactual Synthesis}
\label{sec:Counterfactual_Synthesis}

Identifying semantic counterfactuals necessitates a manageable parametrization of the semantic space for effective exploration. For ease of notation, we denote $(\Delta \mathbf{s})_{i}$ as the global edit direction for $i^{th}$ attribute $a_i \in \mathcal{A}$ from the user input. After these $N$ attributes are provided and the edit directions are calculated, we initialize the control vectors $\mathbf{w}$ of length $N$ where the $i^{th}$ element $w_i$ controls the strength of the $i^{th}$ edit direction. Our counterfactual edit will be a linear combination of normalized edit directions: $\mathbf{s}_{edit} = \sum_{i=1}^N w_i \frac{(\Delta \mathbf{s})_{i}}{||(\Delta \mathbf{s})_{i}||}$.

The black arrows in Fig.~\ref{fig:model} show the forward inference to synthesize counterfactual images. Given the parametrization of attribute editing strengths and the final loss value, our framework searches for counterfactual examples in the optimizable edit weight space. The original sampled image is $\mathbf{x} = G_{\phi}(\mathbf{s})$, and the counterfactual image is
\begin{align}
\label{eq:total_edit}
    \mathbf{\hat{x}} = G_{\phi}(\mathbf{s} + \mathbf{s}_{edit}) = G_{\phi}\left(\mathbf{s} + \sum_{i=1}^N w_i \frac{(\Delta \mathbf{s})_{i}}{||(\Delta \mathbf{s})_{i}||}\right),
\end{align}
which is obtained by minimizing the following loss, $\mathcal{L}$, that is the weighted sum of three terms:
\begin{align}
    \mathcal{L} (\mathbf{s}, \mathbf{w}) &= \alpha \mathcal{L}_{target}(\mathbf{\hat{x}}) + \beta \mathcal{L}_{struct}(\mathbf{\hat{x}}) + \gamma \mathcal{L}_{attr}(\mathbf{\hat{x}}).
\end{align}
 We back-propagate to optimize $\mathcal{L}$ w.r.t the weights of the edit directions $\mathbf{w}$, shown as the red pipeline in Fig.~\ref{fig:model}.

 The targeted adversarial loss $\mathcal{L}_{target}$ for binary attribute classifiers minimizes the distance between the current model prediction $f_{\theta}(\mathbf{\hat{x}})$ with the flip of original prediction $\hat{p}_{cls} = 1 - f_{\theta}(\mathbf{x})$. In the case of an eyeglass classifier on a person wearing eyeglasses, $\mathcal{L}_{target}$ will guide the optimization to search $\mathbf{w}$ such that the model predicts no eyeglasses. For a keypoint detector, the adversarial loss will minimize the distance between the model keypoint prediction with a set of \textit{random} points $\hat{p}_{kp} \sim \mathcal{N}$:
\begin{align}
    &\text{(binary classifier) }\mathcal{L}_{target}(\mathbf{\hat{x}}) = {L}_{CE}(f_{\theta}(\mathbf{\hat{x}}), \hat{p}_{cls}),\\
    &\text{(keypoint detector) }\mathcal{L}_{target}(\mathbf{\hat{x}}) = {L}_{MSE}(f_{\theta}(\mathbf{\hat{x}}), \hat{p}_{kp}).
\end{align}

If we only optimize $\mathcal{L}_{target}$ w.r.t the global edit directions, it is possible that the method will not preserve image statistics of the original image and can include the particular attribute that we are diagnosing. To constrain the optimization, we added a structural loss $\mathcal{L}_{struct}$ and  an attribute consistency loss $\mathcal{L}_{attr}$ to avoid  generation collapse. $\mathcal{L}_{struct}$ \cite{SSIM} aims to preserve global image statistics of the original image $\mathbf{x}$ including image contrasts, background, or shape identity  during the adversarial editing. While $\mathcal{L}_{attr}$ enforces that the target attribute (perceived ground truth) be consistent on the style edits. For example, when diagnosing the eyeglasses classifier, \ourmodel preserves the original status of eyeglasses and precludes direct eyeglasses addition/removal. 
\begin{align}
    \mathcal{L}_{struct}(\mathbf{\hat{x}}) &= L_{SSIM}(\mathbf{\hat{x}}, \mathbf{x})\\
    \mathcal{L}_{attr}(\mathbf{\hat{x}}) &= {L}_{CE}(\operatorname{CLIP}(\mathbf{\hat{x}}), \operatorname{CLIP}(\mathbf{x})) % \|\mathbf{s}_{edit}\|_2^2
\end{align}

Given a pretrained target model $f_{\theta}$, a domain-specific style generator $G_{\phi}$, and a text-driven attribute space $\mathcal{A}$, our goal is to sample an original style vector $\mathbf{s}$ for each image and search its counterfactual edit strength $\mathbf{\hat{w}}$:
\begin{align}
    \mathbf{\hat{w}} &= \argmin_{\mathbf{w}} \mathcal{L} (\mathbf{s}, \mathbf{w}).% \argmin_{w}\mathcal{L}\left(f_{\theta},G_{\phi},\mathcal{A} \right)
\end{align}
\looseness=-1
Unless otherwise stated, we iteratively update $ \mathbf{w}$ as:  

\begin{align}
    \mathbf{w} &= \mbox{clamp}_{\mathrm{[-\epsilon, \epsilon]}}(\mathbf{w} - \eta \nabla_{\mathbf{w}} \mathcal{L}),
\end{align}
where $\eta$ is the step size and $\epsilon$ is the clamp bound to avoid synthesis collapse caused by exaggerated edit. Note that the maximum counterfactual effectiveness does not indicate the maximum edit strength (i.e., $w_i = \epsilon$), since the attribute edit direction does not necessarily overlap with the target classifier direction. The attribute change is bi-directional, as the $w_i$ can be negative in Eq.~\ref{eq:total_edit}. Details of using other optimization approaches (e.g., linear approximation \cite{madry2018towards}) will be discussed in Appendix C. 

\subsection{Attribute Sensitivity Analysis}
\label{sec:Attribute_Sensitivity_Analysis}

Single-attribute counterfactual reflects the sensitivity of target model on the individual attribute. By optimizing independently along the edit direction for a single attribute and averaging the model probability changes over images, our model generates independent sensitivity score $h_{i}$ for each attribute $a_{i}$:
\begin{align}
h_{i} = \mathbb{E}_{\mathbf{x}\sim\mathcal{P}(\mathbf{x}), \mathbf{\hat{x}} = \text{\ourmodel}(\mathbf{x}, a_{i}) }|f_{\theta}(\mathbf{x}) - f_{\theta}(\mathbf{\hat{x}})|.
\end{align} 
The sensitivity score $h_{i}$ is the probability difference between the original image  $ \mathbf{x} $ and generated image $\mathbf{\hat{x}}$, at the most counterfactual point when changing attribute $a_{i}$.

\begin{figure*}[h]
    \centering
    \begin{subfigure}[b]{\linewidth}
    \includegraphics[width=0.245\linewidth]{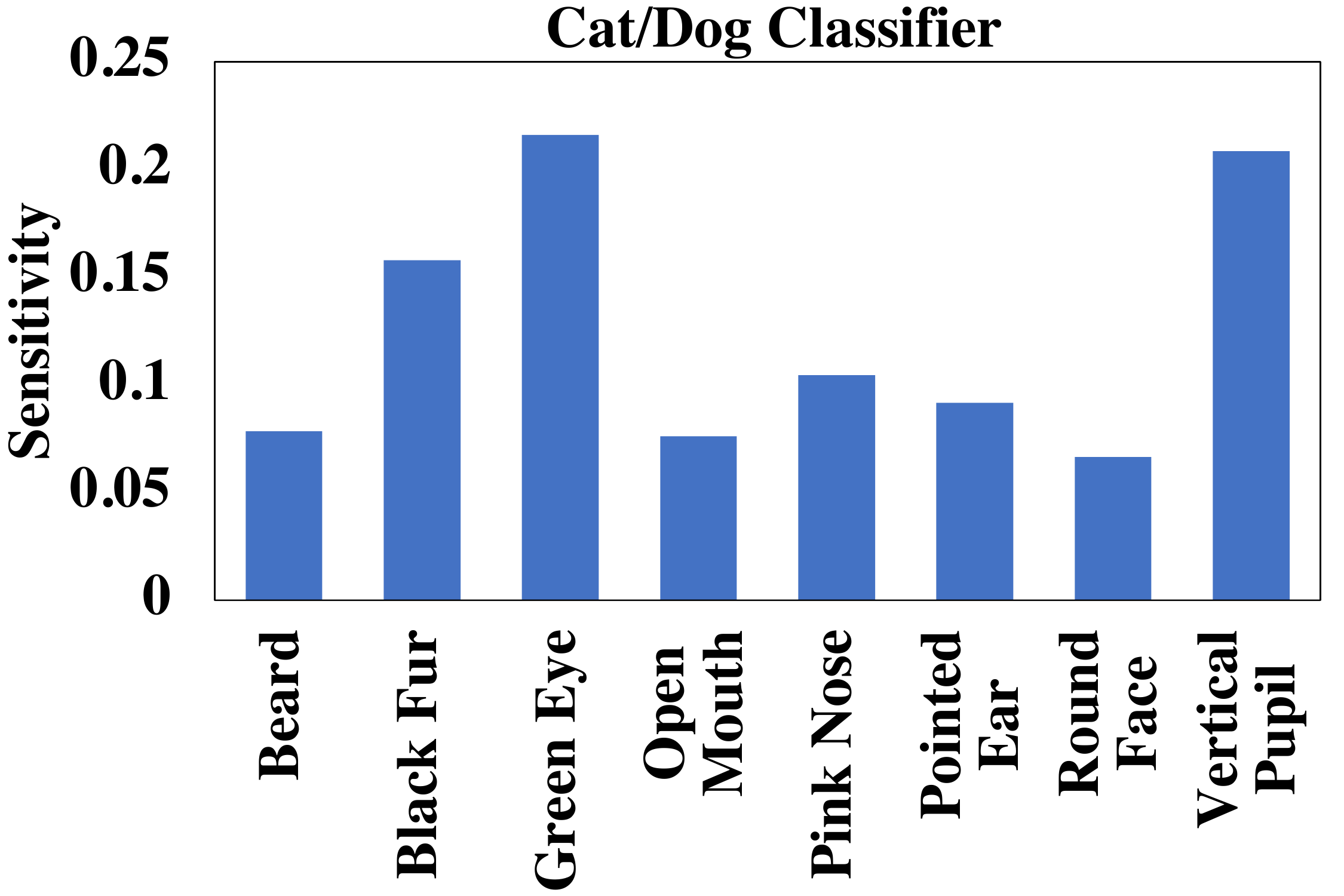}
    \includegraphics[width=0.245\linewidth]{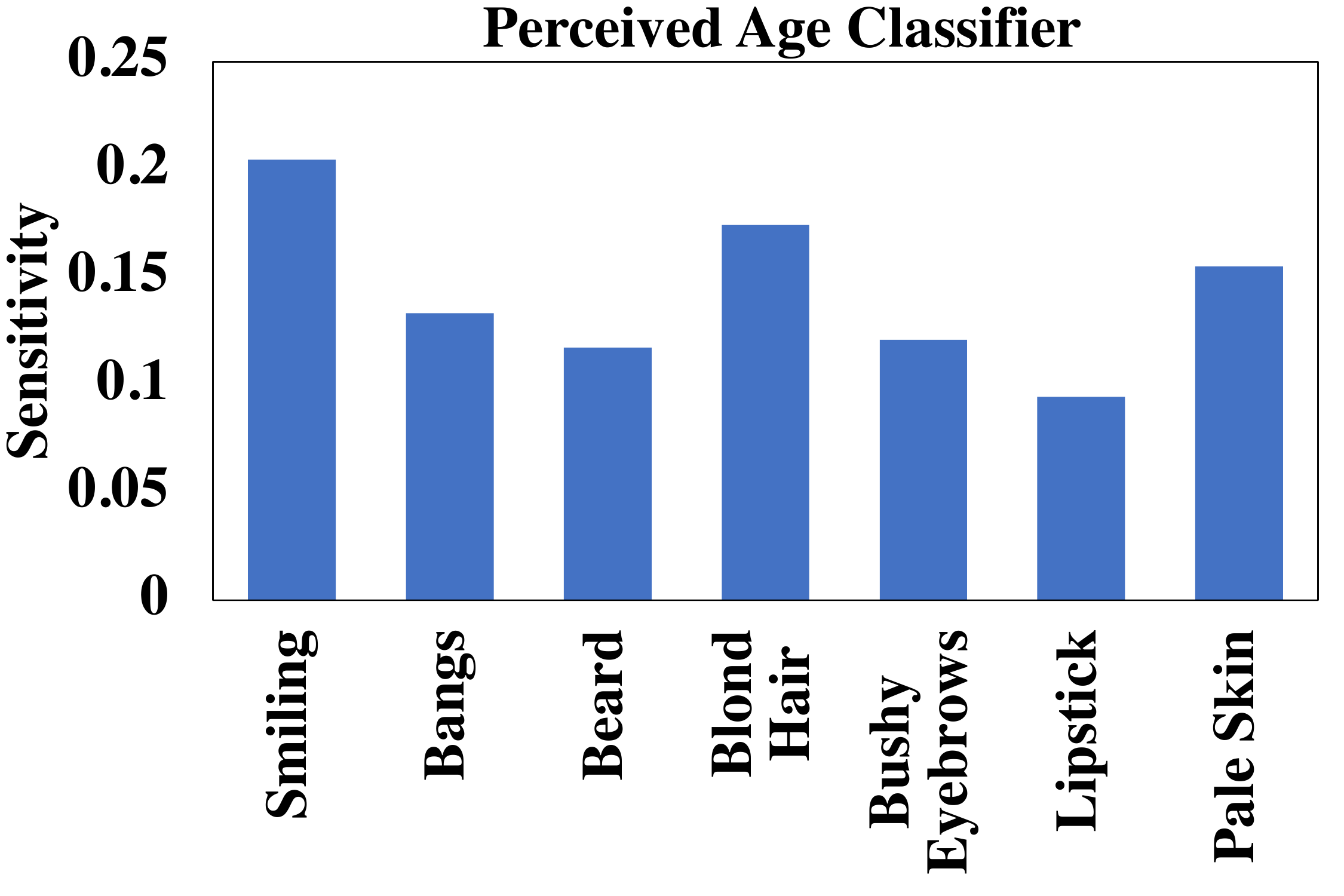}
    \includegraphics[width=0.245\linewidth]{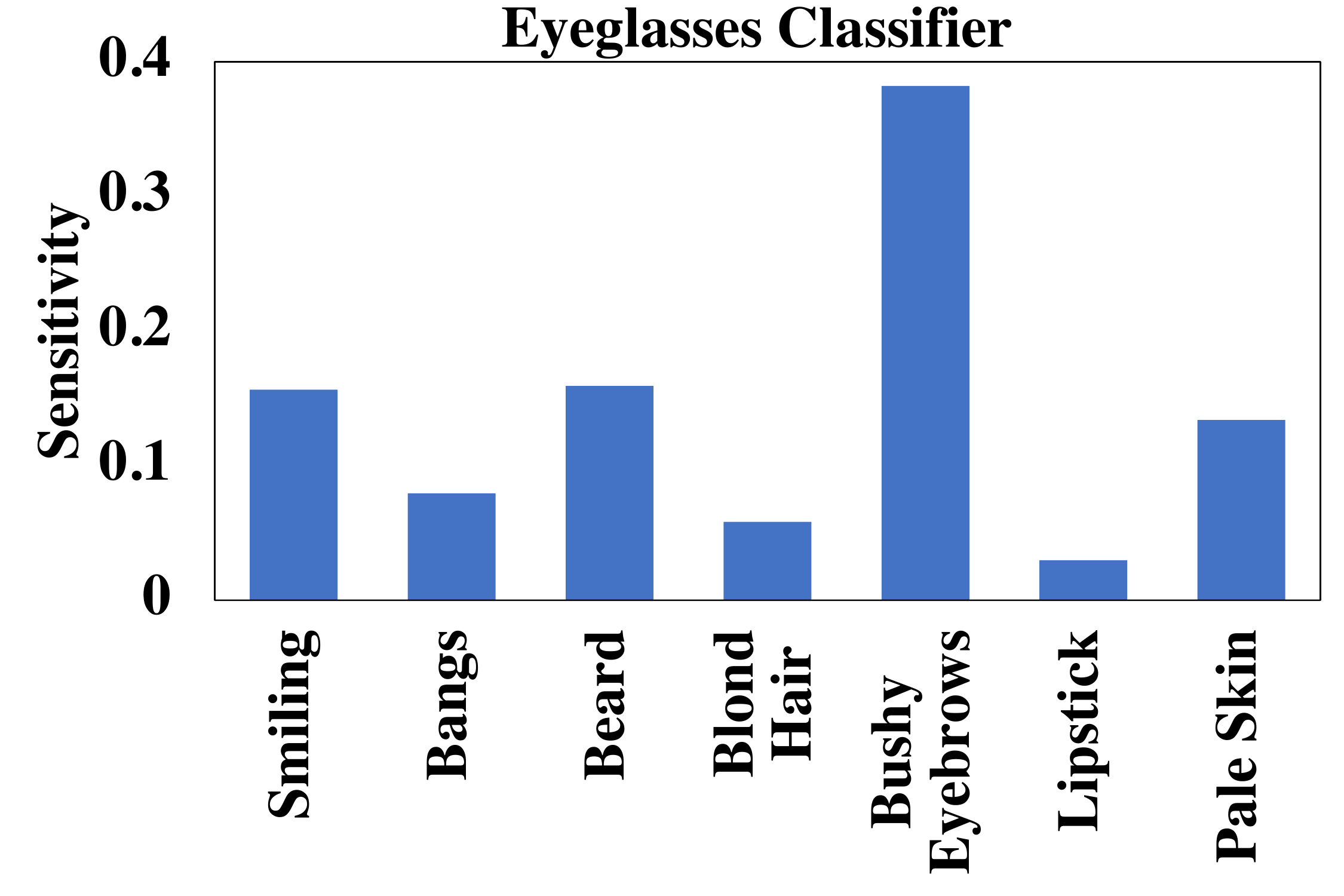}
    \includegraphics[width=0.245\linewidth]{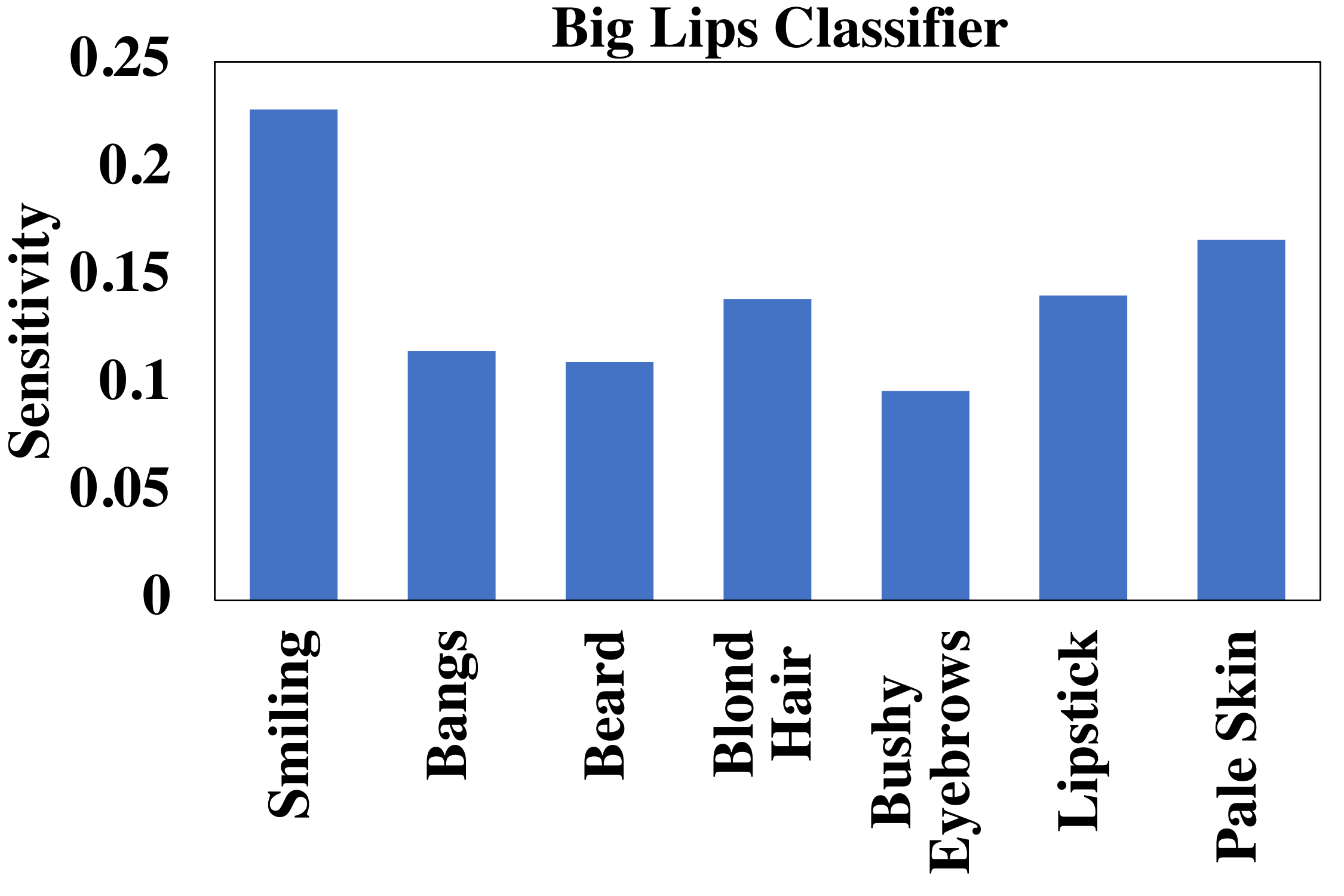}
    \caption{Model diagnosis histograms generated by \ourmodel on four facial attribute classifiers.}
    \label{fig:histograms_vanilla}
    \end{subfigure}
    \begin{subfigure}[b]{\linewidth}
    \includegraphics[width=0.245\linewidth]{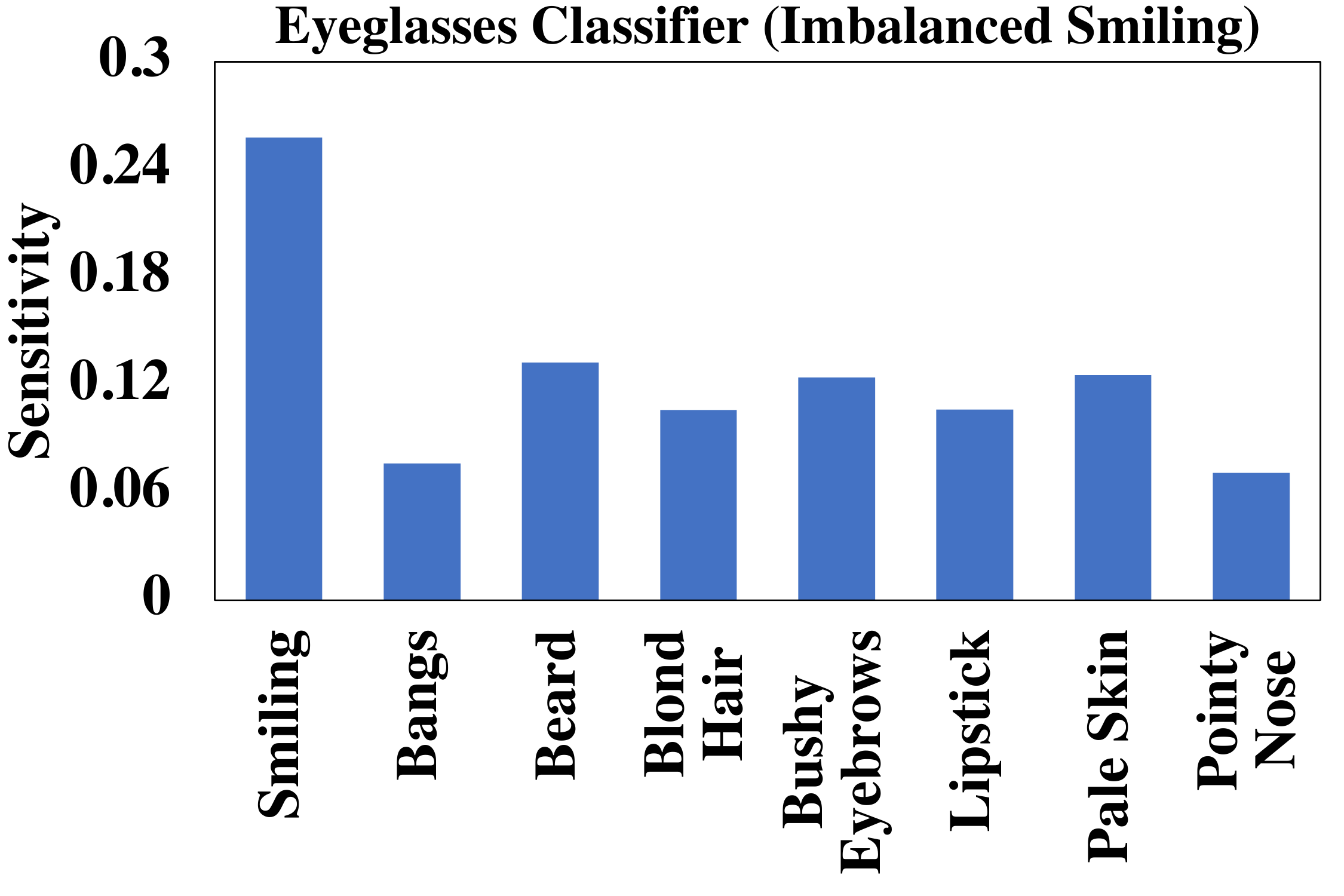}
    \includegraphics[width=0.245\linewidth]{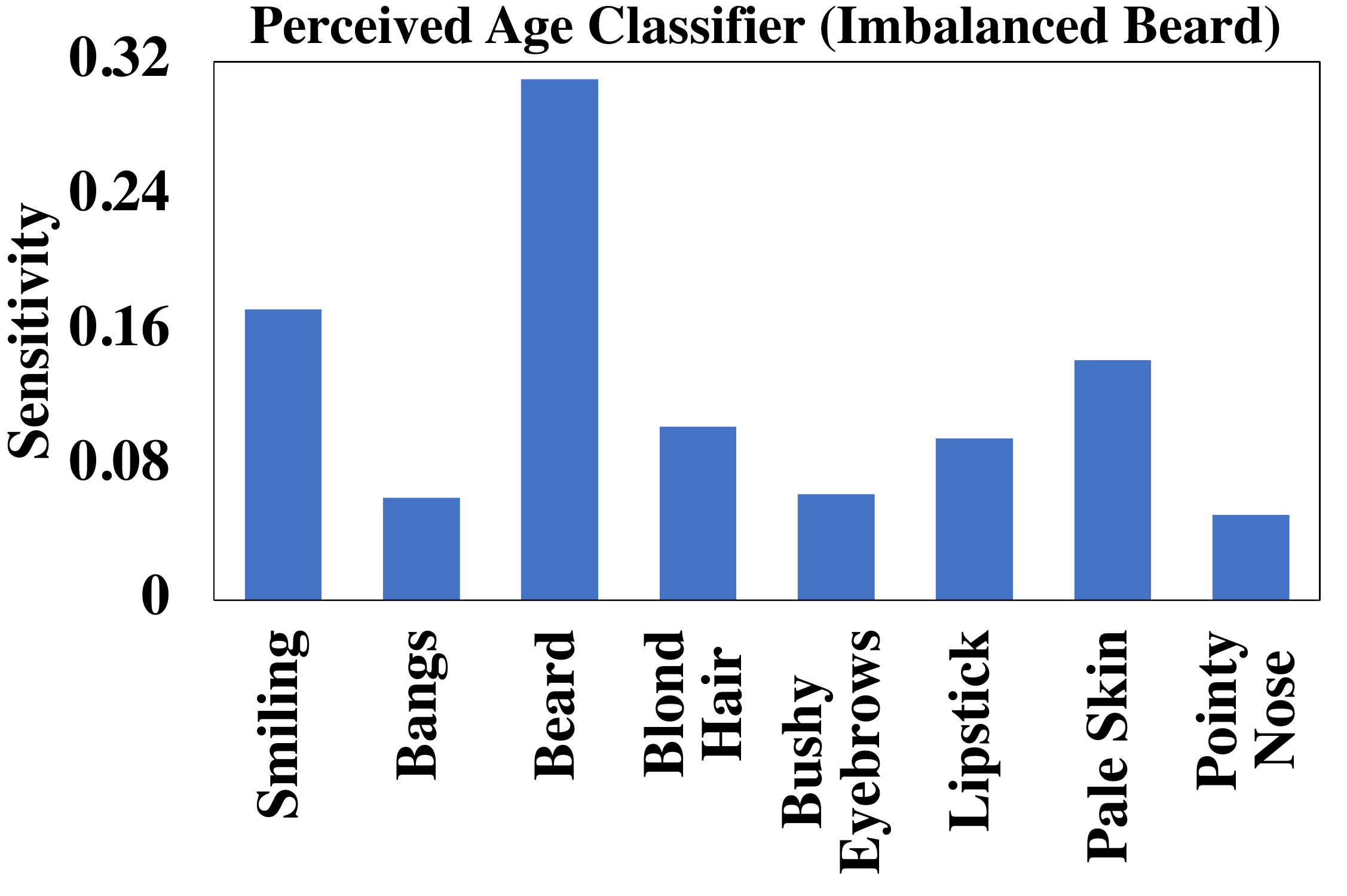}
    \includegraphics[width=0.245\linewidth]{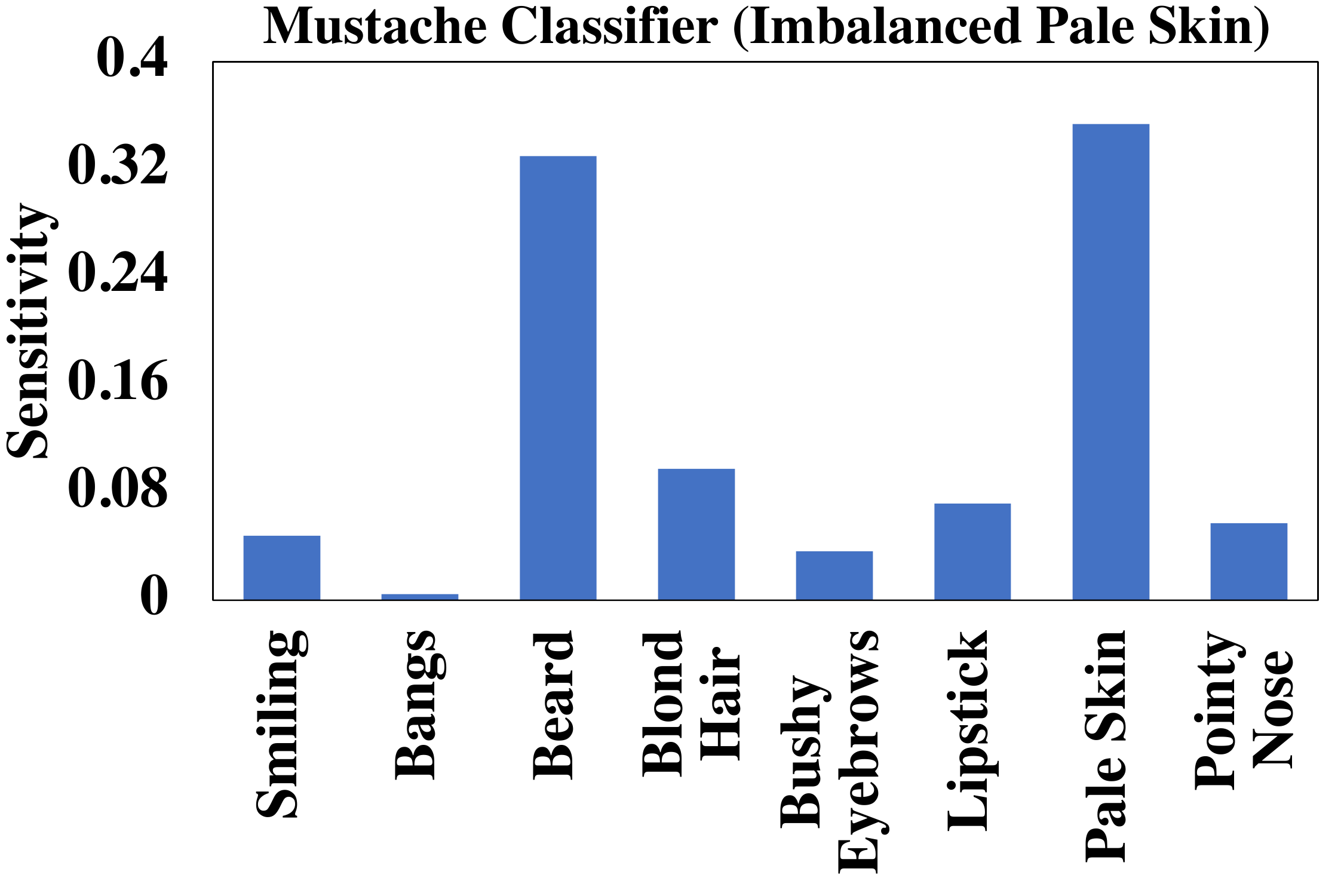}
    \includegraphics[width=0.245\linewidth]{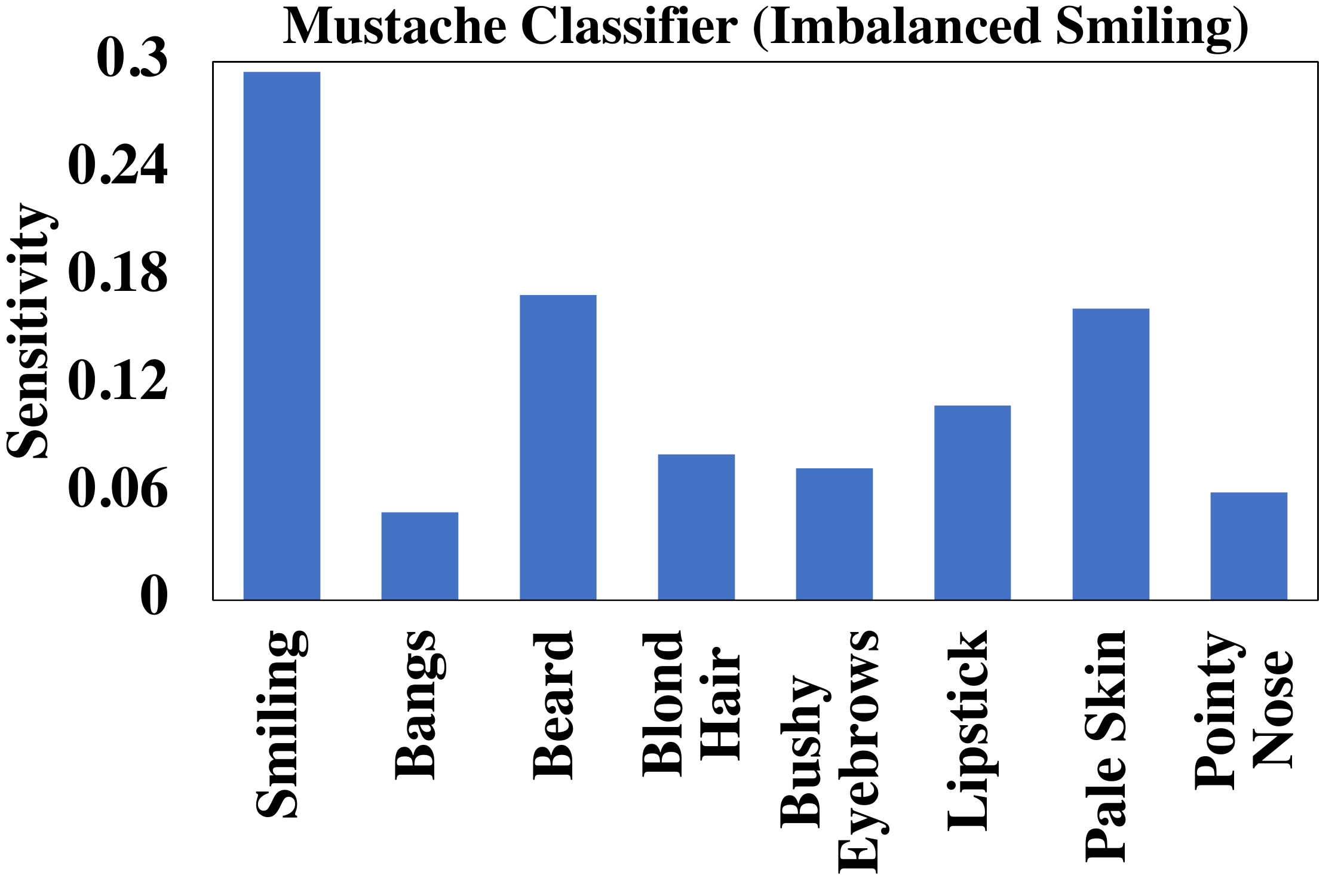}
    \caption{Model diagnosis histograms generated by \ourmodel on four classifiers trained on manually-crafted imbalance data.}
    \label{fig:histograms_unbalanced}
    \end{subfigure}
    \vspace{-5mm}
    \caption{Model diagnosis histograms generated by \ourmodel. The vertical axis values reflect the attribute sensitivities calculated by averaging the model probability change over all sampled images. The horizontal axis is the attribute space input by user.}
    \label{fig:histograms}
    \vspace{-3mm}
\end{figure*}

We synthesize a number of images from $ \mathcal{G}_\phi$, then iteratively compute the sensitivity for each given attribute, and finally normalize all sensitivities to draw the histogram as shown in Fig.~\ref{fig:histograms}. The histogram indicates the sensitivity of the evaluated model $f_{\theta}$ on each of the user-defined attributes. Higher sensitivity of one attribute means that the model is more easily affected by that attribute.

\begin{figure*}[h]
    \centering
    \includegraphics[width=0.49\linewidth]{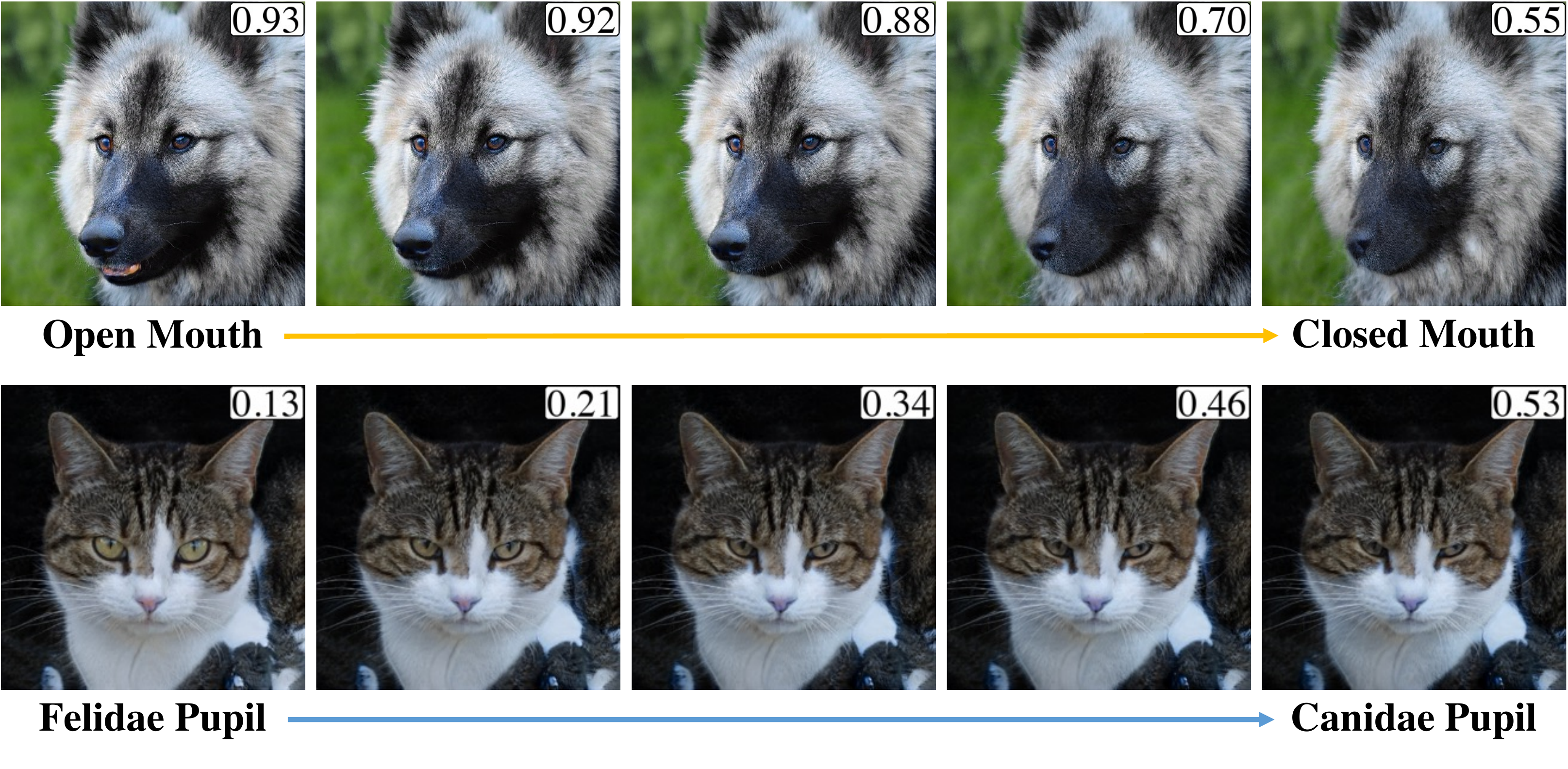}
    \includegraphics[width=0.49\linewidth]{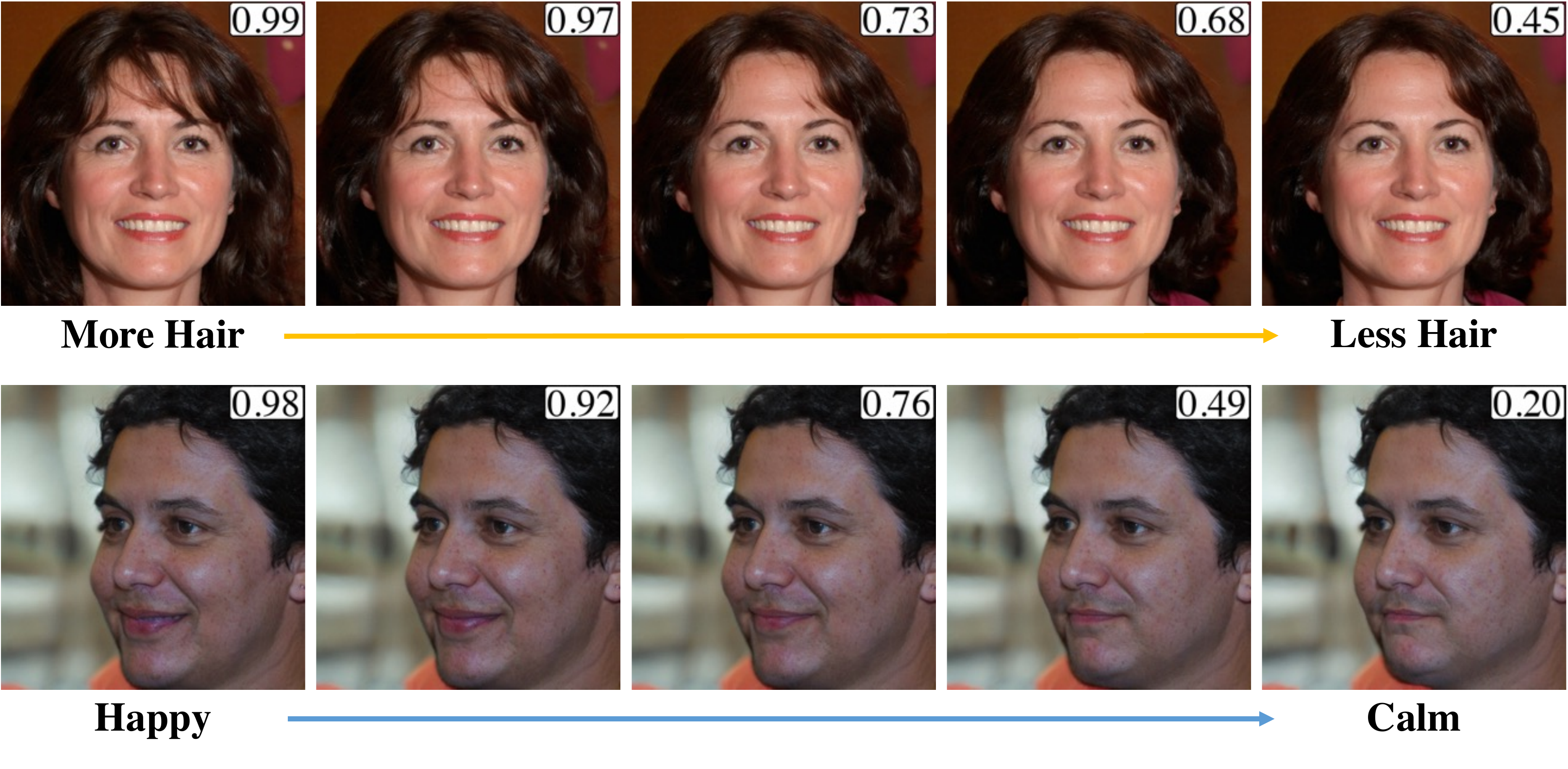}
    \vspace{-2mm}
    \caption{Effect of progressively generating counterfactual images on (left) cat/dog classifier (0-Cat / 1-Dog), and (right) perceived age classifier (0-Senior / 1-Young). Model probability prediction during the process is attached at the top right corner.}
    \label{fig:age_classifier_single}
    \vspace{-2mm}
\end{figure*}

\subsection{Counterfactual Training}
\label{sec:ct}

The multi-attribute counterfactual approach visualizes semantic combinations that cause the model to falter, providing valuable insights for enhancing the model's robustness.
We naturally adopt the concept of iterative adversarial training \cite{madry2018towards} to robustify the target model. For each iteration, \ourmodel receives the target model parameter and returns a batch of mutated counterfactual images with the model's original predictions as labels. Then the target model will be trained on the counterfactually-augmented images to achieve the robust goal:
\begin{equation}
\small
\label{eq:counterfactual_training}
    \theta^{*} = \argmin_{\theta} \mathbb{E}_{\mathbf{x}\sim\mathcal{P}(\mathbf{x}), \mathbf{\hat{x}} = \text{\ourmodel}(\mathbf{x}, \mathcal{A}) } {L}_{CE}(f_{\theta}(\mathbf{\hat{x}}), f_{\theta}(\mathbf{x}))
\end{equation}
where batches of $\mathbf{x}$ are randomly sampled from the StyleGAN generator $ \mathcal{G}_\phi$. 
In the following, we abbreviate the process as Counterfactual Training (CT). Note that, although not explicitly expressed in Eq.~\ref{eq:counterfactual_training}, the CT process is a min-max game. \ourmodel synthesizes counterfactuals to maximize the variation of model prediction (while persevering the perceived ground truth), and the target model is learned with the counterfactual images to minimize the variation.

\section{Experimental Results}
\label{sec:experimental_results}

This section describes the experimental validations on the effectiveness and reliability of \ourmodel. First, we describe the model setup in Sec.~\ref{sec:experiment_setups}. Sec.~\ref{sec:single_attr_diagnosis} and Sec.~\ref{sec:validation_diagnosis} visualize and validate the model diagnosis results for the single-attribute setting. In Sec.~\ref{sec:multiple_attr_diagnosis}, we show results on synthesized multiple-attribute counterfactual images and apply them to counterfactual training.

\subsection{Model Setup}
\label{sec:experiment_setups}
{\bf Pre-trained models:} We used Stylegan2-ADA \cite{Karras2020ada} pretrained on FFHQ \cite{2019stylegan} and AFHQ \cite{choi2020starganv2} as our base generative networks, and the pre-trained CLIP model \cite{CLIP}  which is parameterized by ViT-B/32. We followed StyleCLIP \cite{2021StyleCLIP} setups to compute the channel relevance matrices $\mathcal{M}$.

{\bf Target models:} Our classifier models are ResNet50 with single fully-connected head initialized by TorchVision\footnote{https://pytorch.org/blog/how-to-train-state-of-the-art-models-using-torchvision-latest-primitives/}. In training the binary classifiers, we use the Adam optimizer with learning rate 0.001 and batch size 128. We train binary classifiers for \textit{Eyeglasses, Perceived Gender, Mustache, Perceived Age} attributes on CelebA and for \textit{cat/dog} classification on AFHQ. For the 98-keypoint detectors, we used the HR-Net architecture~\cite{WangSCJDZLMTWLX19} on WFLW~\cite{wayne2018lab}. %Unless explicitly mentioned, our approach samples 1000 images from StyleGAN for each diagnosis by histogram.

\subsection{Visual Model Diagnosis: Single-Attribute}
\label{sec:single_attr_diagnosis}
Understanding where deep learning model fails is
an essential step towards building trustworthy models. Our proposed work allows us to generate counterfactual images (Sec.~\ref{sec:Counterfactual_Synthesis}) and provide insights on sensitivities of the target model (Sec.~\ref{sec:Attribute_Sensitivity_Analysis}). This section visualizes the counterfactual images in which only one attribute is modified at a time. 

Fig. \ref{fig:age_classifier_single} shows the single-attribute counterfactual images. Interestingly (but not unexpectedly), 
we can see that reducing the hair length or joyfulness causes the age classifier more likely to label the face to an older person. Note that our approach is extendable to multiple domains, as we change the cat-like pupil to dog-like, the dog-cat classification tends towards the dog. 
Using the counterfactual images, we can conduct model diagnosis with the method mentioned in Sec.~\ref{sec:Attribute_Sensitivity_Analysis}, on which attributes the model is sensitive to. In the histogram generated in model diagnosis, a higher bar means the model is more sensitive toward the corresponding attribute.

Fig.~\ref{fig:histograms_vanilla} shows the model diagnosis histograms on regularly-trained classifiers. For instance, the cat/dog classifier histogram shows outstanding sensitivity to green eyes and vertical pupil.
The analysis is intuitive since these are cat-biased traits rarely observed in dog photos. Moreover, the histogram of eyeglasses classifier shows that the mutation on bushy eyebrows is more influential for flipping the model prediction. 
It potentially reveals the positional correlation between eyeglasses and bushy eyebrows. The advantage of single-attribute model diagnosis is that the score of each attribute in the histogram are independent from other attributes, enabling unambiguous understanding of exact semantics that make the model fail. Diagnosis results for additional target models can be found in Appendix B.

\subsection{Validation of Visual Model Diagnosis} 
\label{sec:validation_diagnosis}
Evaluating whether our zero-shot sensitivity histograms (Fig.~\ref{fig:histograms}) explain the true vulnerability is a difficult task, since we do not have access to a sufficiently large and balanced test set fully annotated in an open-vocabulary setting. To verify the performance, we create synthetically imbalanced cases where the model bias is known. We then compare our results with a supervised diagnosis setting~\cite{sia}. In addition, we will validate the decoupling of the attributes by CLIP. 

\vspace{-2mm}
\subsubsection{Creating imbalanced classifiers}
\label{sec:creating_imbalance_classifiers}
\begin{figure}[t]
    \begin{subfigure}[b]{\linewidth}
        \label{fig:histogram_attgan_1}
         \centering
         \includegraphics[width=\linewidth]{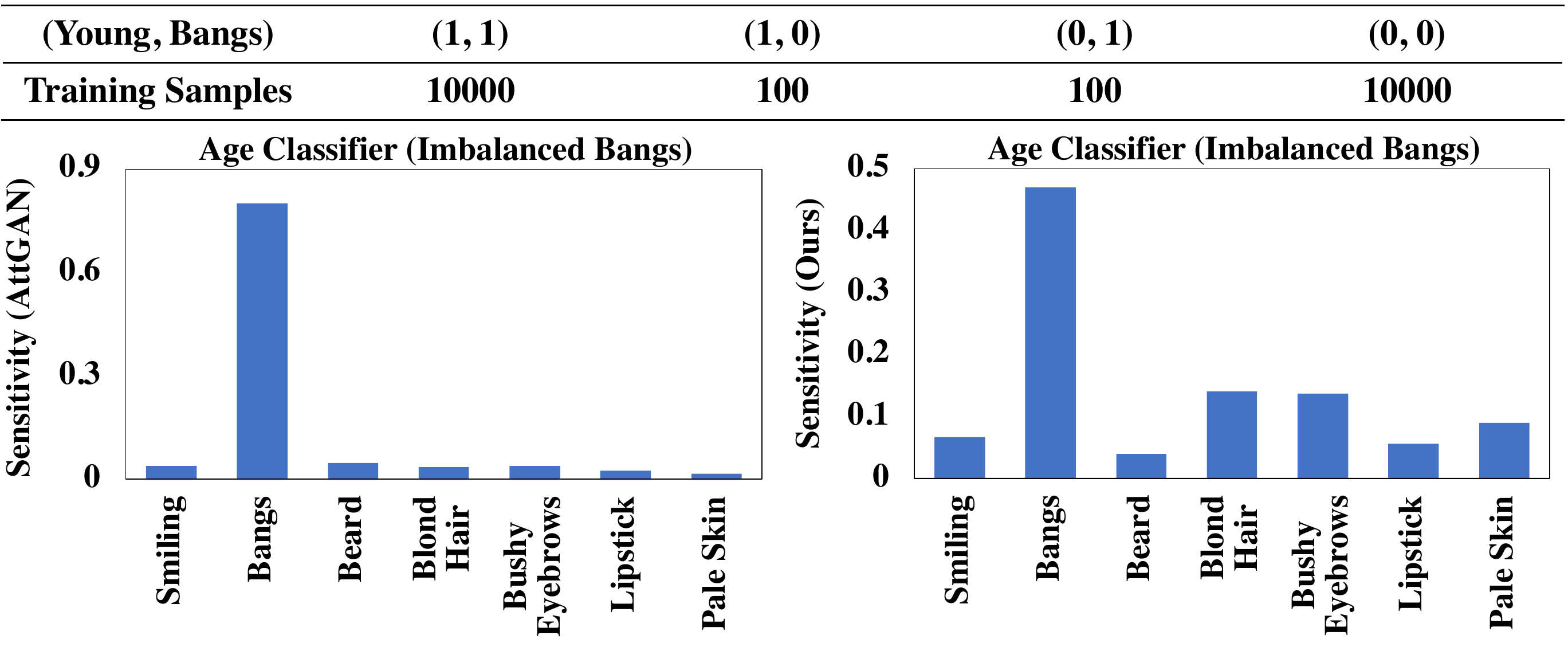}\\
    \end{subfigure}
    \begin{subfigure}[b]{\linewidth}
    \label{fig:histogram_attgan_2}
         \includegraphics[width=\linewidth]{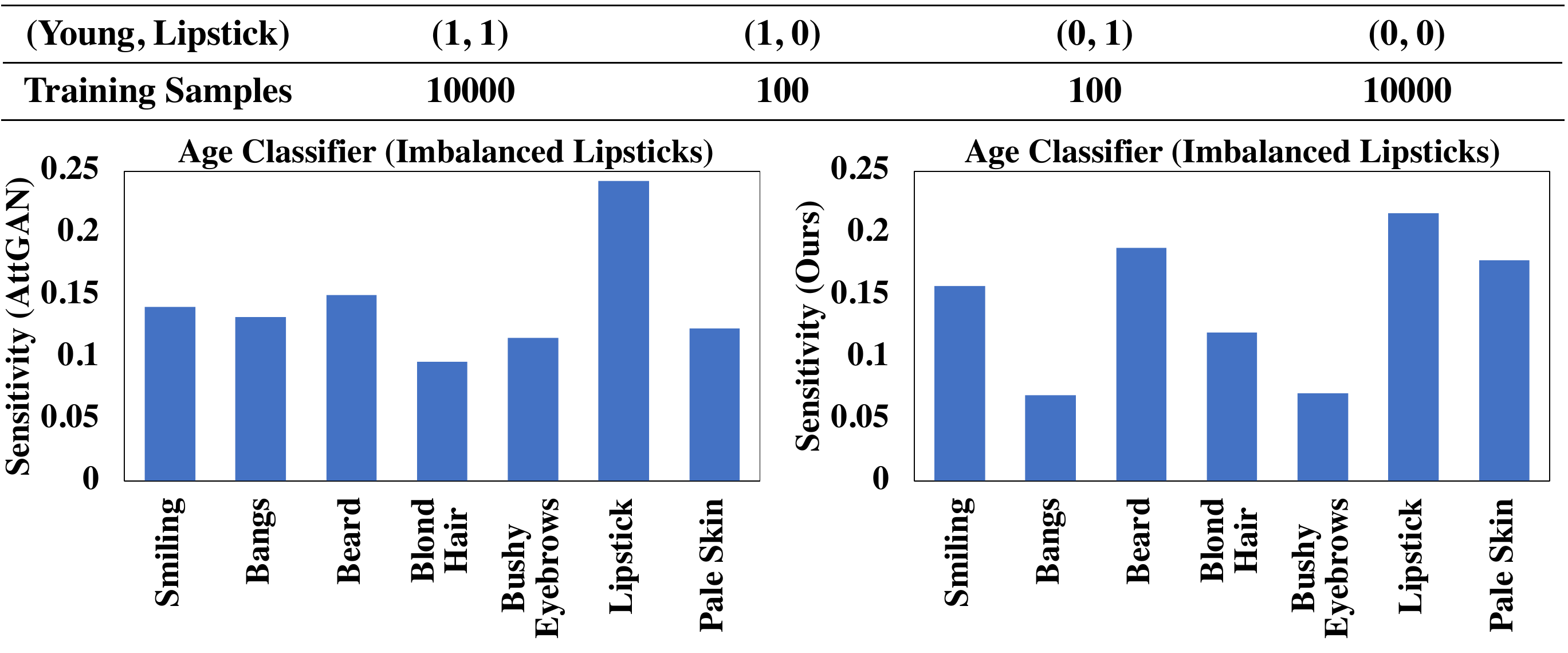}
    \end{subfigure}
        \vspace{-6mm}
         \caption{ The sensitivity of the age classifier is evaluated with \ourmodel (right) and AttGAN (left), achieving comparable results. }
         \label{fig:histogram_attgan}
         \vspace{-1mm}
    %  \end{subfigure}
\end{figure}

\vspace{-1mm}
In order to evaluate whether our sensitivity histogram is correct, we train classifiers that are highly imbalanced towards a known attribute and see whether \ourmodel can detect the sensitivity w.r.t the attribute. For instance, when training the perceived-age classifier (binarized as Young in CelebA), we created a dataset on which the trained classifier is strongly sensitive to Bangs (hair over forehead). The custom dataset is a CelebA training subset that consists of $20,200$ images. More specifically, there are $10,000$ images that have both young people that have bangs, represented as $(1,1)$, 
and $10,000$ images of people that are not young and have no bangs, represented as $(0,0)$. The remaining combinations of $(1,0)$ and $(0,1)$ have only 100 images.
With this imbalanced dataset, bangs is the attribute that dominantly correlates with whether the person is young, and hence the perceived-age classifier would be highly sensitive towards bangs.
% will learn that bangs is the most sensitive attribute to predict age. 
See Fig.~\ref{fig:histogram_attgan} (the right histograms) for an illustration of the sensitivity histogram computed by our method for the case of an age classifier with bangs (top) and lipstick (bottom) being imbalanced. 

 We trained multiple imbalanced classifiers with this methodology,  and visualize the model diagnosis histograms of these imbalanced classifiers in Fig.~\ref{fig:histograms_unbalanced}. We can observe that the \ourmodel histograms successfully detect the synthetically-made bias, which are shown as the highest bars in histograms. See the caption for more information. 

\begin{figure}[t]
    \begin{subfigure}[b]{0.49\linewidth}
        \centering
        \includegraphics[width=\linewidth]{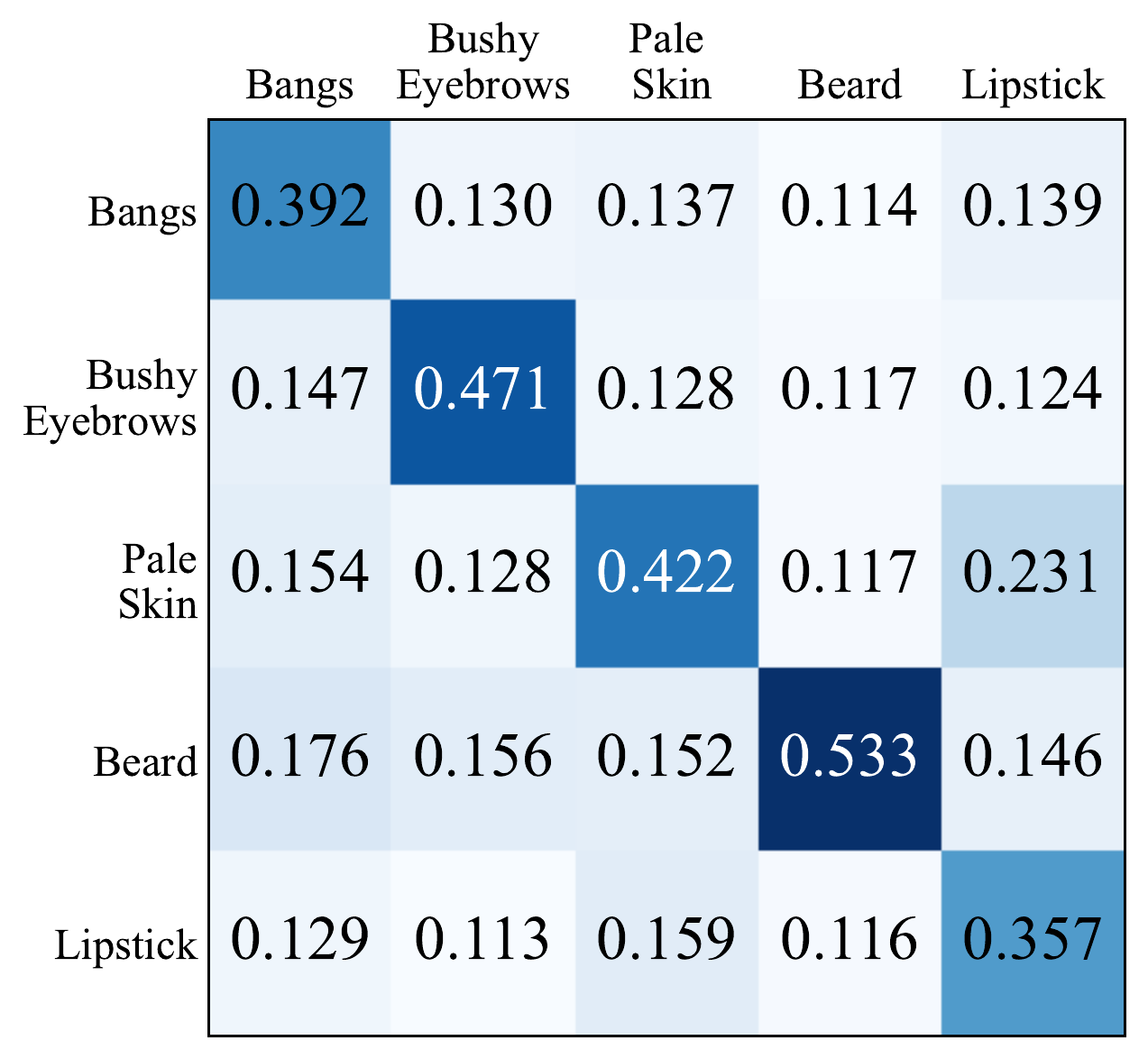}
        \caption{Mustache classifier}
        \label{fig:matrix_CLIP_Score_a}
    \end{subfigure}
    \begin{subfigure}[b]{0.49\linewidth}
        \centering
        \includegraphics[width=\linewidth]{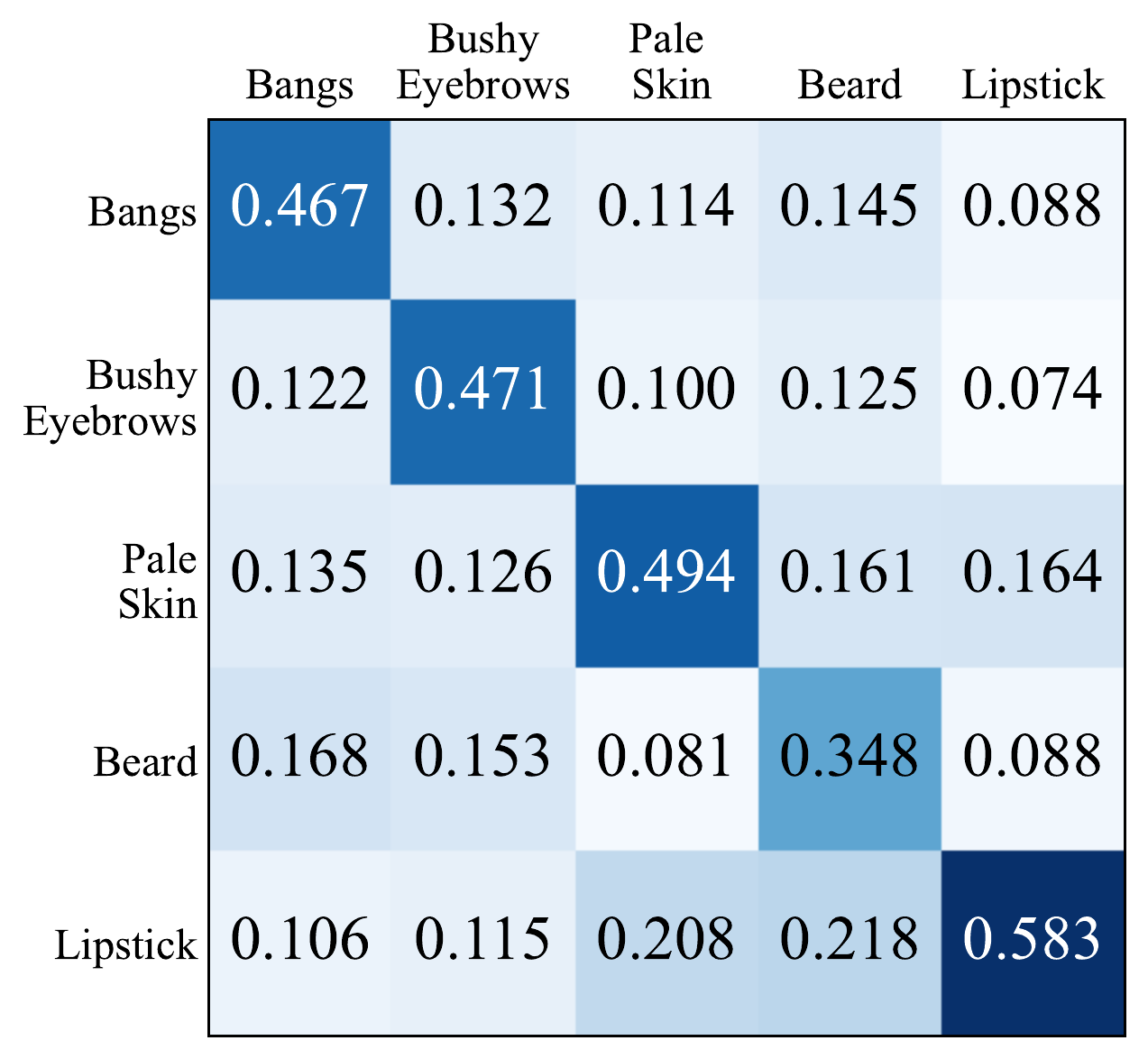}
        \caption{Perceived age classifier}
        \label{fig:matrix_CLIP_Score_b}
    \end{subfigure}
    \vspace{-2mm}
    \caption{Confusion matrix of CLIP score variation (vertical axis) when perturbing attributes (horizontal axis). This shows that attributes in \ourmodel are highly decoupled. }
    \label{fig:matrix_CLIP_Score}
    \vspace{-3mm}
\end{figure}

\begin{figure*}[ht]
    \centering
    \includegraphics[width=\linewidth]{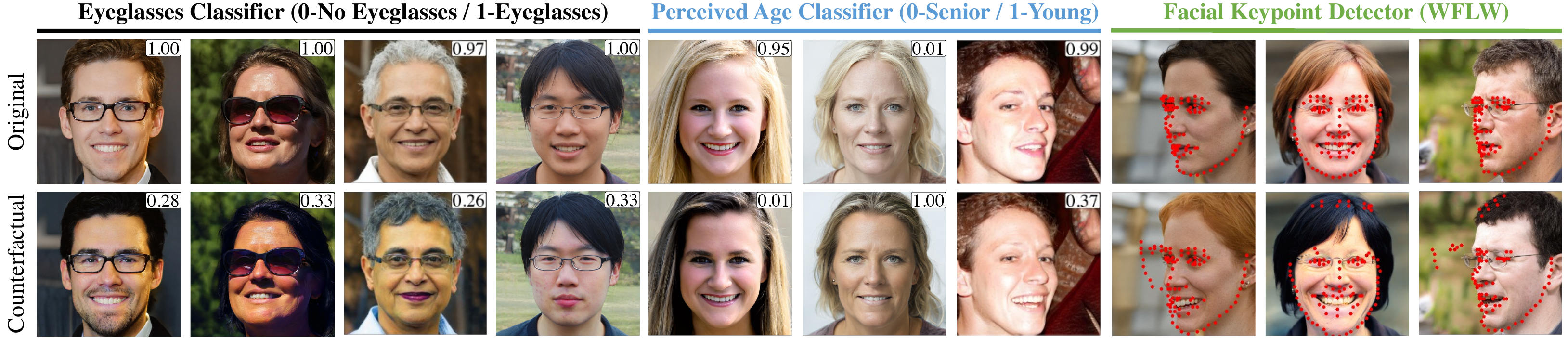}
    \caption{Multi-attribute counterfactual in faces. The model probability is documented in the upper right corner of each image.}
    \label{fig:human_classifier_multiattr}
    \vspace{-4mm}
\end{figure*}

\vspace{-2mm}
\subsubsection{Comparison with supervised diagnosis}
\vspace{-1mm}
We also validated our histogram by comparing it with the case in which we have access to a generative model that has been explicitly trained to disentangle attributes.  We follow the work on~\cite{sia} and used AttGAN~\cite{attGAN} trained on the CelebA training set over $15$ attributes\footnote{\textit{Bald, Bangs, Black\_Hair, Blond\_Hair, Brown\_Hair, Bushy\_Eyebrows, Eyeglasses, Male, Mouth\_Slightly\_Open, Mustache, No\_Beard, Pale\_Skin, Young, Smiling, Wearing\_Lipstick.}}.
After the training converged, we performed adversarial learning in the attribute space of AttGAN and create a sensitivity histogram using the same approach as Sec.~\ref{sec:Attribute_Sensitivity_Analysis}. Fig.~\ref{fig:histogram_attgan} shows the result of this method on the perceived-age classifier which is made biased towards bangs.  As anticipated, the AttGAN histogram (left) corroborates the histogram derived from our method (right). Interestingly, unlike \ourmodel, AttGAN show less sensitivity to remaining attributes. This is likely because AttGAN has a latent space learned in a supervised manner and hence attributes are better disentangled than with StyleGAN. Note that AttGAN is trained with a fixed set of attributes; if a new attribute of interest is introduced, the dataset needs to be re-labeled and AttGAN retrained. ZOOM, however, merely calls for the addition of a new text prompt.  More results in Appendix B.

\vspace{-2mm}
\subsubsection{Measuring disentanglement of attributes}
\vspace{-1mm}
Previous works demonstrated that the StyleGAN's latent space can be entangled~\cite{interfacegan, EditinginStyle}, adding undesired dependencies when searching single-attribute counterfactuals. This section verifies that our framework can disentangle the attributes and mostly edit the desirable attributes.

We use CLIP as a super annotator to measure attribute changes during single-attribute modifications. For $1,000$ images, we record the attribute change after performing adversarial learning in each attribute, and average the attribute score change. Fig.~\ref{fig:matrix_CLIP_Score} shows the confusion matrix during single-attribute counterfactual synthesis. The horizontal axis is the attribute being edited during the optimization, and the vertical axis represents the CLIP prediction changed by the process. For instance, the first column of Fig.~\ref{fig:matrix_CLIP_Score_a} is generated when we optimize over bangs for the mustache classifier. We record the CLIP prediction variation. It clearly shows that bangs is the dominant attribute changing during the optimization. From the main diagonal of matrices, it is evident that the \ourmodel mostly perturbs the attribute of interest. The results indicate reasonable disentanglement among attributes.

\subsection{Visual Model Diagnosis: Multi-Attributes}
\label{sec:multiple_attr_diagnosis}
\begin{figure}[t]
    \centering
    \includegraphics[width=\linewidth]{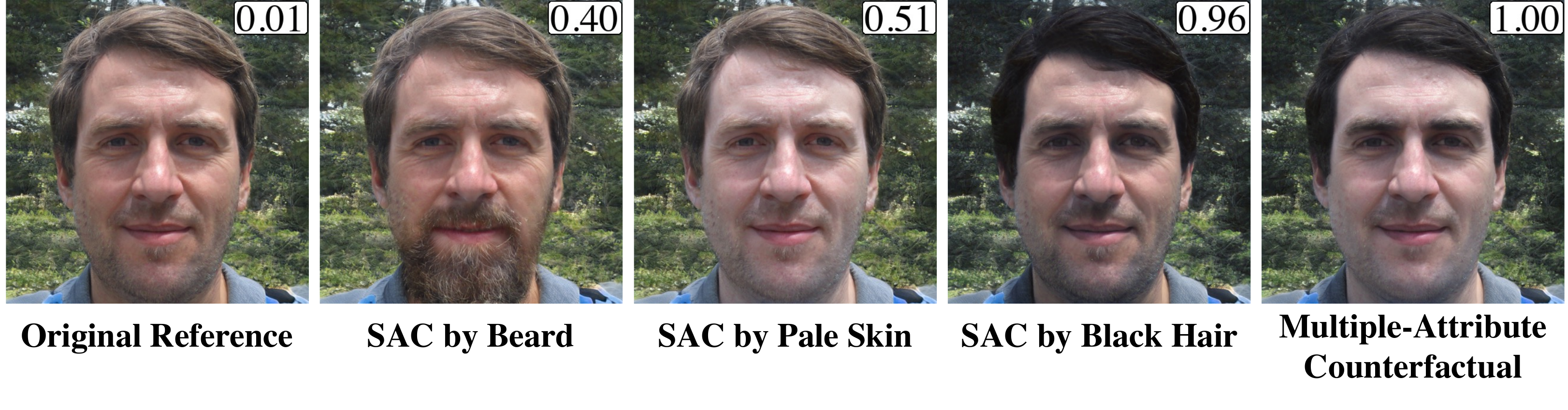}\\
    \vspace{-1mm}
    \includegraphics[width=\linewidth]{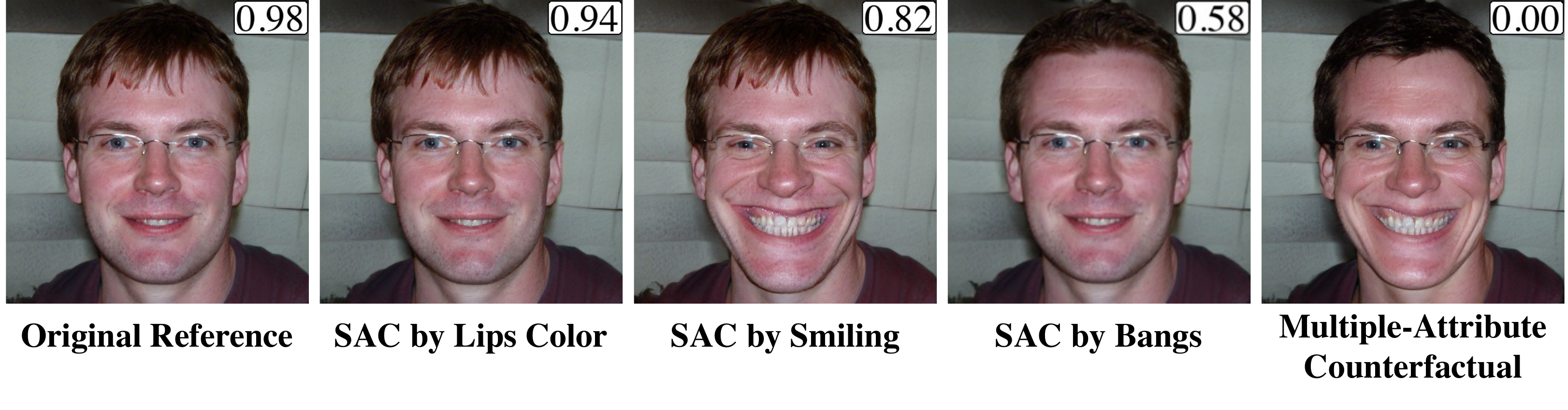}
    \vspace{-8mm}
    \caption{ Multiple-Attribute Counterfactual (MAC, Sec.~\ref{sec:multiple_attr_diagnosis}) compared with Single-Attribute Counterfactual (SAC, Sec.~\ref{sec:single_attr_diagnosis}). We can see that optimization along multiple directions enable the generation of more powerful counterfactuals.}
    \label{fig:multiple_attribute_is_more_powerful}
    \vspace{-4mm}
\end{figure}

In the previous sections, we have visualized and validated single-attribute model diagnosis histograms and counterfactual images. 
In this section, we will assess \ourmodel's ability to produce counterfactual images by concurrently exploring multiple attributes within $\mathcal{A}$, the domain of user-defined attributes.  The approach conducts multi-attribute counterfactual searches across various edit directions, identifying distinct semantic combinations that result in the target model's failure. By doing so, we can effectively create more powerful counterfactuals images (see Fig.~\ref{fig:multiple_attribute_is_more_powerful}).

\begin{figure}[t]
    \centering
    \includegraphics[width=\linewidth]{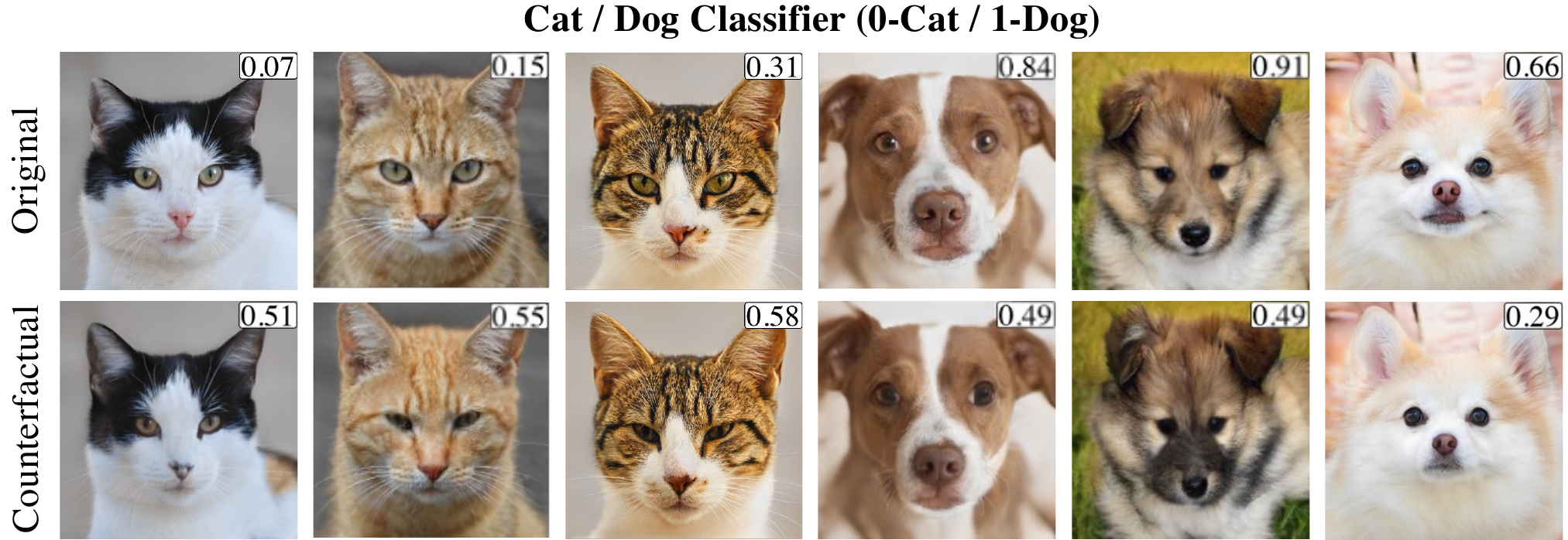}
    \caption{Multi-attribute counterfactual on Cat/Dog classifier. The number in each image is the predicted probability of being a dog.}
    \label{fig:dog_classifier_multiattr}
    \vspace{-2mm}
\end{figure}

Fig.~\ref{fig:human_classifier_multiattr} and Fig.~\ref{fig:dog_classifier_multiattr} show examples of multi-attribute counterfactual
images generated by \ourmodel, against human and animal face classifiers. 
It can be observed that multiple face attributes such as lipsticks or hair color are edited in Fig.~\ref{fig:human_classifier_multiattr}, and various cat/dog attributes like nose pinkness, eye shape, and fur patterns are edited in Fig.~\ref{fig:dog_classifier_multiattr}. 
These attribute edits are blended to affect the target model prediction. Appendix B further illustrates \ourmodel counterfactual images for semantic segmentation, multi-class classification, and a church classifier. By mutating semantic representations, \ourmodel reveals semantic combinations as outliers where the target model underfits.

In the following sections, we 
will use the Flip Rate (the percentage of counterfactuals that flipped the model prediction) and Flip Resistance (the percentage of counterfactuals for which the model successfully withheld its prediction) to evaluate the multi-attribute setting. 
\begin{figure}[t]
    \centering
    \begin{subfigure}[b]{\linewidth}
    \includegraphics[width=0.495\linewidth]{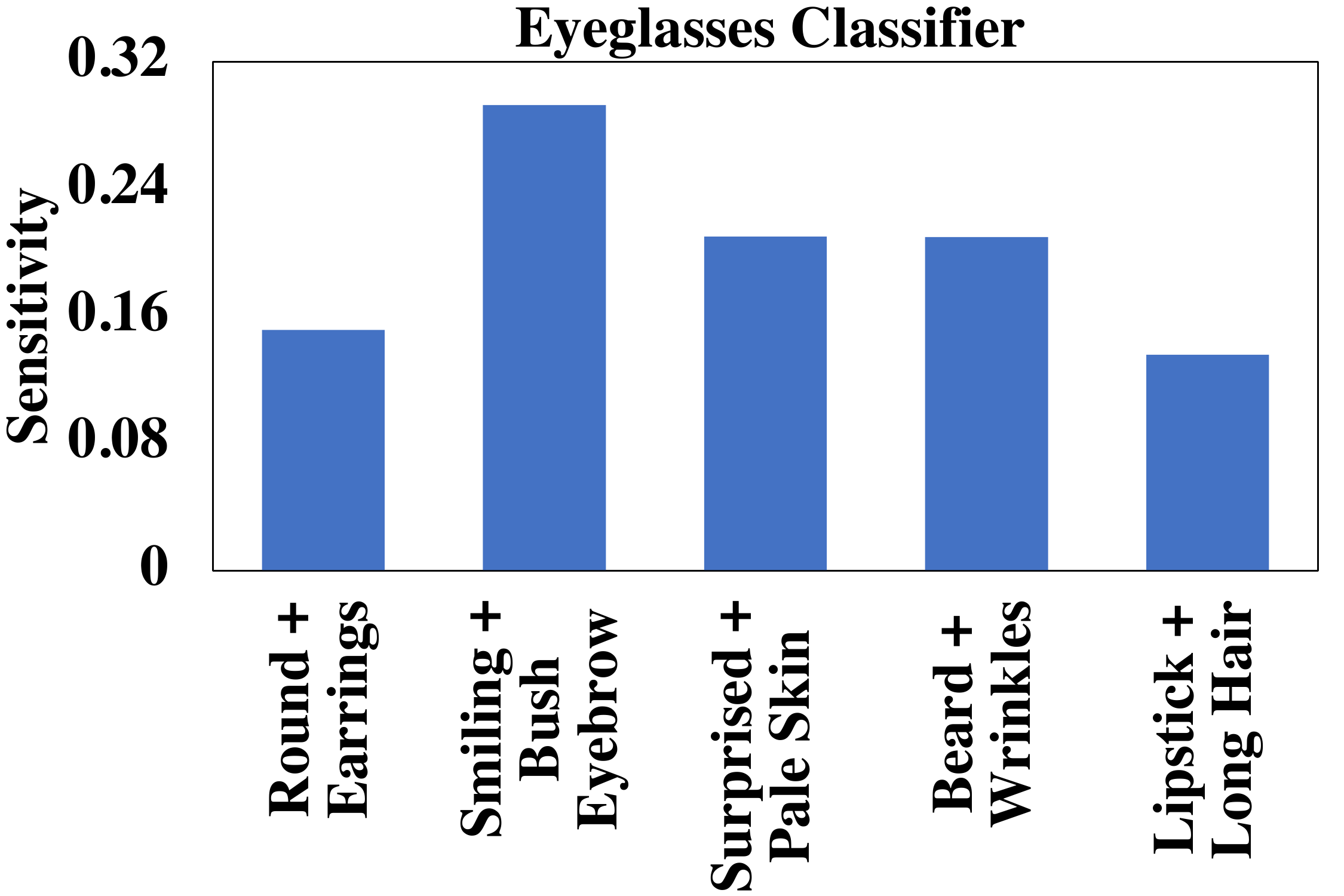}
    \includegraphics[width=0.495\linewidth]{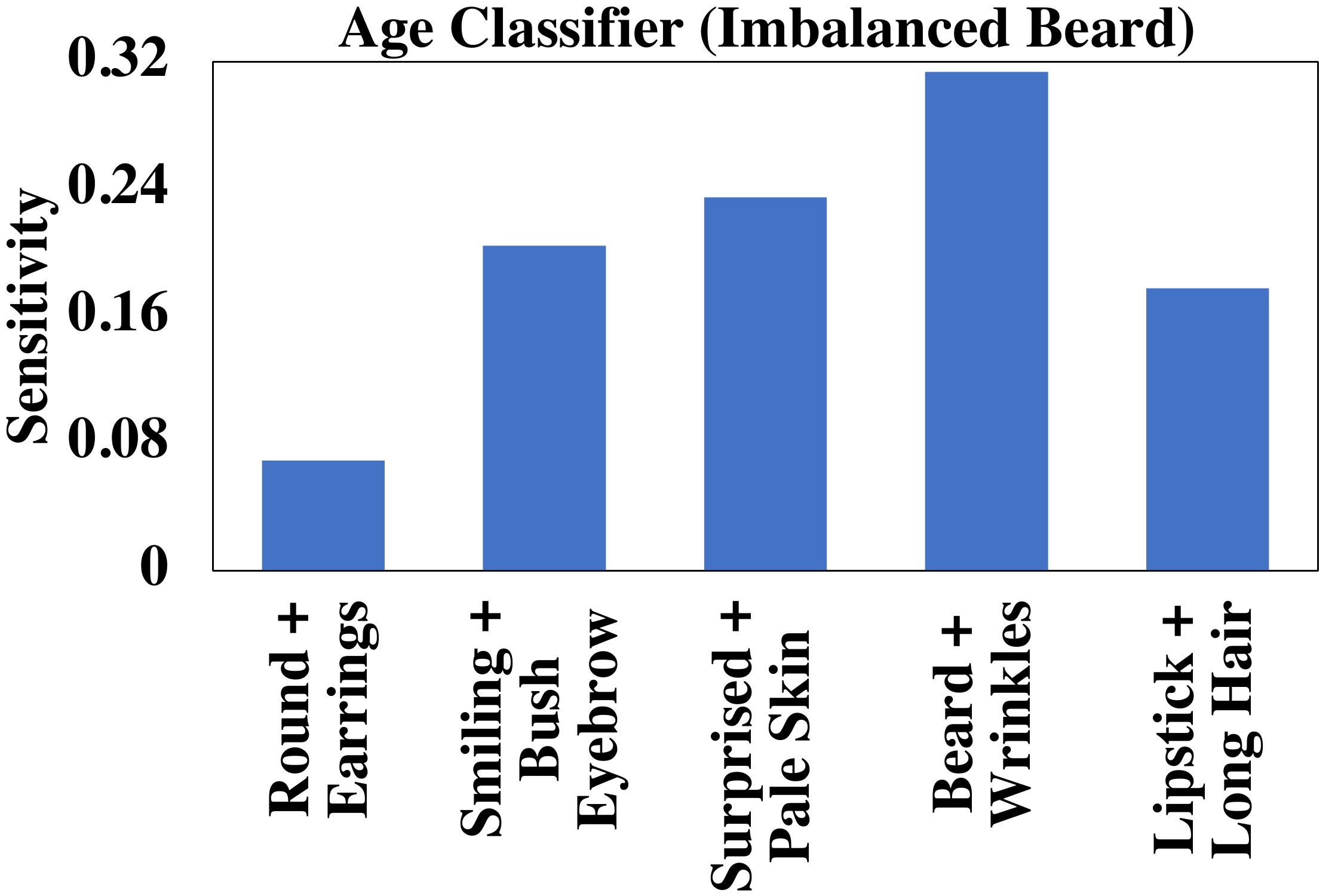}
    \caption{Sensitivity histograms generated by \ourmodel on attribute combinations.}
    \label{fig:histograms_combination}
    \end{subfigure}\\
    \begin{subfigure}[b]{\linewidth}
    \includegraphics[width=\linewidth]{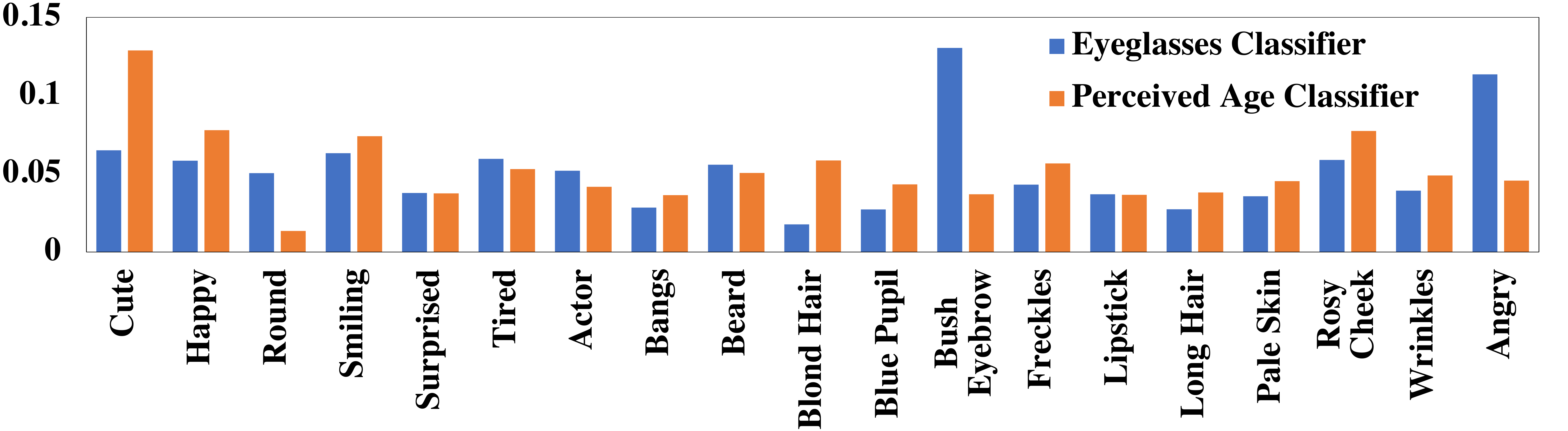}
    \caption{Model diagnosis by \ourmodel over $19$ attributes. Our framework is generalizable to analyze facial attributes of various domains.}
    \label{fig:histograms_grand}
    \end{subfigure}
    \vspace{-6mm}
    \caption{Customizing attribute space for \ourmodel.}
    \label{fig:multiple_attribute_histogram}
    \vspace{-4mm}
\end{figure}
\vspace{-3mm}
\subsubsection{Customizing attribute space}
\vspace{-2mm}
\looseness=-1

In some circumstances,  users may finish one round of model diagnosis and proceed to another round by adding new attributes, or trying a new attribute space.
The linear nature of attribute editing (Eq.~\ref{eq:total_edit}) in \ourmodel makes it possible to easily add or remove attributes. 
Table~\ref{tab:model_flip_rate} shows the flip rates results when adding new attributes into $\mathcal{A}$ for perceived age classifier and big lips classifier.  We can observe that a different attribute space will results in different effectiveness of counterfactual images. Also, increasing the search iteration will make counterfactual more effective (see last row). 
 Note that neither re-training the StyleGAN nor user-collection/labeling of data is required at any point in this procedure.  Moreover, Fig.~\ref{fig:histograms_combination} shows the model diagnosis histograms generated with combinations of two attributes. Fig.~\ref{fig:histograms_grand} demonstrates the capability of \ourmodel in a rich vocabulary setting where we can analyze attributes that are not labeled in existing datasets~\cite{liu2015celeba,MAAD}.
 
\vspace{-4mm}
\subsubsection{Counterfactual training results}
\label{sec:ct_result}
\vspace{-1mm}

This section evaluates regular classifiers trained on CelebA~\cite{liu2015celeba} and counterfactually-trained (CT) classifiers on a mix of CelebA data and counterfactual images as described in Sec.~\ref{sec:ct}. Table \ref{tab:ct_acc_table} presents accuracy and flip resistance (FR) results. CT outperforms the regular classifier. FR is assessed over 10,000 counterfactual images, with FR-25 and FR-100 denoting Flip Resistance after 25 and 100 optimization iterations, respectively. Both use $\eta=0.2$ and $\epsilon=30$. We can observe that the classifiers after CT are way less likely to be flipped by counterfactual images while maintaining a decent accuracy on the CalebA testset. Our approach robustifies the model by increasing the tolerance toward counterfactuals. Note that CT slightly improves the CelebA classifier when trained on a mixture of CelebA images (original images) and the counterfactual images generated with a generative model  trained in the FFHQ~\cite{2019stylegan} images (different domain).

\begin{table}[t]
  \centering
  \footnotesize
  \begin{tabular}{@{}lccc@{}}
     \toprule
     Method & \makecell{AC Flip Rate (\%)} & \makecell{BC Flip Rate (\%)} \\
     \midrule
     Initialize \ourmodel by $\mathcal{A}$                        & 61.95 &  83.47\\
     + Attribute: Beard                                           &  72.08 & 90.07\\
     + Attribute: Smiling                                        &  87.47 &  \textbf{96.27}\\
     + Attribute: Lipstick                                         &  90.96 &  94.07\\
     + Iterations increased to 200                                &  \textbf{92.91} &  94.87\\
     \bottomrule
  \end{tabular}
  \caption{\label{tab:model_flip_rate} Model flip rate study. The initial attribute space $\mathcal{A} =$ \{Bangs, Blond Hair, Bushy Eyebrows, Pale Skin, Pointy Nose\}. AC is the perceived age classifier and BC is the big lips classifier.} 
  \vspace{-3mm}
\end{table}

\begin{table}[t]
    \centering
    \footnotesize
    \begin{tabular}{ccccc}
        \toprule
         Attribute & \makecell{Metric} & \makecell{Regular (\%)} & \makecell{CT (Ours) (\%)} \\

\midrule
        \multirow{3}{*}{Perceived Age} & CelebA Accuracy   & 86.10 & \textbf{86.29}   \\
        & \ourmodel FR-25  & 19.54 & \textbf{97.36}  \\
        & \ourmodel FR-100  & 9.04 & \textbf{95.65}  \\
        \midrule
        \multirow{3}{*}{Big Lips} & CelebA Accuracy   & 74.36 & \textbf{75.39}    \\
        & \ourmodel FR-25  & 14.12 & \textbf{99.19}  \\
        & \ourmodel FR-100  & 5.93 & \textbf{88.91}  \\
        \bottomrule
    \end{tabular}
    \caption{\label{tab:ct_acc_table} Results of network inference on CelebA original images and \ourmodel-generated counterfactual. The CT classifier is significantly less prone to be flipped by counterfactual images, while test accuracy on CelebA remains performant.}
    \vspace{-6mm}
\end{table}

\vspace{-2mm}

\section{Conclusion and Discussion} \label{conclusion_and_future}
\looseness=-1
\vspace{-2mm}

In this paper, we present \ourmodel, a zero-shot model diagnosis framework that generates sensitivity histograms based on 
user's input of natural language attributes. 
\ourmodel assesses failures and generates corresponding sensitivity histograms for each attribute.  A significant advantage
of our technique is its ability to analyze the failures of a target deep model without the need for laborious collection and annotation of test sets. \ourmodel effectively visualizes the correlation between attributes and model outputs, elucidating model behaviors and intrinsic biases.

Our work has three primary limitations. First, users should possess domain knowledge as their input (text of attributes of interest) should be relevant to the target domain.  Recall that it is a small price to pay for model evaluation without an annotated test set. Second, StyleGAN2-ADA struggles with generating out-of-domain samples. Nevertheless, our adversarial learning framework can be adapted to other generative models (e.g., stable diffusion), and the generator can be improved by training on more images. We have rigorously tested our generator with various user inputs, confirming its effectiveness for regular diagnosis requests. Currently, we are exploring stable diffusion models to generate a broader range of classes while maintaining the core concept. Finally, we rely on a pre-trained model like CLIP which we presume to be free of bias and capable of encompassing all relevant attributes.

{\bf Acknowledgements: }We would like to thank George Cazenavette, Tianyuan Zhang, Yinong Wang, Hanzhe Hu, Bharath Raj for suggestions in the presentation and experiments. We sincerely thank Ken Ziyu Liu, Jiashun Wang, Bowen Li, and Ce Zheng for revisions to improve this work.

%------------------------------------------------------------------------

%%%%%%%%% REFERENCES
{\small
\bibliographystyle{ieee_fullname}
\bibliography{PaperForArXiv}
}

% %------------------------------------------------------------------------
\clearpage
\newpage
\appendix
\section{Validation for CLIP-guided Editing}
\label{sec:appendix_sectiona}

Our methodology relies on CLIP-guided fine-grained image editing to provide adequate model diagnostics. It is critical to verify CLIP's ability to link language and visual representations. This section introduces two techniques for validating CLIP's capabilities.

\subsection{Visualization for edited images}
In this section, we analyze the decoupling of attribute editing used in StyleCLIP~\cite{2021StyleCLIP} in our domain.

\begin{figure*}[h]
    \centering
    \captionsetup{justification=centering}
    \begin{subfigure}[t]{0.60\linewidth}
        \includegraphics[width=\linewidth]{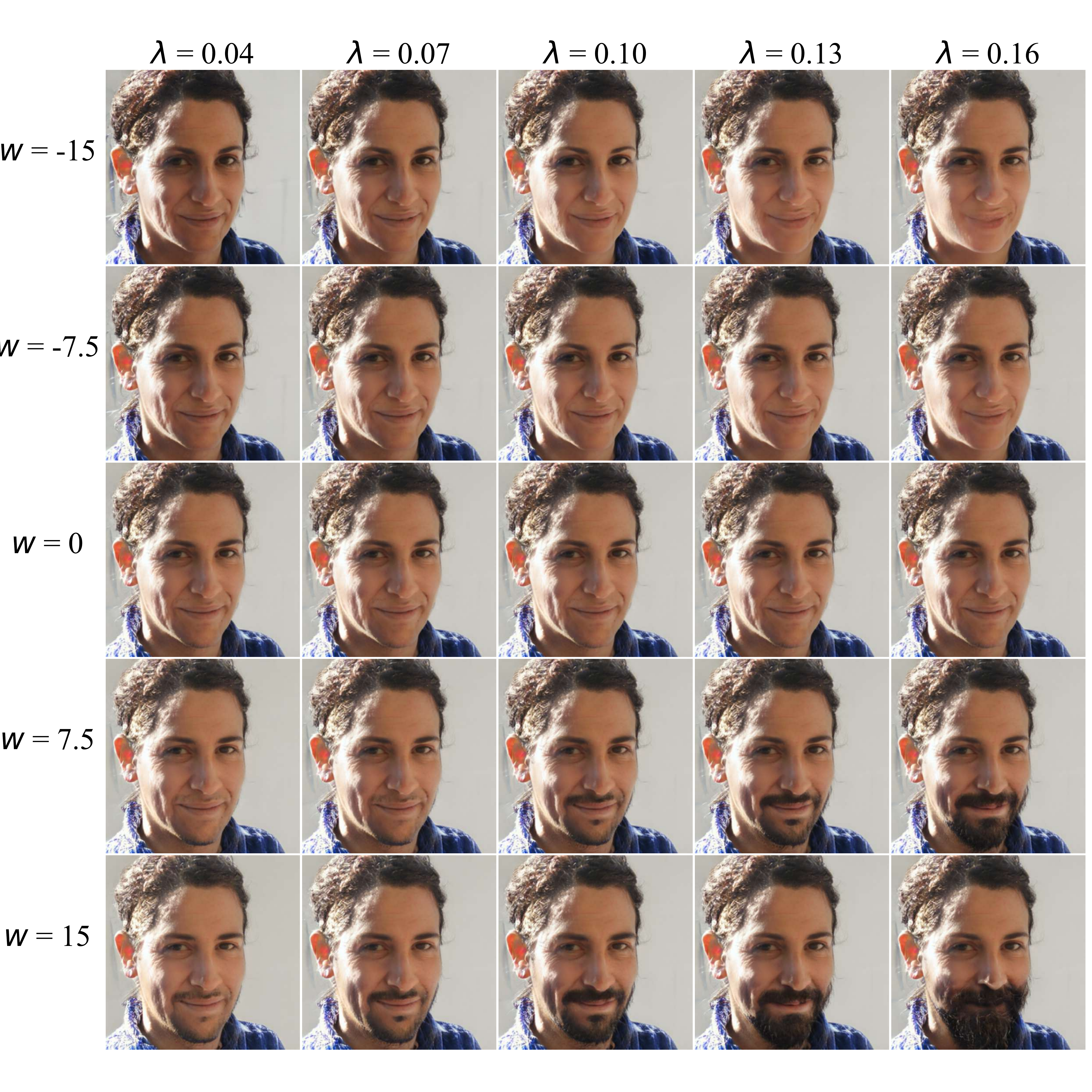}
    \caption{Effect of $\lambda$ values for editing beard.}
    \label{fig:appendix_lambda_1}
    \end{subfigure}\\
    \begin{subfigure}[t]{0.60\linewidth}
        \centering
        \captionsetup{justification=centering}
        \includegraphics[width=\linewidth]{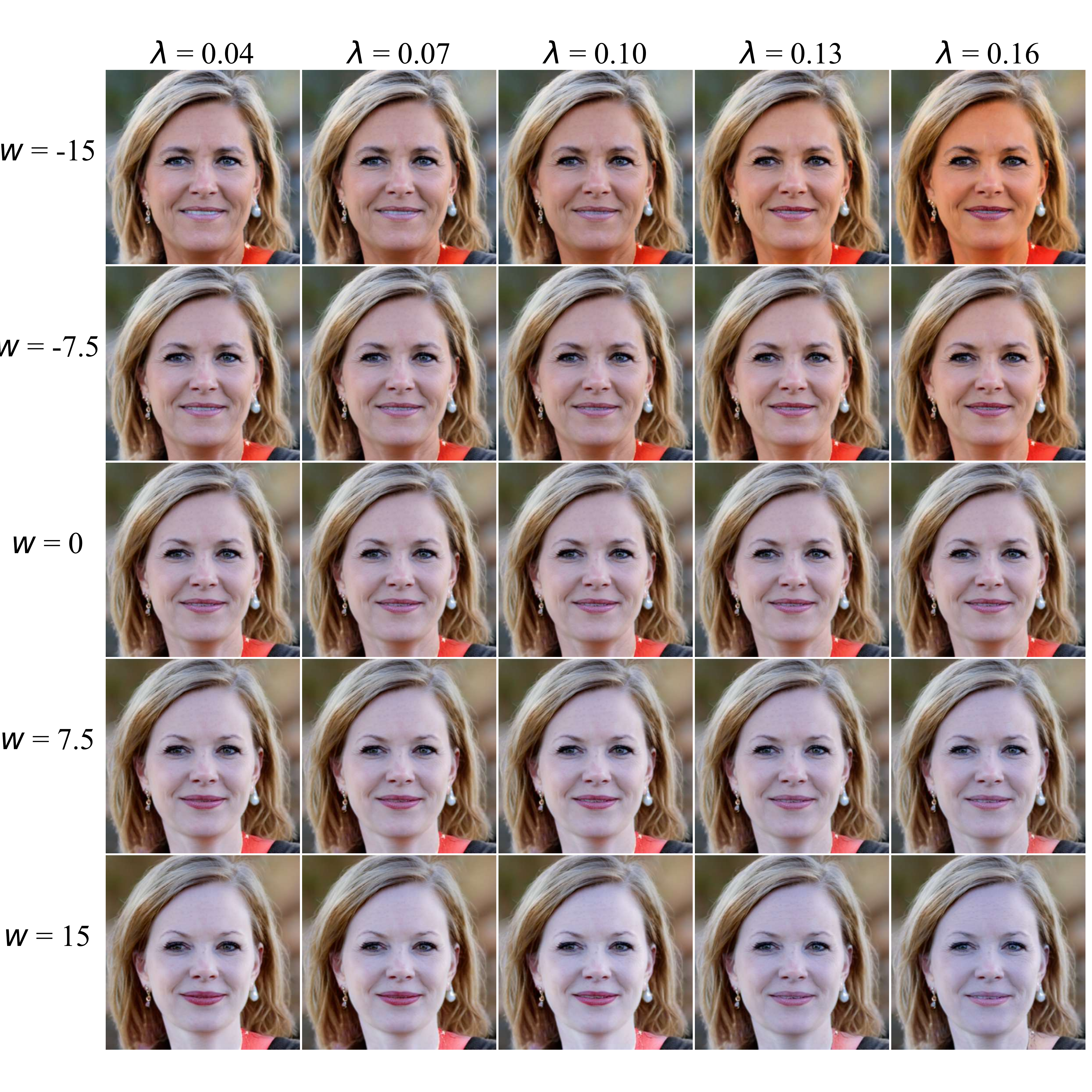}
        \caption{Effect of $\lambda$ values for editing pale skin.}
        \label{fig:appendix_lambda_2}
    \end{subfigure}
\caption{Visualization of the effect of different $\lambda$ values.}
\label{fig:appendix_lambda}
\end{figure*}

\textbf{Effect of $\lambda$.} Fig.~\ref{fig:appendix_lambda} shows the effect of $\lambda$ in Equation 2 of the main text~\cite{2021StyleCLIP} . 
Originally in StyleCLIP, this filter parameter (denoted as $\beta$ in~\cite{2021StyleCLIP}) helps the style disentanglement for editing. 
As we have normalized the edit vectors, which contributes to disentanglement in our framework, the impact of $\lambda$ on style disentanglement is reduced. Consequently, $\lambda$ primarily influences intensity control and denoising.

\begin{figure*}[t]
\centering
    \begin{subfigure}[b]{0.48\linewidth}
        \centering
        \includegraphics[width=\linewidth]{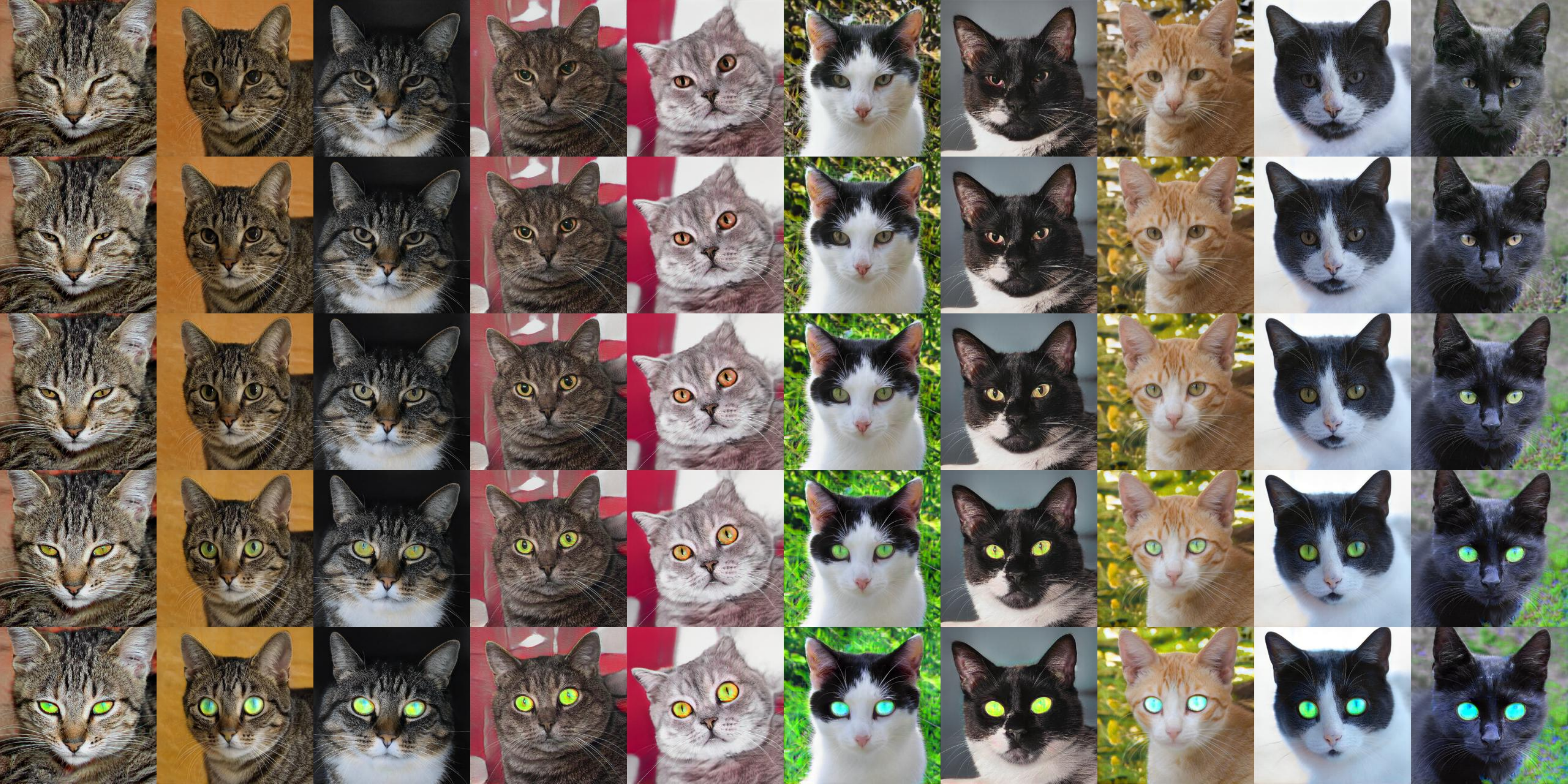}
        \caption{Attribute editing: a cat with green eyes.}
    \end{subfigure}
    \begin{subfigure}[b]{0.48\linewidth}
        \centering
        \includegraphics[width=\linewidth]{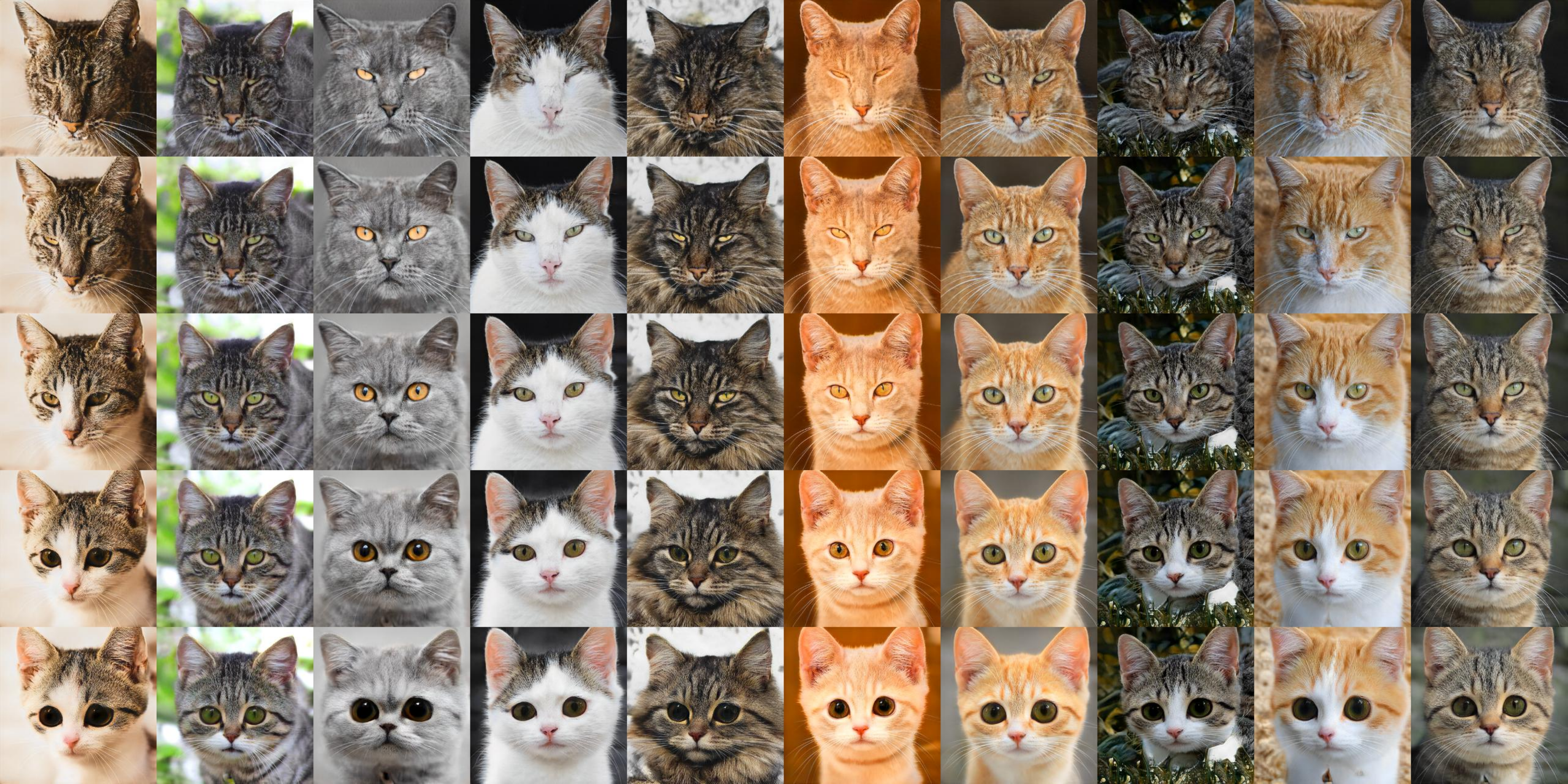}
        \caption{Attribute editing: a cute cat.}
    \end{subfigure}\\
    \begin{subfigure}[b]{0.48\linewidth}
        \centering
        \includegraphics[width=\linewidth]{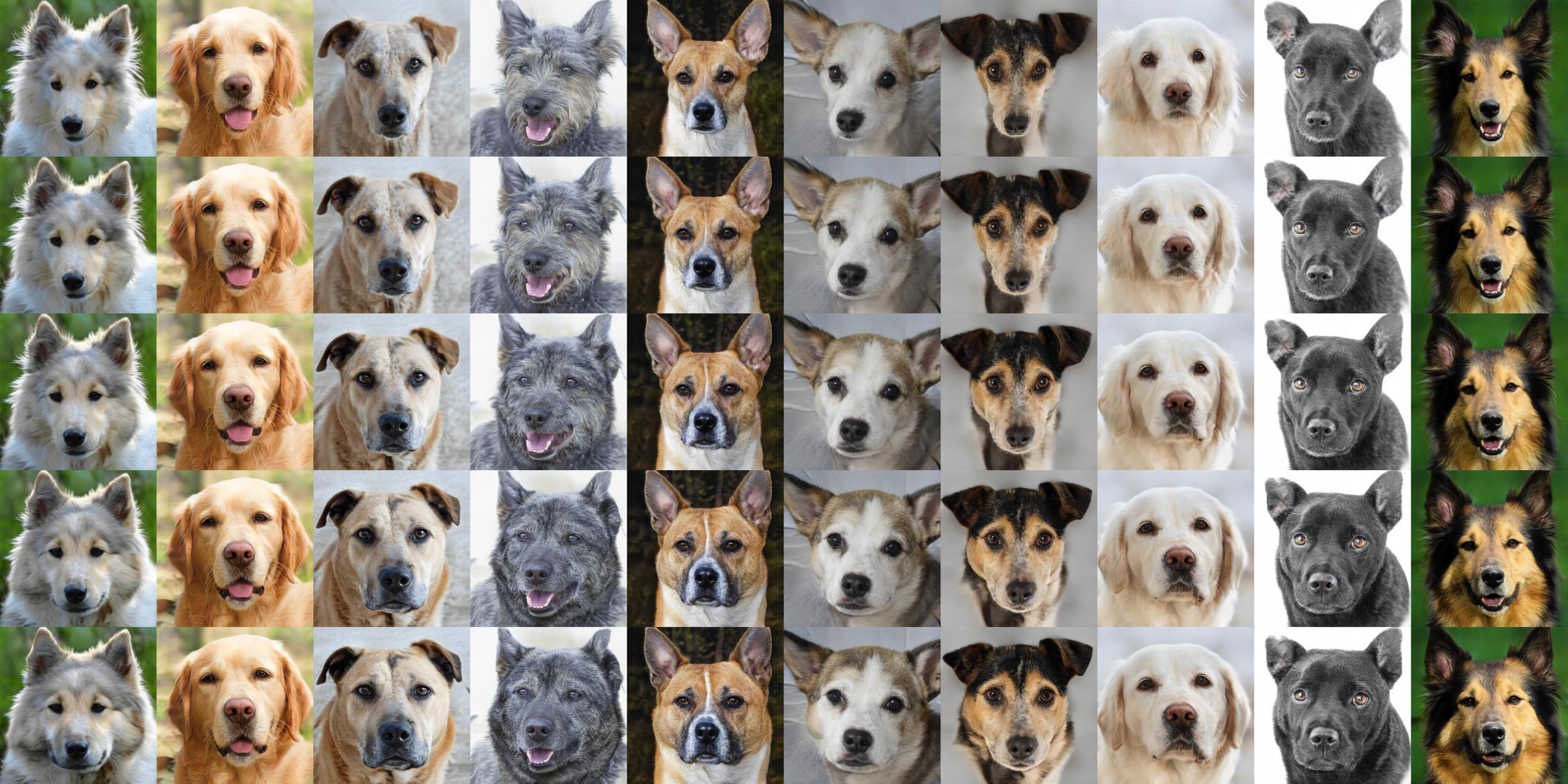}
        \caption{Attribute editing: a dog with round face.}
    \end{subfigure}
    \begin{subfigure}[b]{0.48\linewidth}
        \centering
        \includegraphics[width=\linewidth]{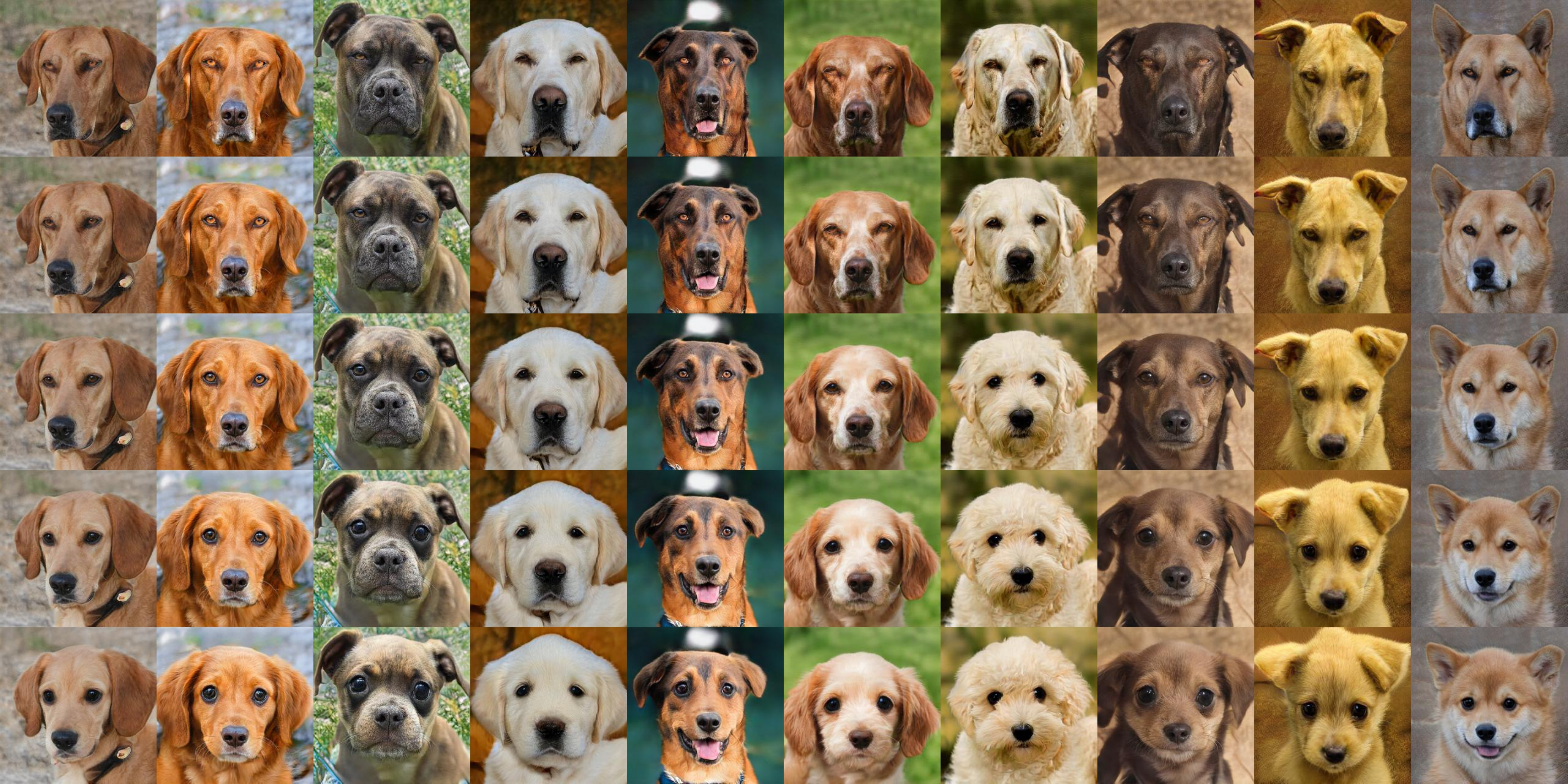}
        \caption{Attribute editing: a cute dog.}
    \end{subfigure}\\
    \begin{subfigure}[b]{0.48\linewidth}
        \centering
        \includegraphics[width=\linewidth]{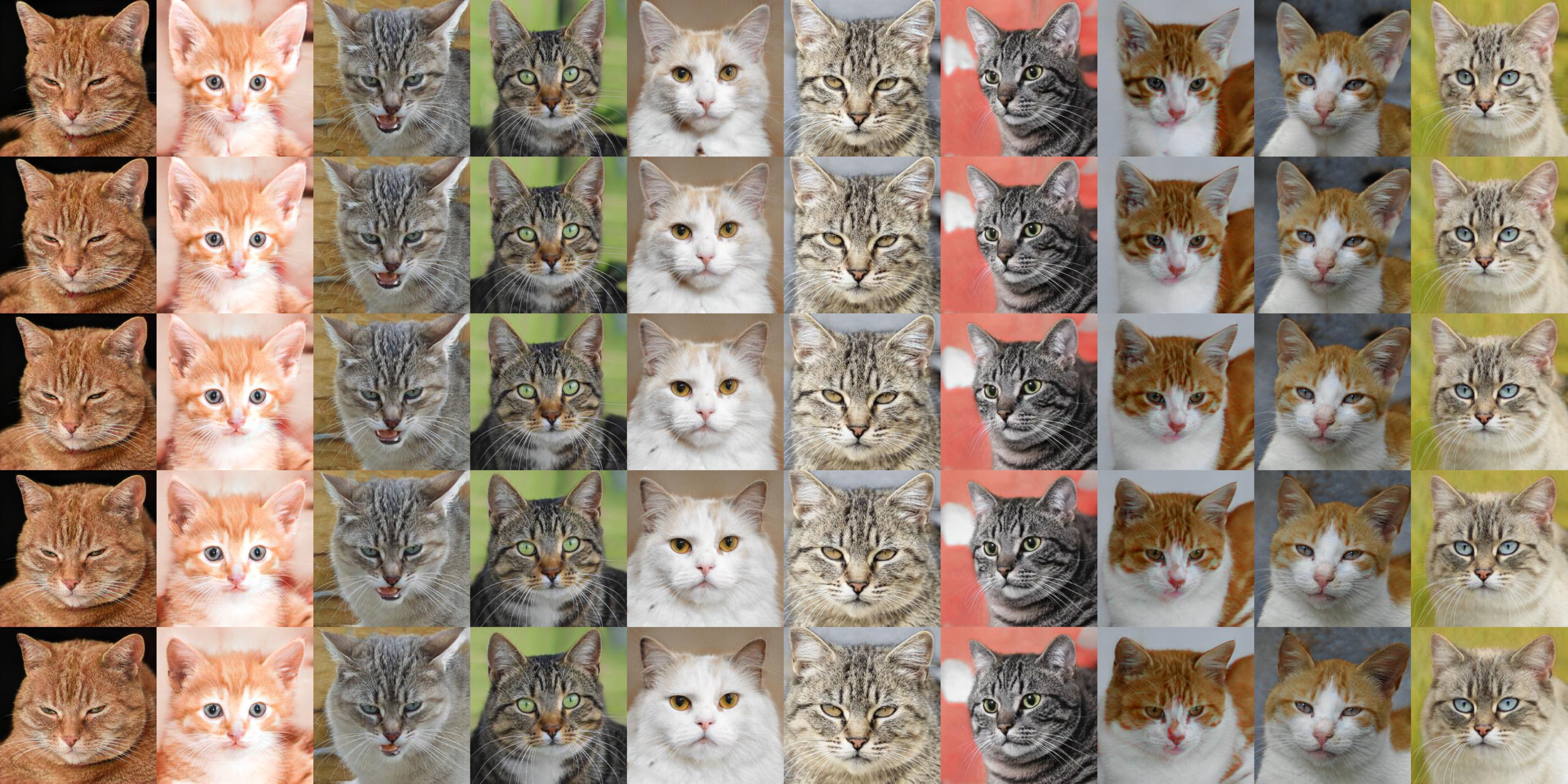}
        \caption{Attribute editing: a cat with round face.}
    \end{subfigure}
    \begin{subfigure}[b]{0.48\linewidth}
        \centering
        \includegraphics[width=\linewidth]{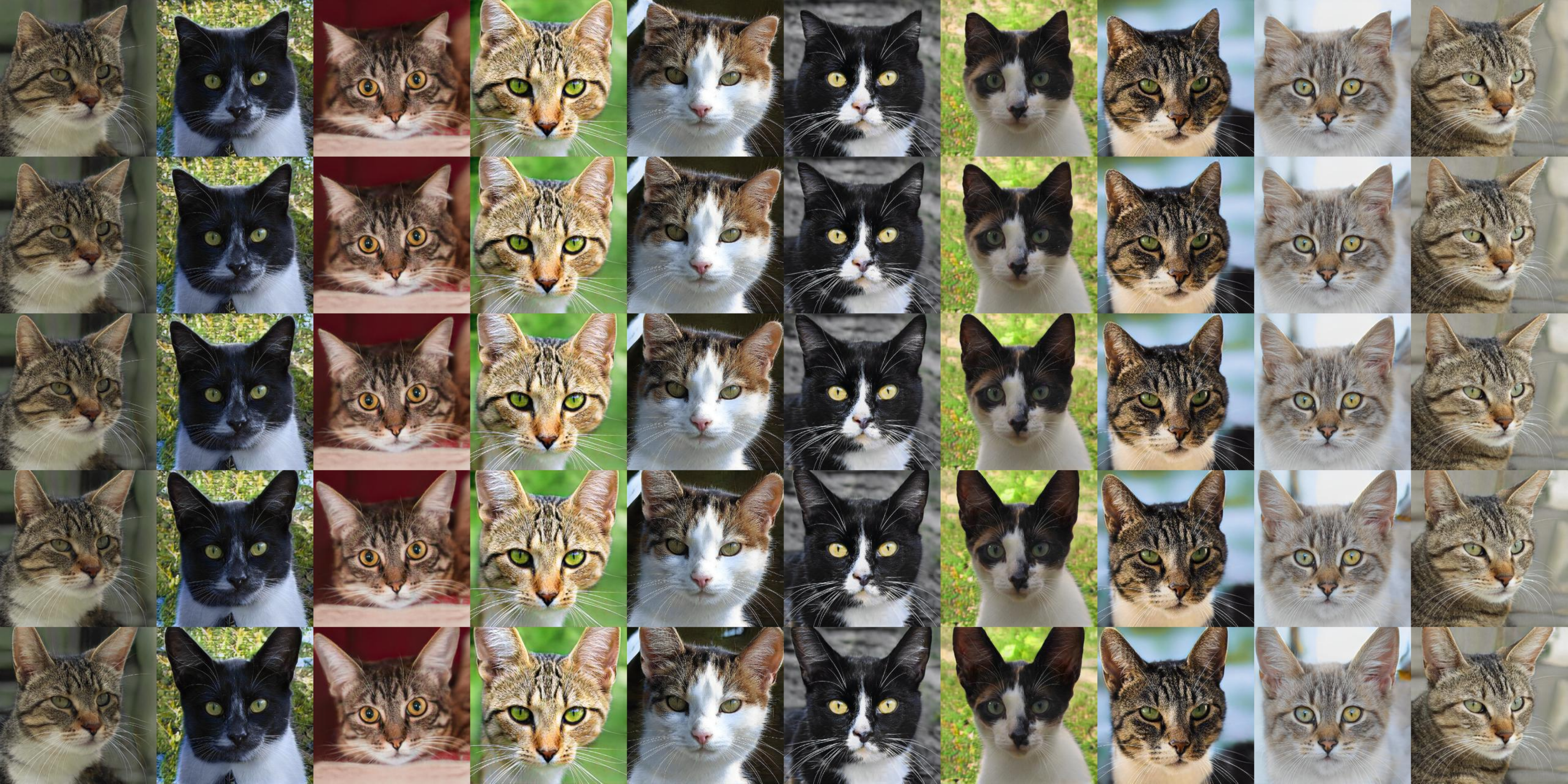}
        \caption{Attribute editing: a cat with pointed ears.}
    \end{subfigure}\\
    \begin{subfigure}[b]{0.48\linewidth}
        \centering
        \includegraphics[width=\linewidth]{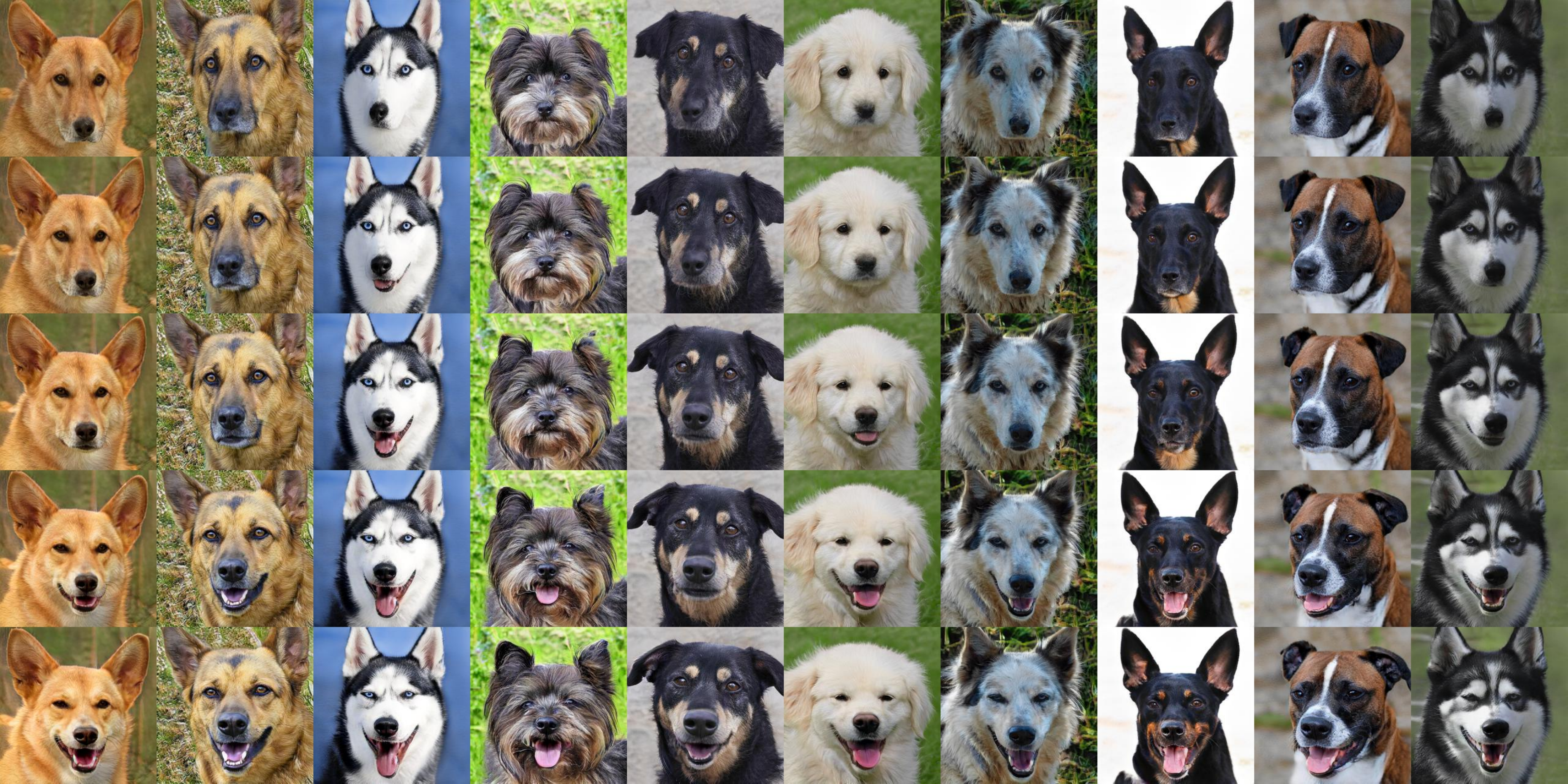}
        \caption{Attribute editing: a dog with open mouth.}
    \end{subfigure}
    \begin{subfigure}[b]{0.48\linewidth}
        \centering
        \includegraphics[width=\linewidth]{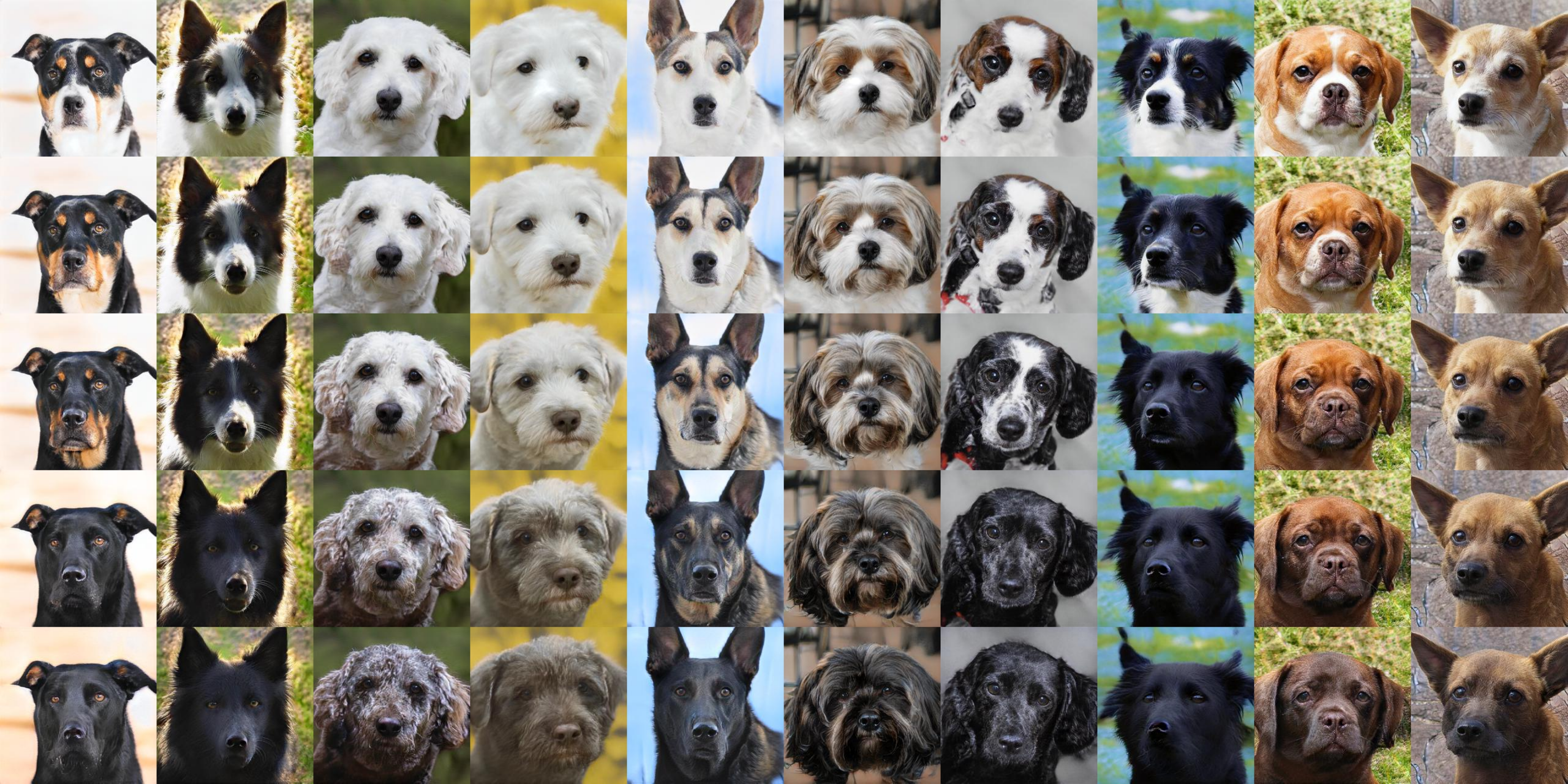}
        \caption{Attribute editing: a black dog.}
    \end{subfigure}
    
    \caption{Visualization of global edit directions by utilizing the StyleCLIP channel relevance matrix. Images are sampled from the AFHQ domain using StyleGAN2-ADA. Every column demonstrates an edited image from edit weight $w = -30$ to $w = 30$. Weights of five images are linearly interpolated as $\{-30, -15, 0, 15, 30\}$. We can see that global edit directions are generalizable on multiple images.}
    \label{fig:appendix_attribute_editing_afhq}
\end{figure*}

\textbf{Single-attribute editing.} Fig.~\ref{fig:appendix_attribute_editing_afhq} and Fig.~\ref{fig:appendix_attribute_editing_ffhq} show a set of images of different object categories by editing different attributes extracted with the global edit directions method (as described in Section 3.2 of the main text). By analyzing the user's input attribute string, we can see that the modified image only alters in the attribute direction while maintaining the other attributes. 

\begin{figure*}[h]
    \centering
    \begin{subfigure}[b]{\linewidth}
        \includegraphics[width=0.49\linewidth]{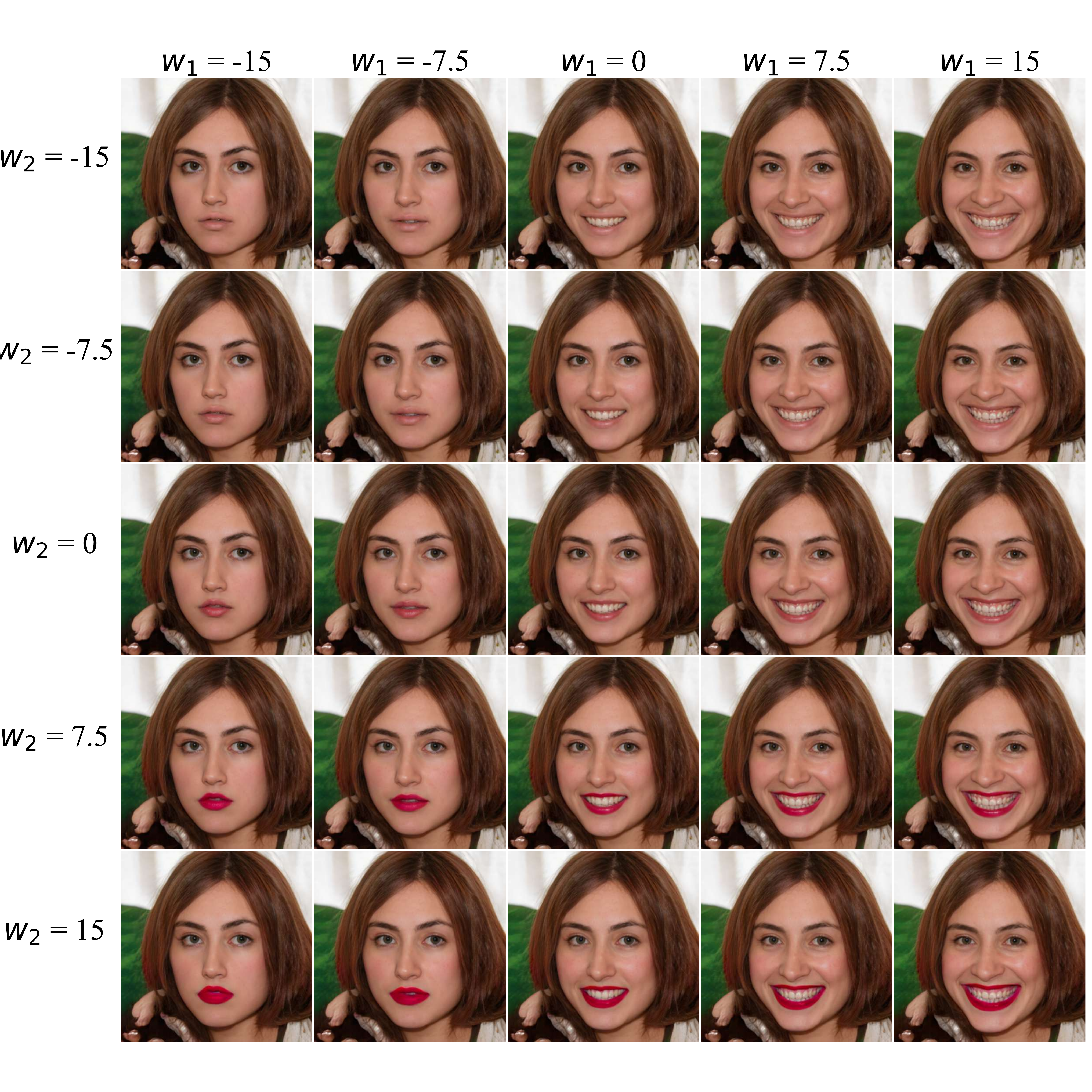}
        \includegraphics[width=0.49\linewidth]{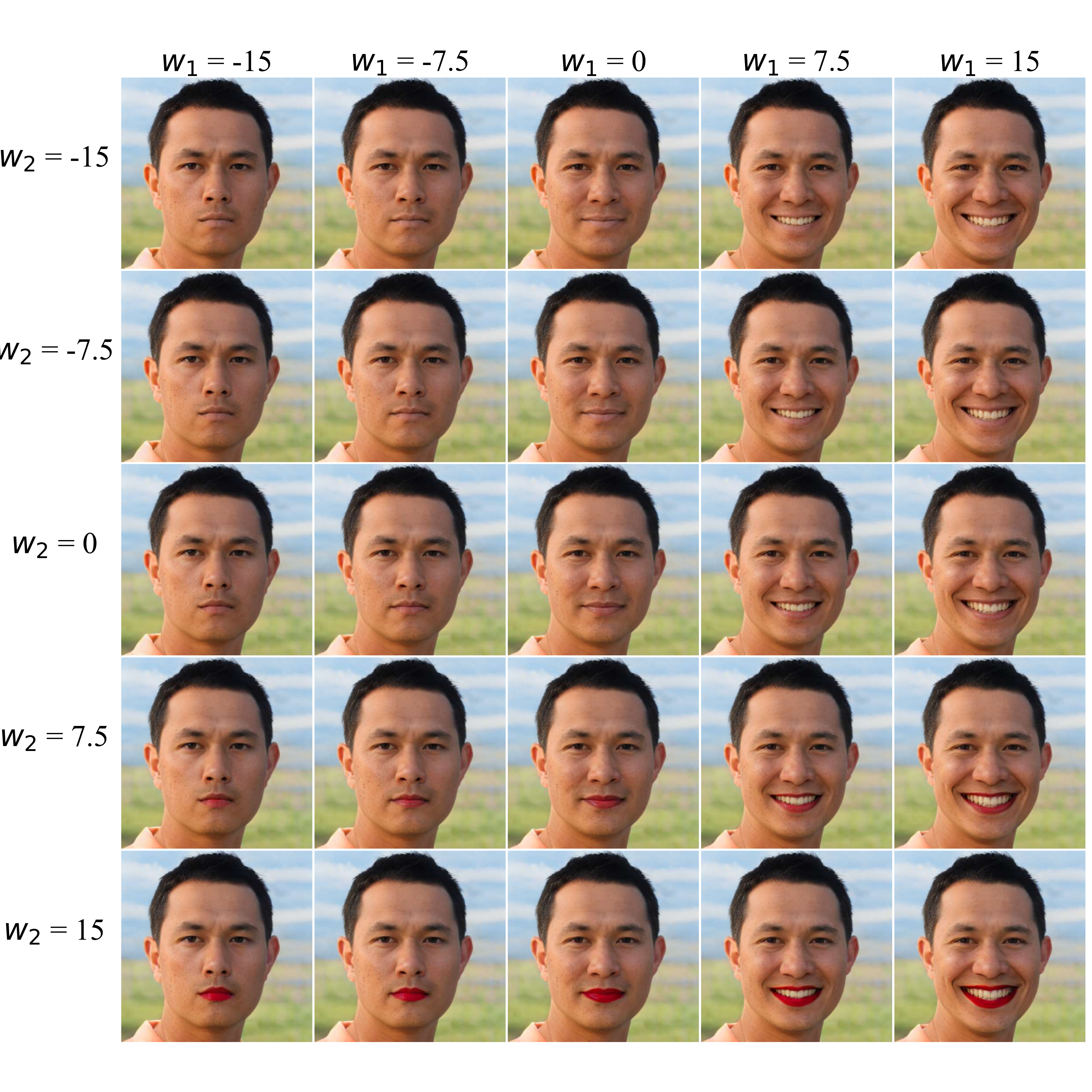}
        \caption{Combination of smiling ($w_1$) and lipstick ($w_2$).}
        \label{fig:interpolation_similing_lipstick}
    \end{subfigure}
    \begin{subfigure}[b]{\linewidth}
        \includegraphics[width=0.49\linewidth]{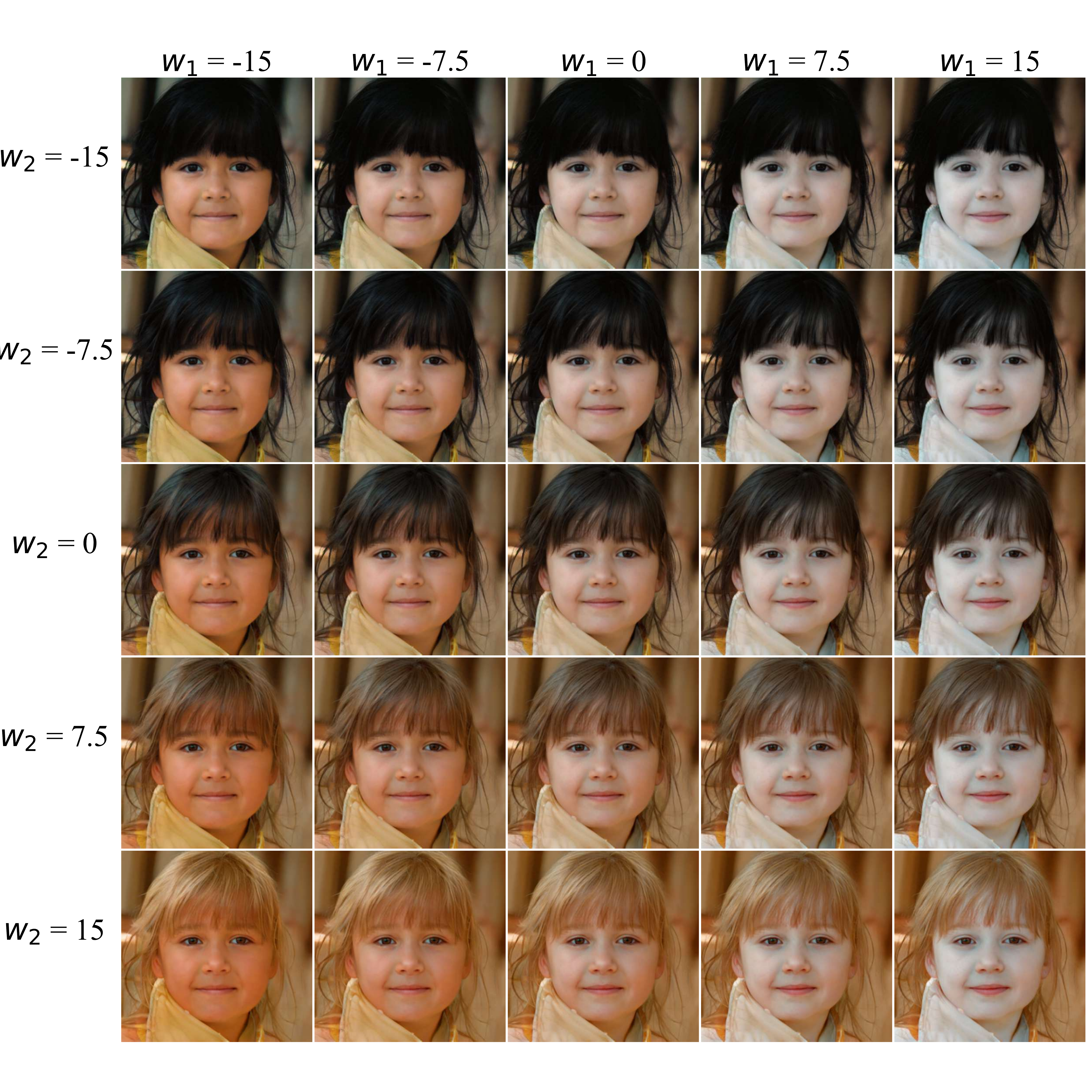}
        \includegraphics[width=0.49\linewidth]{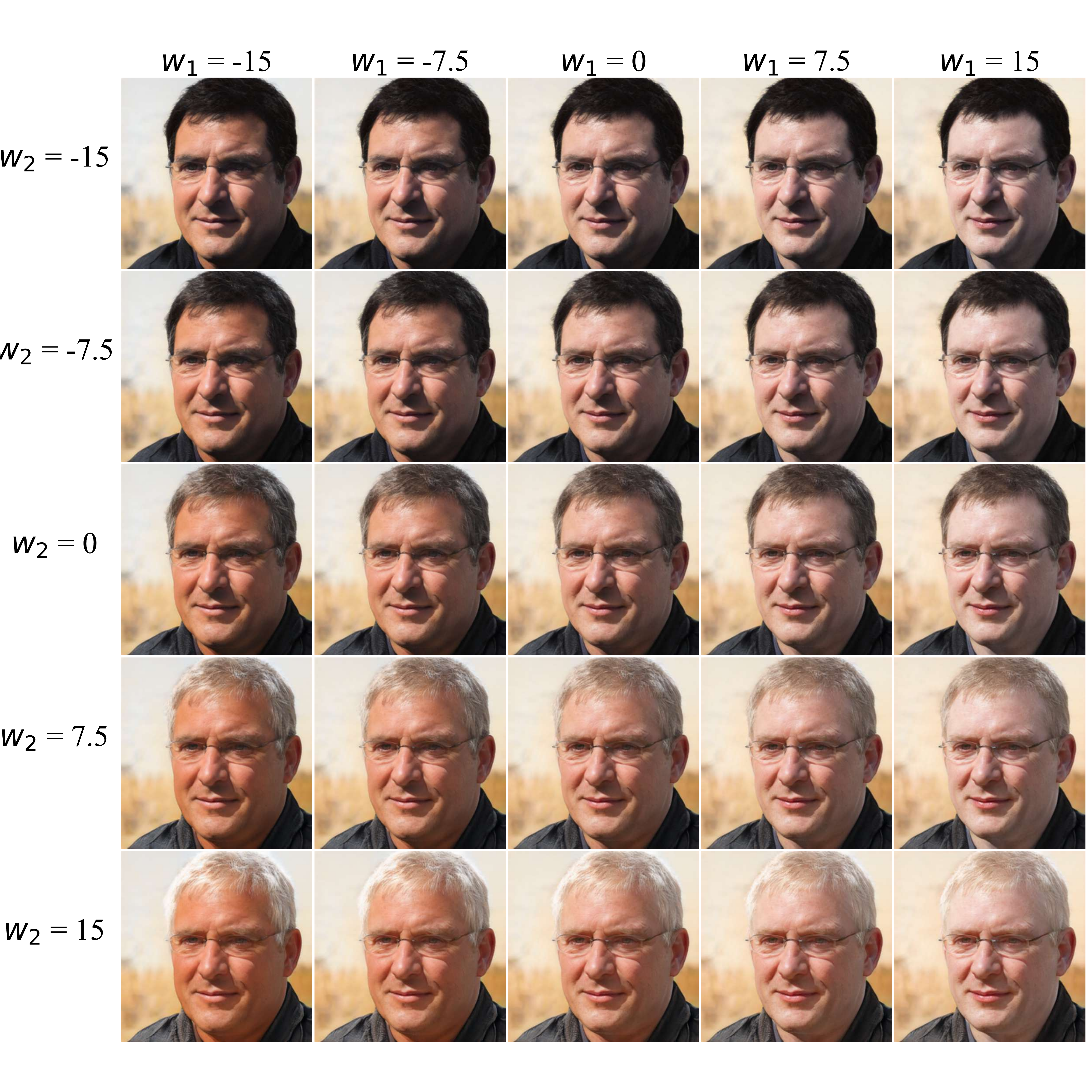}
        \caption{Combination of pale skin ($w_1$) and blond hair ($w_2$).}
        \label{fig:interpolation_paleskin_blondhair}
    \end{subfigure}
    \caption{Visualization of traversing on directional (attribute) style vectors to validate the effectiveness of multiple attribute editing.}
    \label{fig:appendix_interpolation}
\end{figure*}

\textbf{Multiple-attribute editing.} 
We demonstrate the validity of our method for editing multiple attributes through linear combination (as outlined in Equation 3 of the main text) by presenting illustrations of combined edits in Figure ~\ref{fig:appendix_interpolation}.

\subsection{User study for edited images}
To validate that our counterfactual image synthesis preserve fine-grained details and authenticity, we conducted a user study validating two aspects: synthesis fidelity and attribute consistency. 

\begin{figure*}[h]
    \centering
    \begin{subfigure}[t]{0.37\linewidth}
        \centering
        \captionsetup{margin={4pt,4pt}}
        \includegraphics[width=\linewidth]{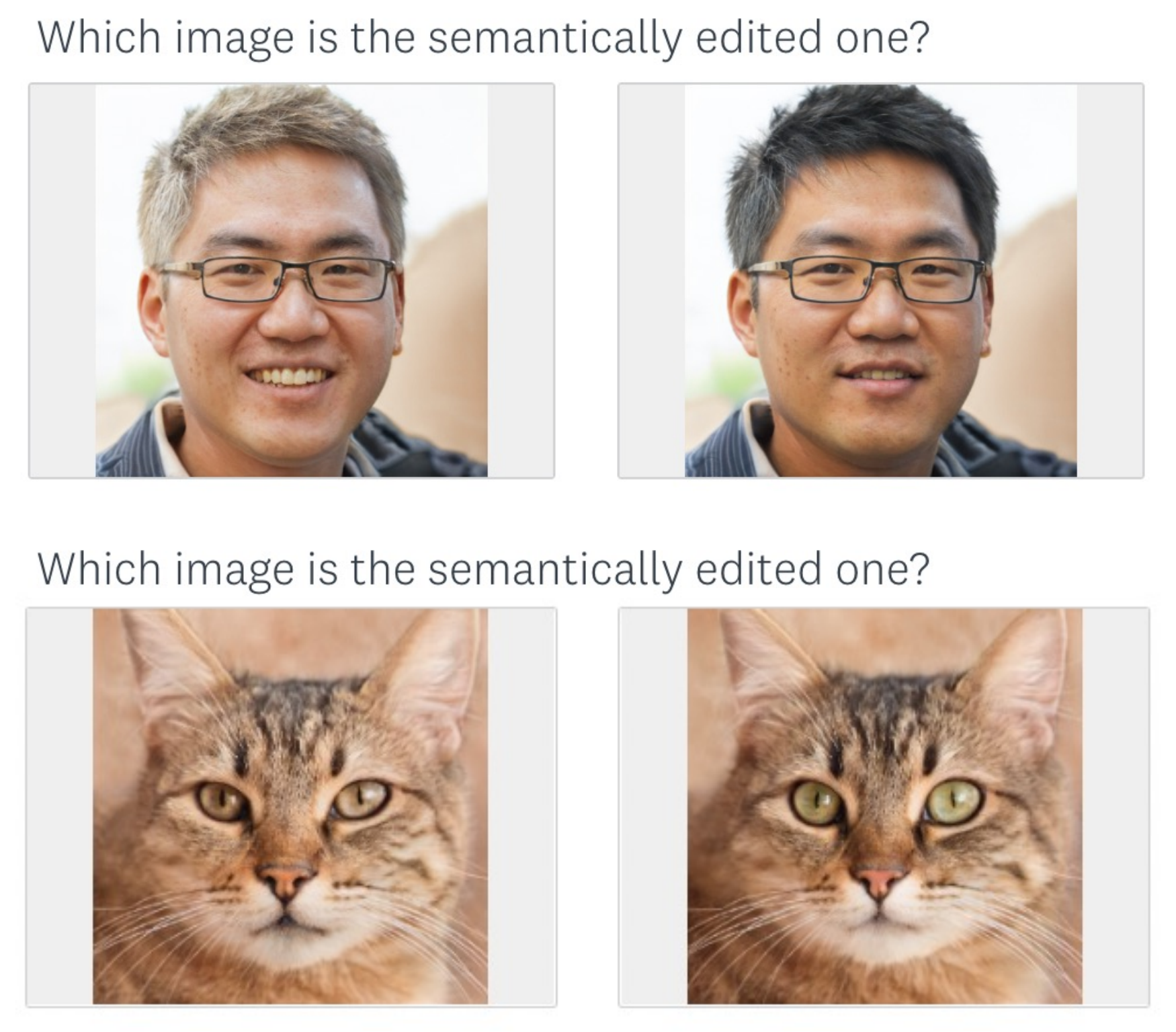}
        \caption{Evaluating visual fidelity. We show two images and let users choose the one that they think is edited. The counterfactuals are generated on Eyeglasses classifier and Cat/Dog classifier. }
        \label{fig:user_study_visual_fidelity}
    \end{subfigure}
    \begin{subfigure}[t]{0.32\linewidth}
        \centering
        \captionsetup{margin={4pt,4pt}}
        \includegraphics[width=\linewidth]{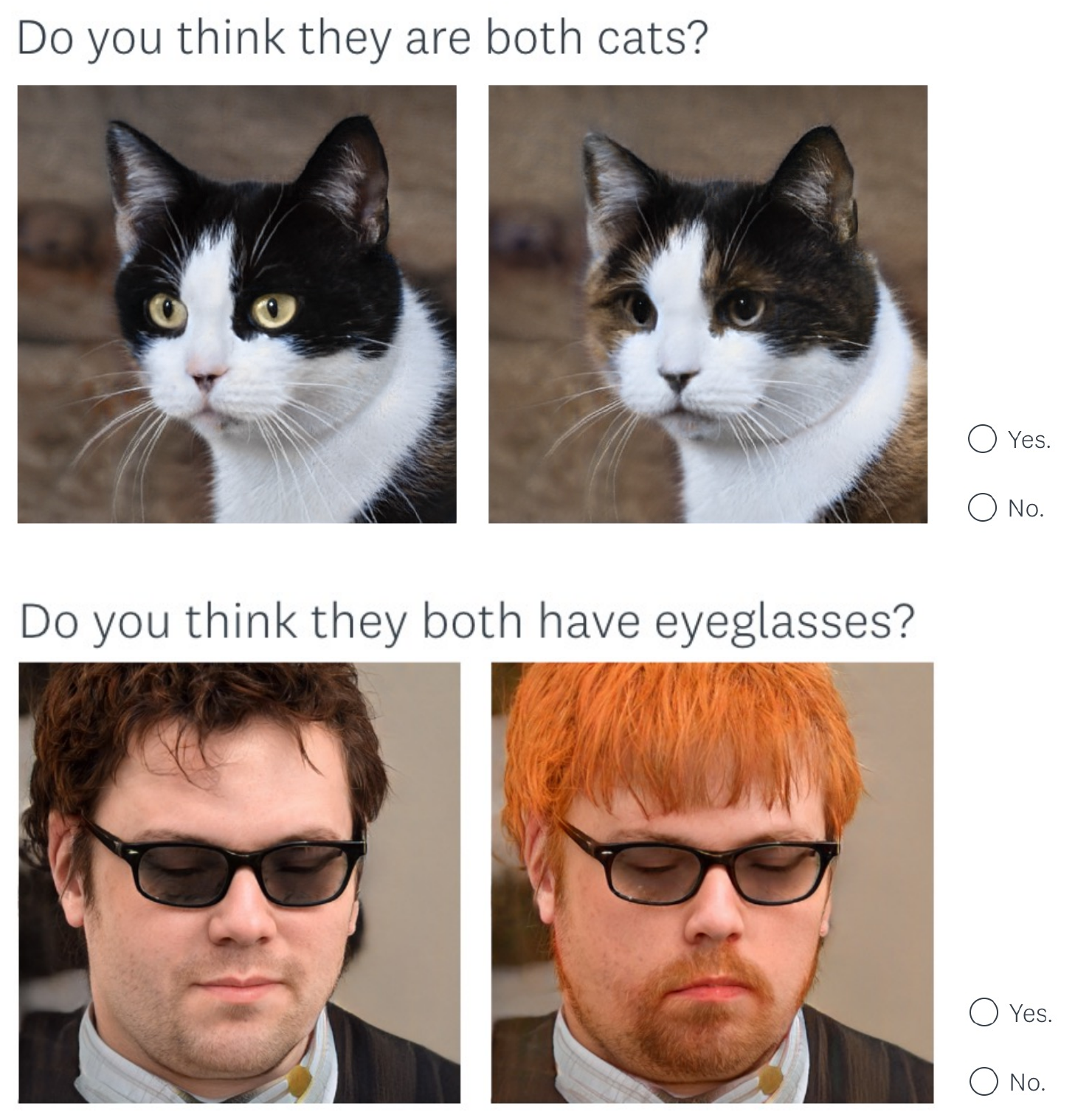}
        \caption{Evaluating attribute consistency. The user classifies whether the ground truth is flipped. Example of counterfactual images on Cat/Dog classifier and Eyeglasses classifier is shown above.}
        \label{fig:user_study_attribute_consistency}
    \end{subfigure}
\caption{Sample questions in the user study. Each user answers $10$ questions for each of the two user studies. }
\end{figure*}

\textbf{User study for synthesis fidelity.} 
The classification of the counterfactual synthesis image vs real images by the user is employed to confirm that no unrealistic artifacts are introduced throughout the process of our model Fig.~\ref{fig:user_study_visual_fidelity} shows sample questions of this study.  In theory, the worst-case scenario is that users can accurately identify the semantic modification and achieve a user recognition rate of $100\%$. Conversely, the best-case scenario would be that users are unable to identify any counterfactual synthesis and make random guesses, resulting in a user recognition rate of $50\%$.

\textbf{User study for attribute consistency.} We ask users whether they agree that the counterfactual and original images are consistent on the ground truth w.r.t. the target classifier. For example, during the counterfactual synthesis for the cat/dog classifier, a counterfactual cat image should stay consistent as a cat. Fig.~\ref{fig:user_study_attribute_consistency} shows another sample questions. The worst case is that the counterfactual changes the ground truth label to affect the target model, which makes the user agreement rate very low (even to zero). 

The user study statistics are presented in Table ~\ref{tab:user_study}. The study involved $34$ participants with at least an undergraduate level of education, who were divided into two groups using separate collector links. The participants themselves randomly selected their group (i.e., the link clicked), and their responses were collected.

\begin{figure*}[t]
    \centering
    \includegraphics[width=0.49\linewidth]{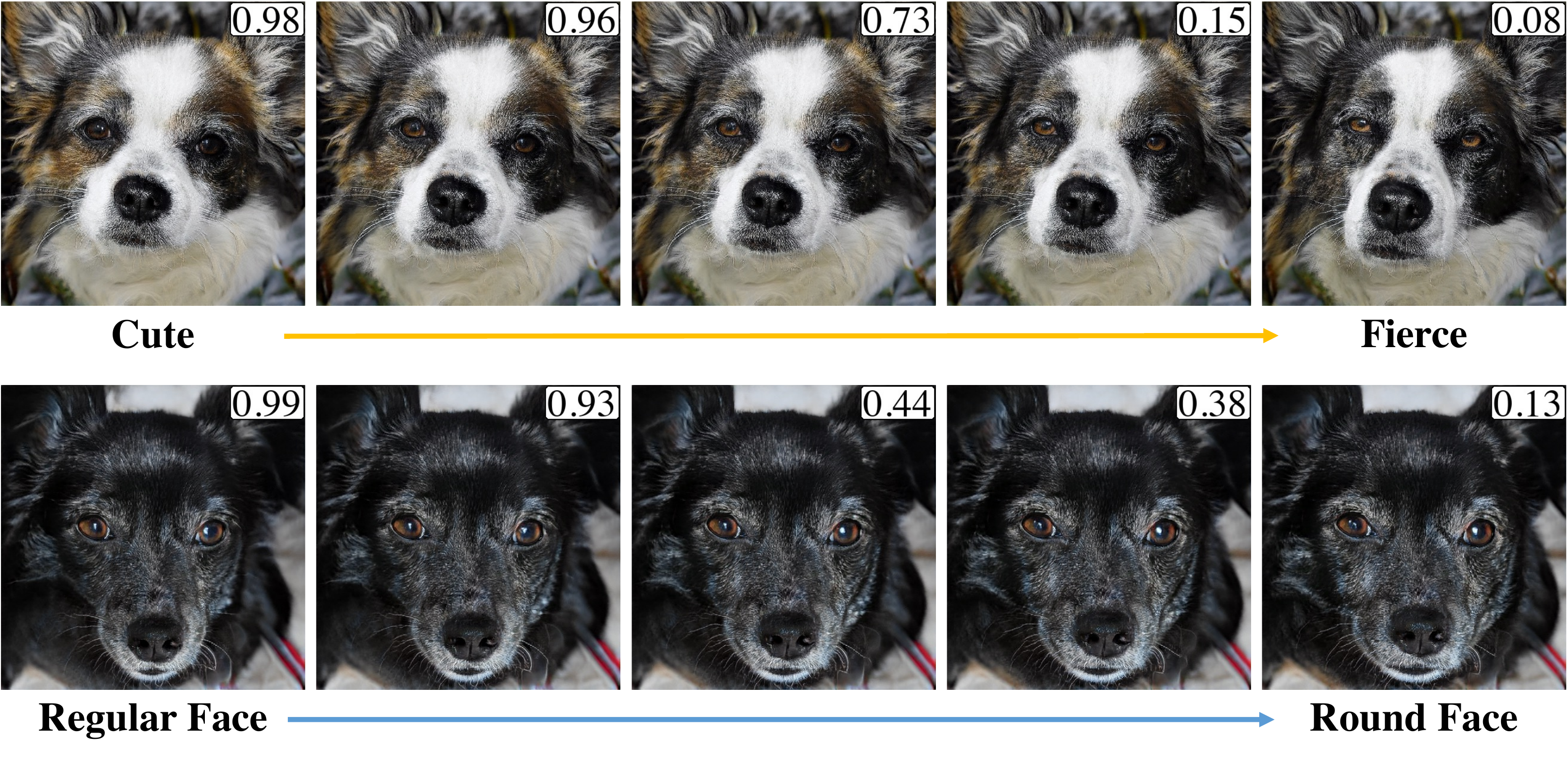}
    \includegraphics[width=0.49\linewidth]{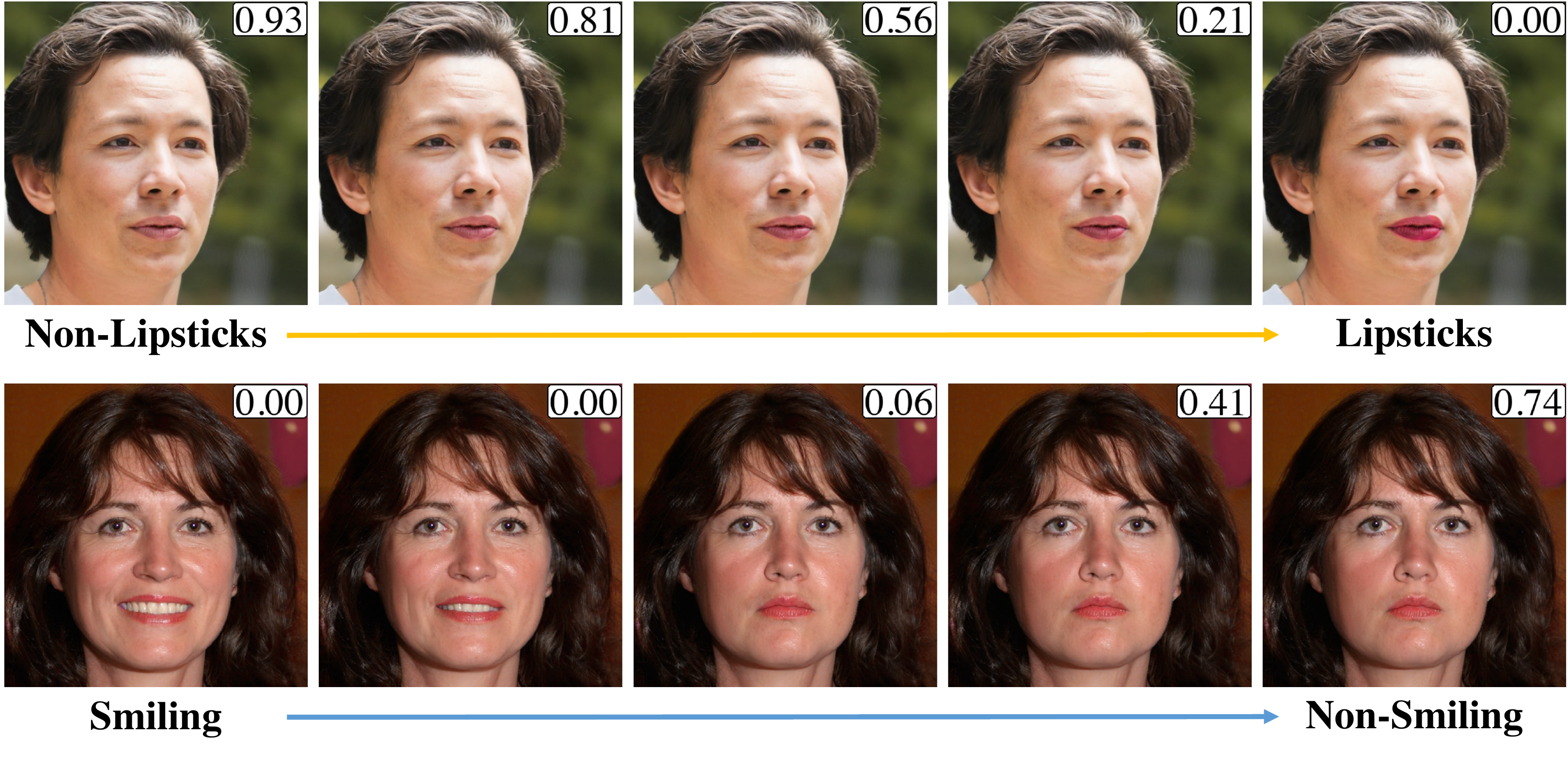}
    \vspace{-3mm}
    \caption{Effect of progressively generating counterfactual images on the Cat/Dog classifier (0-Cat / 1-Dog), and the Perceived Gender classifier (0-Female / 1-Male). Model probability prediction during the process is attached at the top right corner.}
    \label{fig:appendix_classifier_single}
\end{figure*}

The production of high-quality counterfactual images is supported by the difficulty users had in differentiating them. Additionally, the majority of users concurred that the counterfactual images do not change the ground truth concerning the target classifier, confirming that our methodology generates meaningful counterfactuals. However, it should be noted that due to the nature of our recognition system, human volunteers are somewhat more responsive to human faces. As a result, we observed a slightly higher recognition rate in the human face (FFHQ) domain than in the animal face (AFHQ) domain.

\subsection{Stability across CLIP phrasing/wording:} 

\begin{figure*}[t]
  \centering
  % \fbox{\rule{0pt}{0.5in} \rule{0.9\linewidth}{0pt}}
  \includegraphics[width=\linewidth]{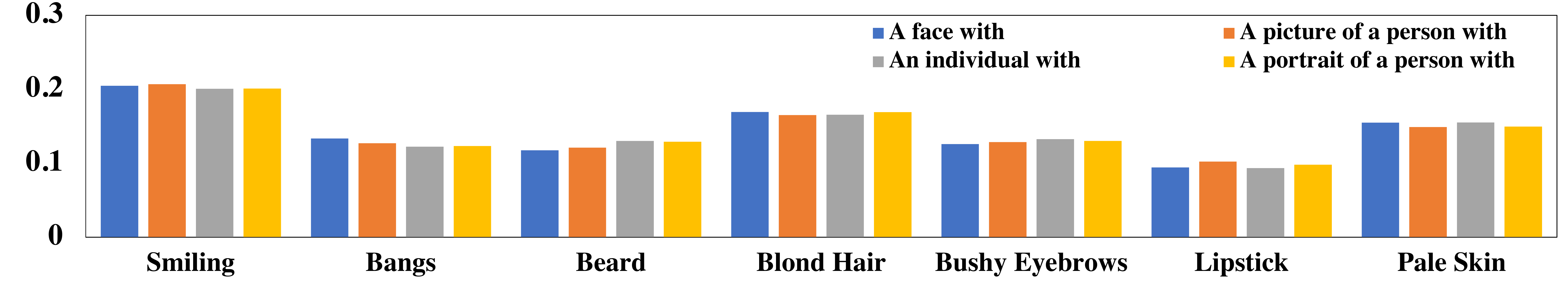}
  \vspace{-8mm}
   \caption{ Sensitivity histograms when using four instances of phrases with a similar concept. Zoom in for better visibility.}
   \label{fig:stability_histogram}
   \vspace{-4mm}
\end{figure*}

It is worth noting that the resulting counterfactual image is affected by the wording of the prompt used. In our framework, we subtract the neutral phrase (such as "a face") after encoding in CLIP space to ensure that the attribute edit direction is unambiguous enough. Our experimentation has shown that as long as the prompt accurately describes the object, comparable outcomes can be achieved. For instance, we obtained similar sensitivity results on the perceived-age classifier using prompts like "a picture of a person with X," "a portrait of a person with X," or other synonyms. Examples of this are presented in Figure ~\ref{fig:stability_histogram}.

\begin{figure*}[t]
  \centering
  % \fbox{\rule{0pt}{0.5in} \rule{0.9\linewidth}{0pt}}
  \includegraphics[width=0.9\linewidth]{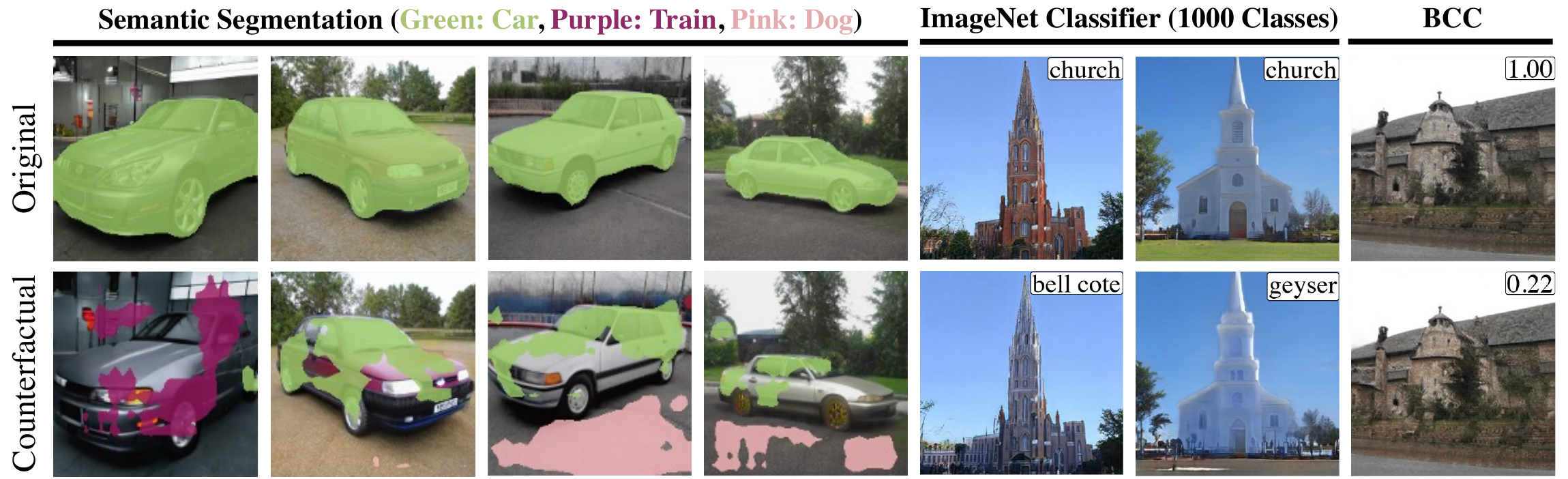}
   \caption{\ourmodel counterfactuals on more tasks (segmentation, multi-class classifier) and additional visual domains (cars, churches). Zoom in for better visibility.}
   \label{fig:more_models}
   \vspace{-3mm}
\end{figure*}

\begin{table}[h]
   \centering
   \small	
   \begin{tabular}{@{}lccc@{}}
     \toprule
    \multicolumn{1}{c}{Name of Study} & Domain & Group 1    & Group 2\\
     \midrule
    \multicolumn{1}{c}{\multirow{2}{*}{\makecell{Synthesis Fidelity (\\Recognition Rate $\downarrow$, \%)}}} & FFHQ &62.12 & 71.79 \\
                                                                & AFHQ & 51.30 & 50.55  \\
                        \midrule
     \multicolumn{1}{c}{\multirow{2}{*}{\makecell{Attribute Consistency (\\Agreement Rate $\uparrow$, \%)}}} & FFHQ &94.12 & 90.76  \\
                                                          & AFHQ & 89.92 & 88.26 \\

     \bottomrule
   \end{tabular}
   %\end{adjustbox}
   \caption{User study results. We can see from the table that our counterfactual synthesis preserves the visual quality and maintains the ground truth labels from the user's perspective.}
   \label{tab:user_study}
   \vspace{-3mm}
\end{table}
% Appendix for Histograms and Counterfactual demo
\section{Additional Results of Model Diagnosis}

\subsection{Additional counterfactual images}
\begin{figure*}[!t]
\vspace{-3mm}
    \centering
    \begin{subfigure}[b]{\linewidth}
    \includegraphics[width=\linewidth]{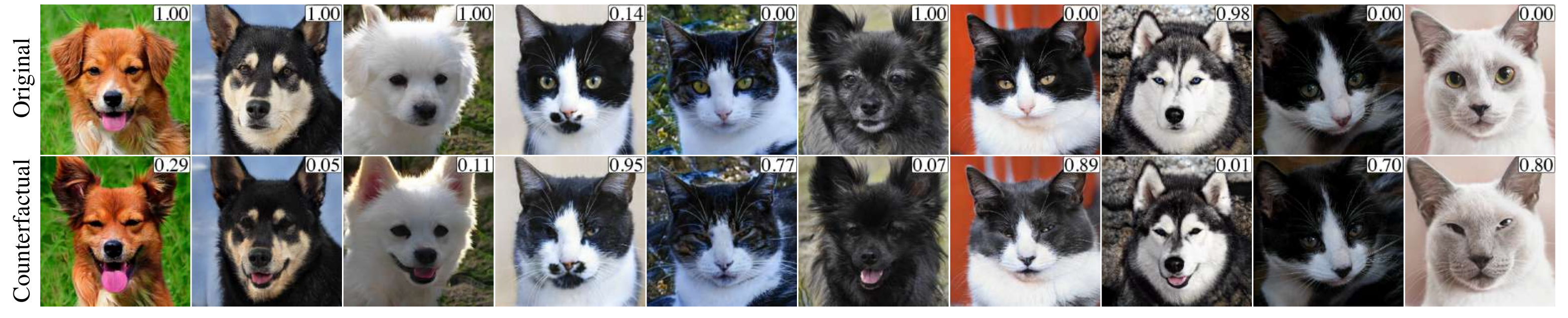}
    \caption{Multiple-attribute counterfactual for cat/dog classifier.}
    \end{subfigure}\\
    \begin{subfigure}[b]{\linewidth}
    \includegraphics[width=\linewidth]{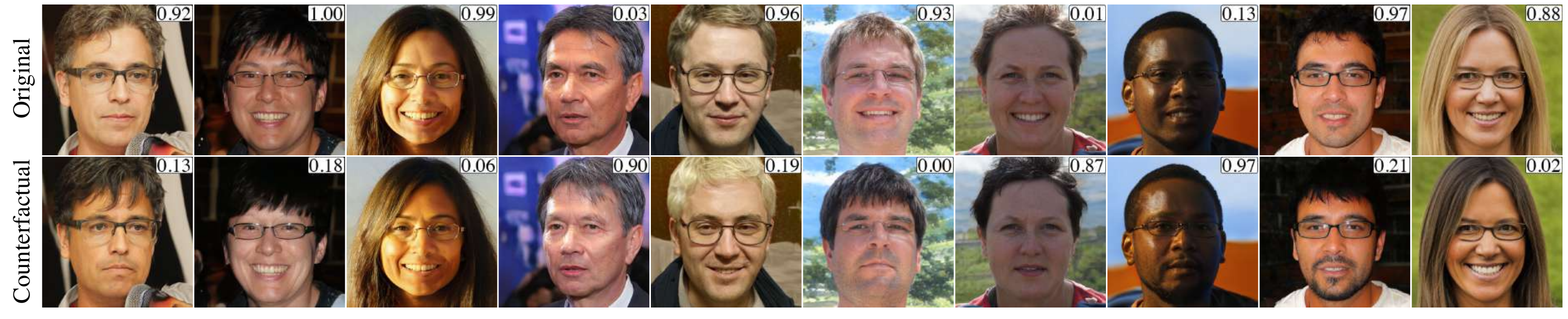}
    \caption{Multiple-attribute counterfactual for eyeglasses classifier.}
    \end{subfigure}\\
    \begin{subfigure}[b]{\linewidth}
    \includegraphics[width=\linewidth]{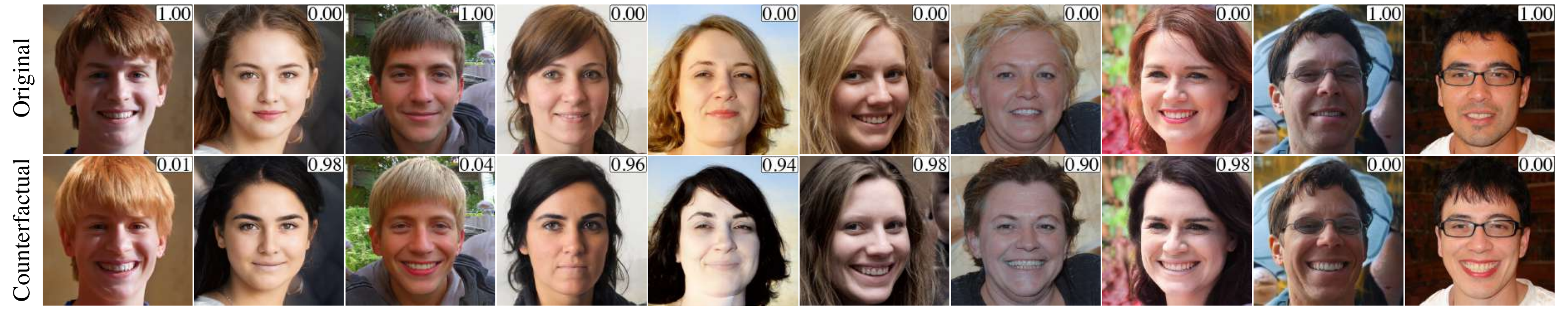}
    \caption{Multiple-attribute counterfactual for perceived gender classifier.}
    \end{subfigure}\\
    \begin{subfigure}[b]{\linewidth}
    \includegraphics[width=\linewidth]{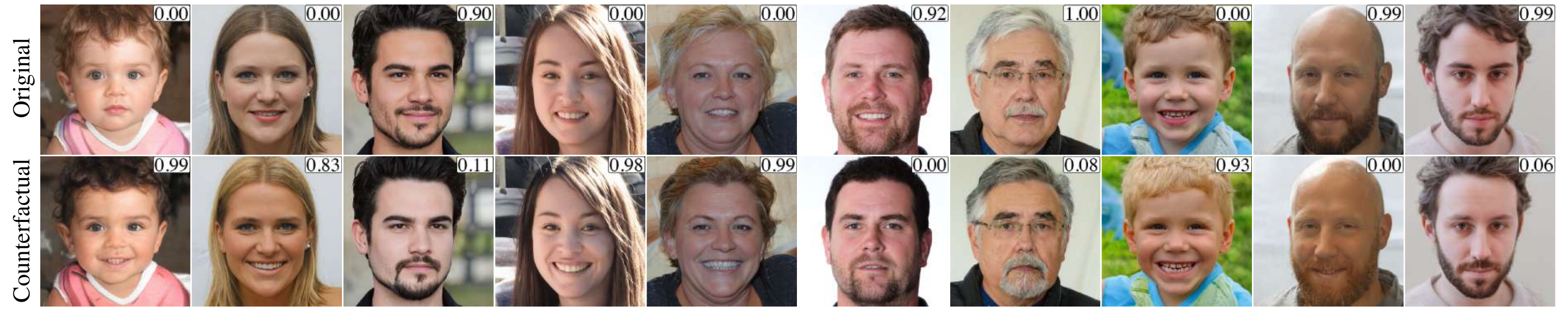}
    \caption{Multiple-attribute counterfactual for mustache classifier.}
    \end{subfigure}\\
    \begin{subfigure}[b]{\linewidth}
    \includegraphics[width=\linewidth]{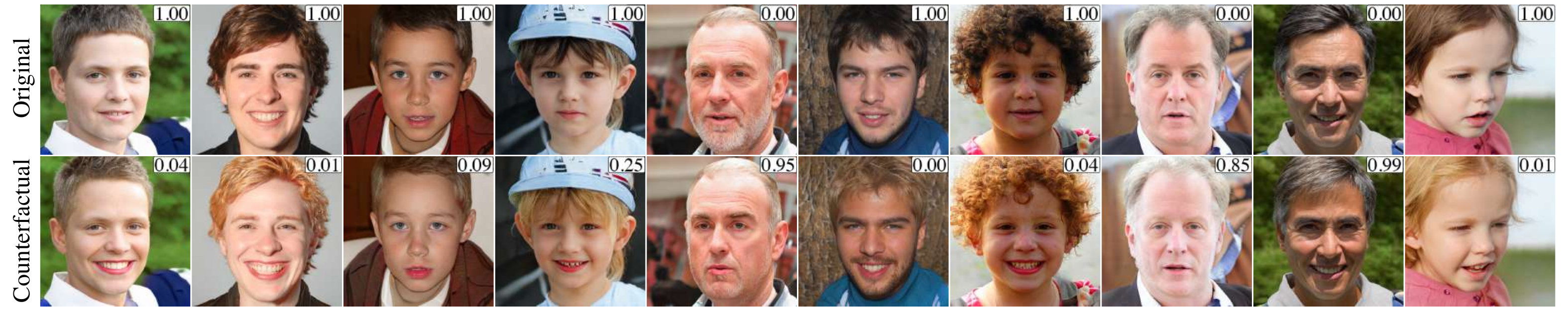}
    \caption{Multiple-attribute counterfactual for perceived age classifier.}
    \end{subfigure}\\
    \caption{Multi-attribute counterfactual in the human face and animal face domain. The right-up corner of each image records the model output prediction.}
    \label{fig:appendix_classifier_multi}
\end{figure*}

Fig.~\ref{fig:appendix_classifier_single} shows more examples of single-attribute counterfactual images on the Cat/Dog and Perceived Gender classifiers. The output prediction is shown in the top-right corner. It shows that the model prediction is flipped without changing the actual target attribute. In addition to binary classification and key-point detection in our manuscript, we further illustrate the extension of \ourmodel counterfactuals on semantic segmentation, multi-class classification, and binary church classifier (BCC) in Fig.~\ref{fig:more_models}. Fig.~\ref{fig:appendix_classifier_multi} shows more examples of multiple-attribute counterfactual images.

\subsection{Additional histograms}
\begin{figure*}[t]
    \centering
    \includegraphics[width=0.33\linewidth]{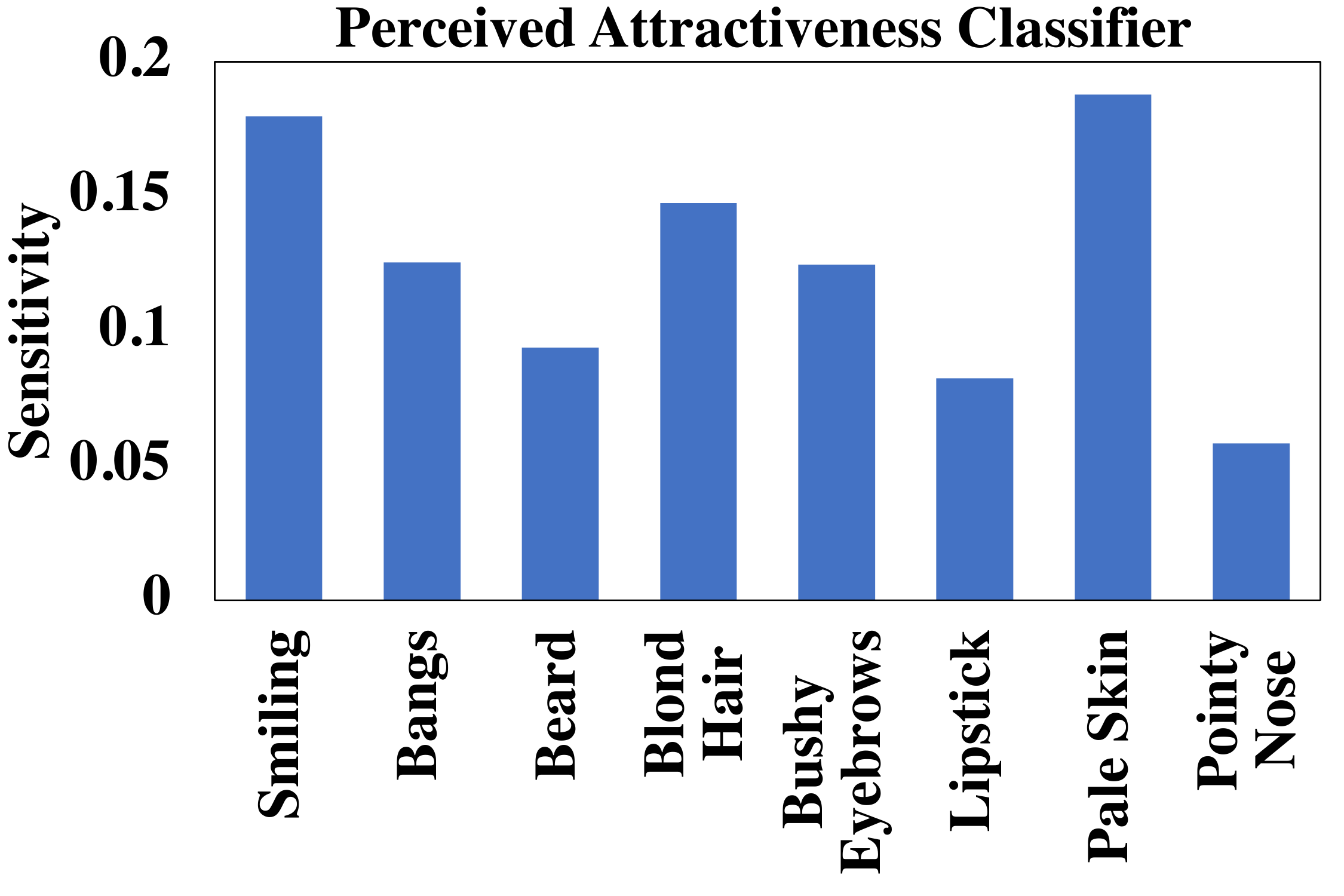}
    \includegraphics[width=0.33\linewidth]{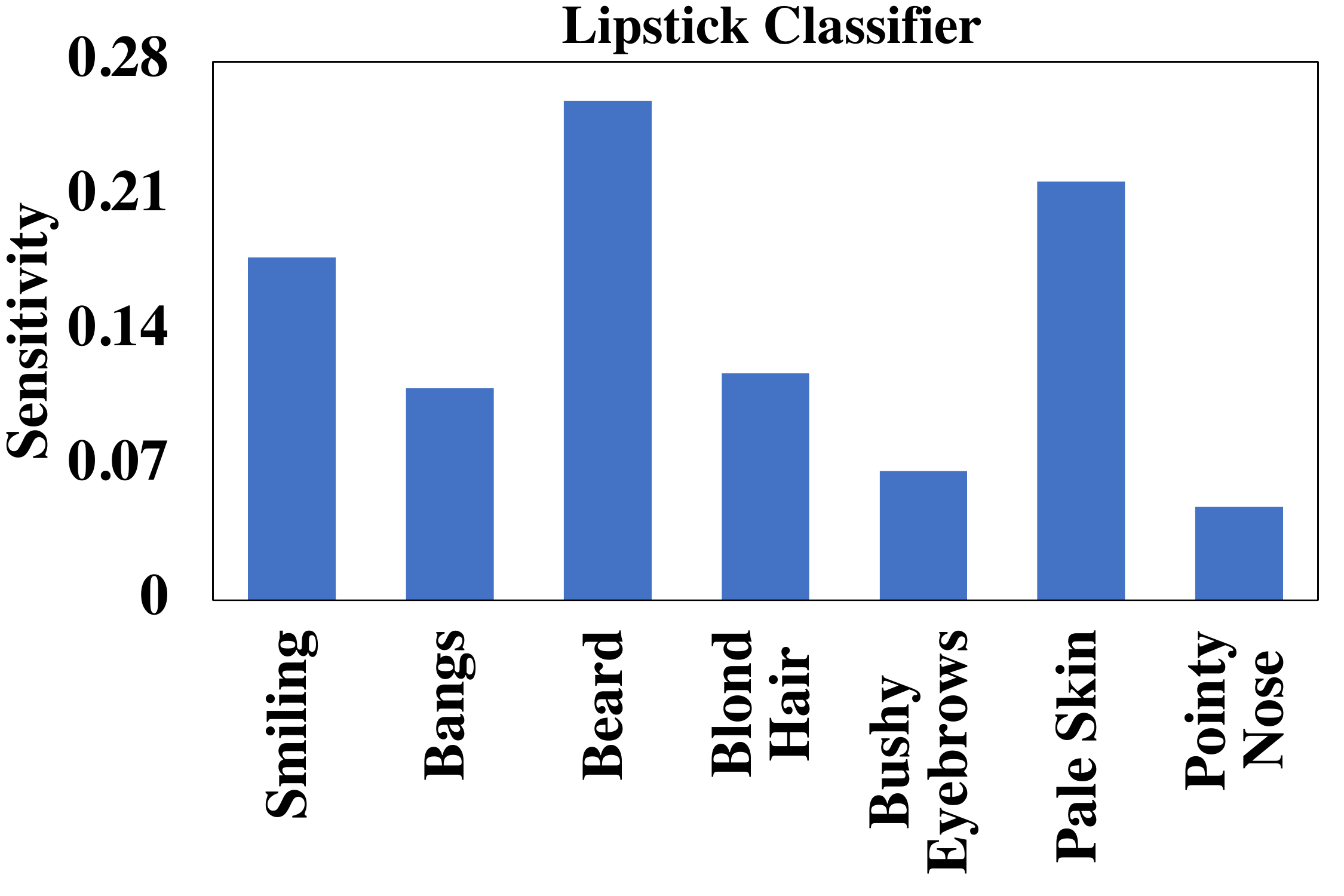}
    \includegraphics[width=0.33\linewidth]{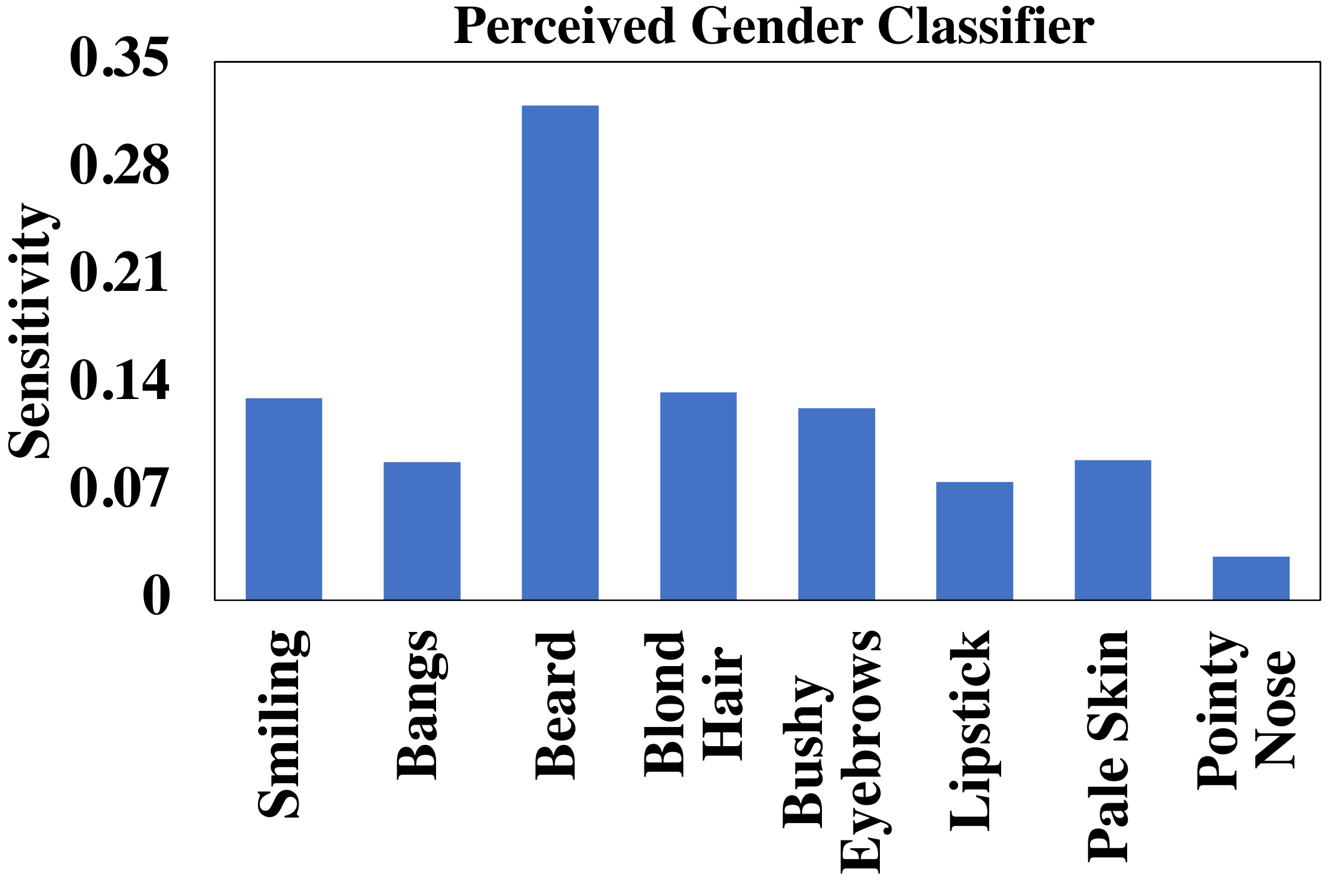}\\
    \includegraphics[width=0.33\linewidth]{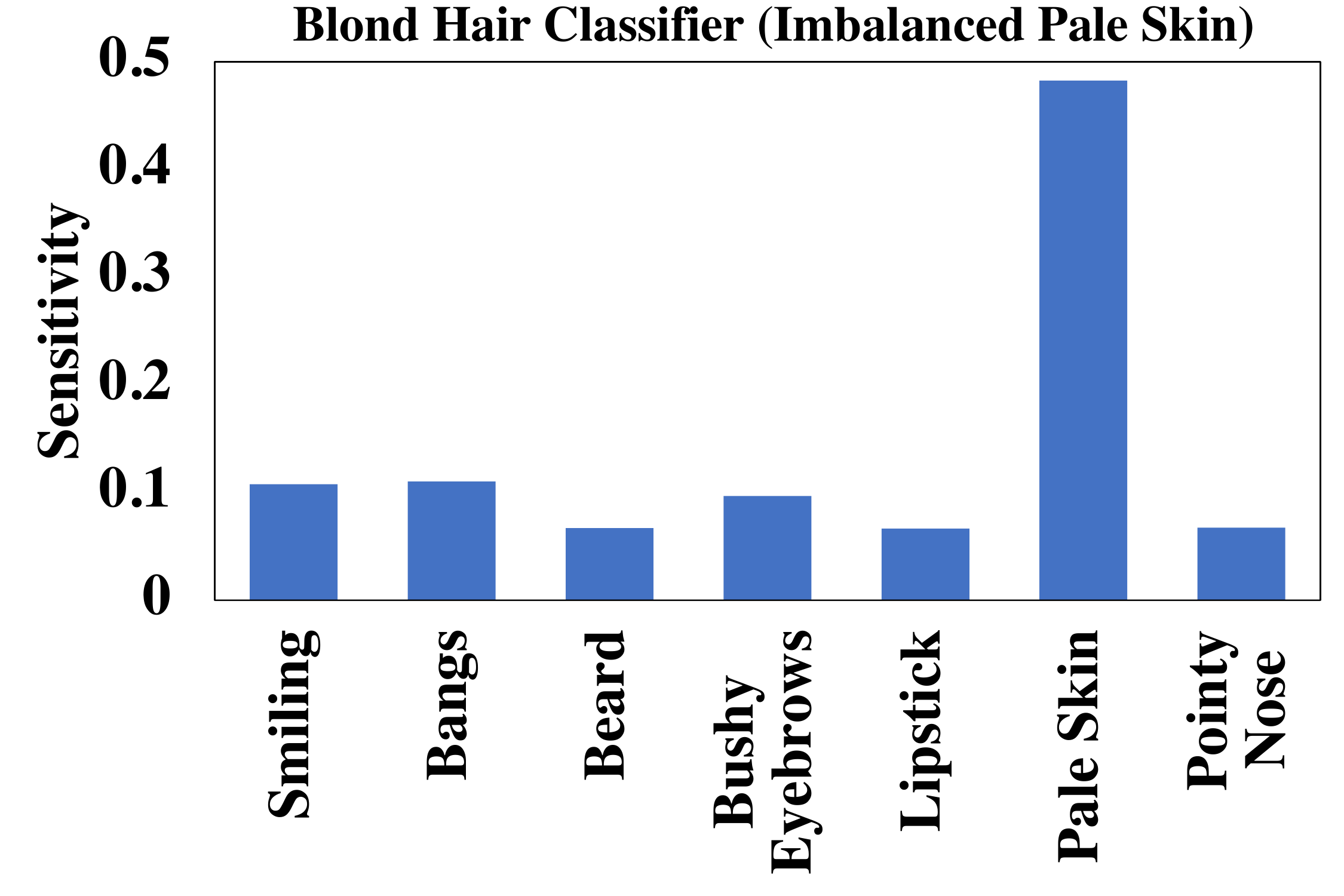}
    \includegraphics[width=0.33\linewidth]{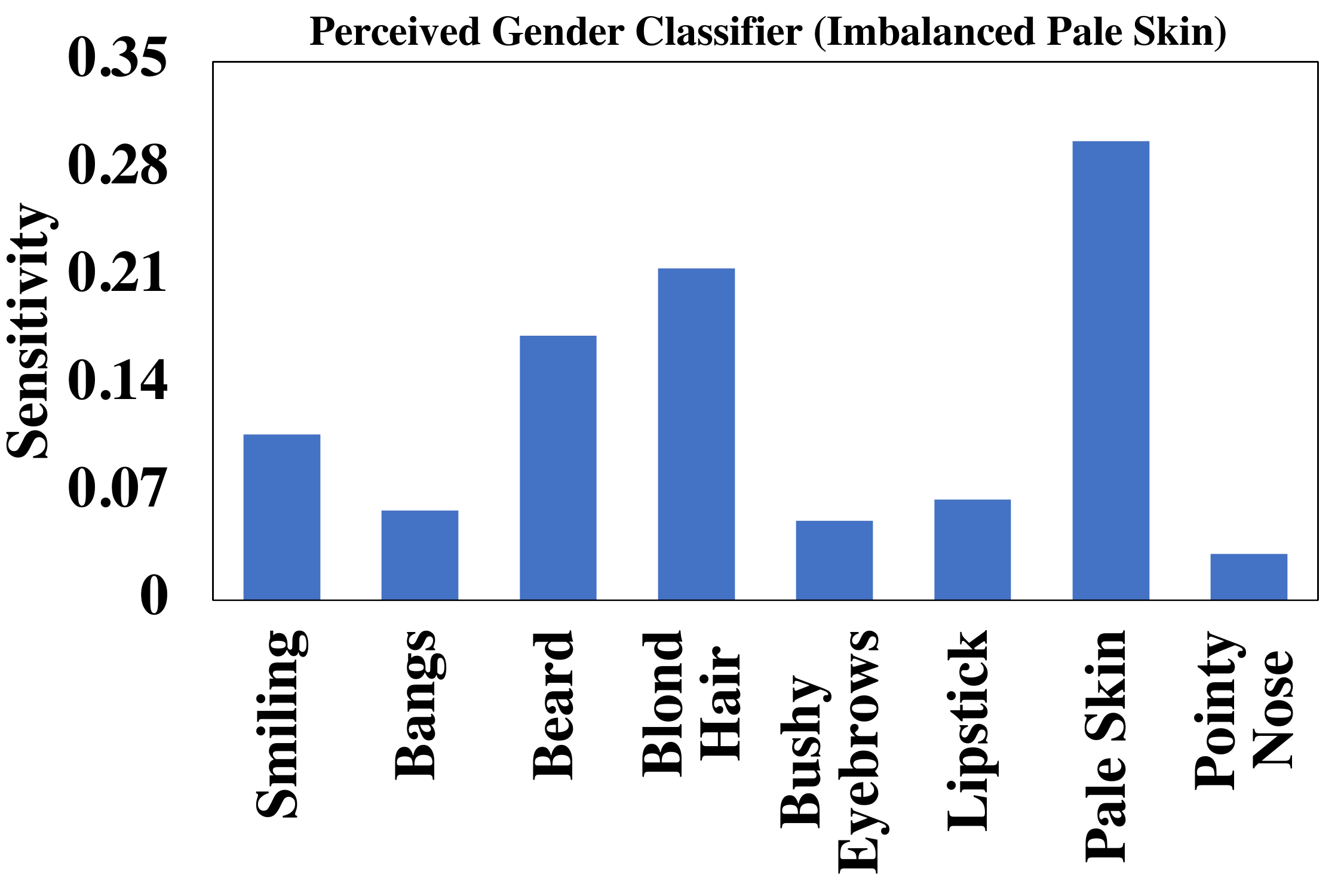}
    \includegraphics[width=0.33\linewidth]{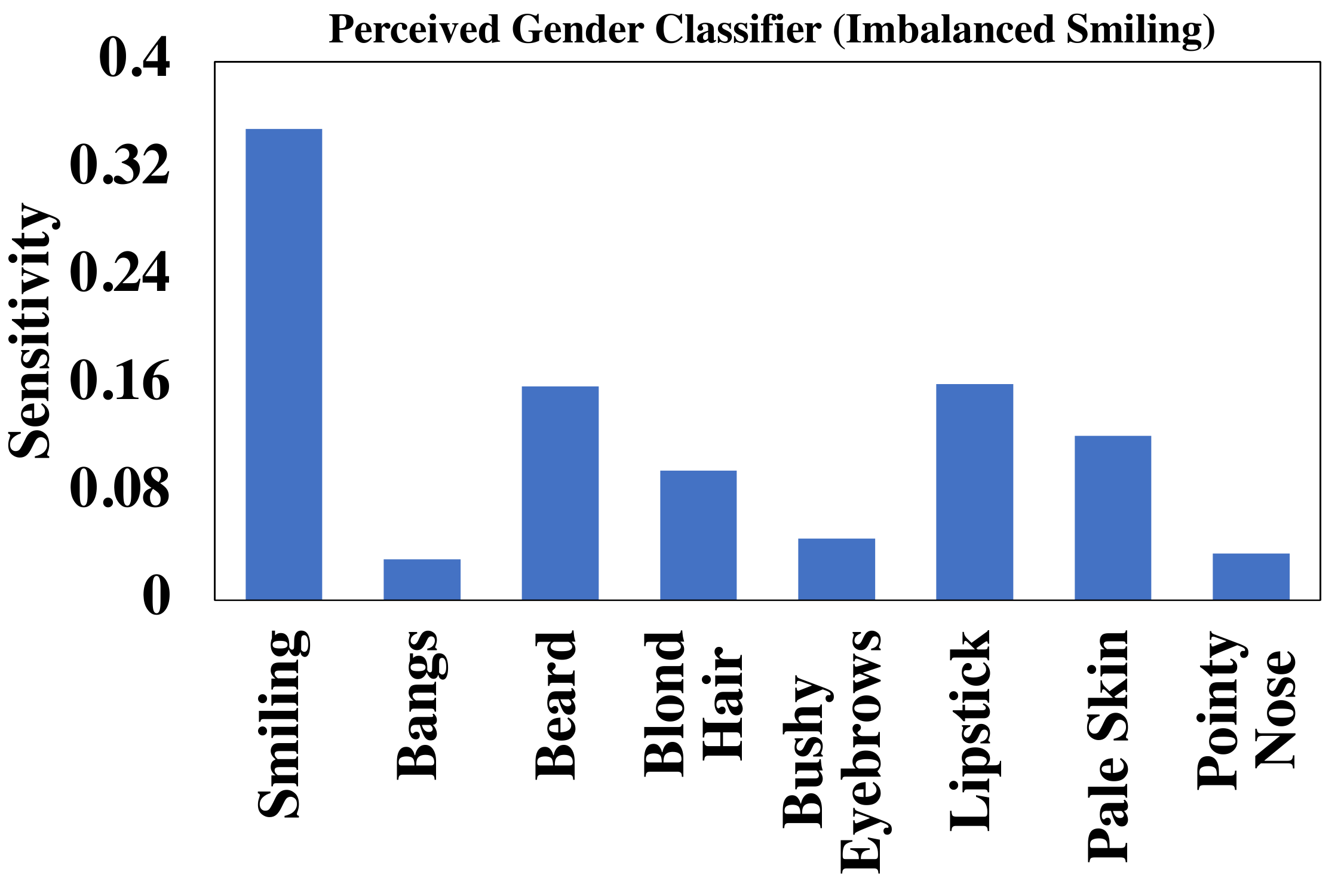}
    \vspace{-5mm}
    \caption{The above histograms show \ourmodel on three regularly trained classifiers on CelebA, and the bottom histograms show \ourmodel successfully detects the bias in the manually-crafted imbalanced classifiers.}
    \label{fig:apendix_histograms}
    \vspace{-1mm}
\end{figure*}

Fig.~\ref{fig:apendix_histograms} shows more histograms on the classifiers trained on CelebA (top) and the classifiers that are intentionally biased  (bottom). The models and datasets are created using the same method described in Section 4 of the main text.
% Details of using other optimization appraoches (e.g., linear approximation
\section{Ablation of the Adversarial Optimization Method}
When there are multiple attributes (i.e., $N>1$) to optimize, linearizing the cost function as grid in high dimensional space will help to efficiently approximate convergence in limited epochs. Specifically, we have the option to adopt PGD \cite{madry2018towards} (i.e., update using $\eta \cdot\operatorname{sign} (\nabla_{\mathbf{w}} \mathcal{L})$) for efficient optimization. We compared generating counterfactuals with and without  projected gradients. Table~\ref{tab:appendix_pgd} shows the visual quality and flip rate of the generated counterfactuals. We can observe that \ourmodel-PGD image quality is finer under Structured Similarity Indexing Method (SSIM) \cite{SSIM}, while \ourmodel-SGD has a higher flip rate. The images from \ourmodel-PGD is finer since the signed method stabilizes the optimization by eliminating problems of gradient vanishing and exploding. 

 \begin{table}[h]
   \small
   \centering
   \begin{tabular}{@{}llcc@{}}
     \toprule
    \multicolumn{1}{c}{Optimization} & Classifier &  SSIM ($\uparrow$) & Flip Rate (\%, $\uparrow$)\\
     \midrule
     \multicolumn{1}{c}{\multirow{3}{*}{SGD}} & Perceived Age & 0.5732 & 67.24  \\
                                              & Perceived Gender & 0.5815 & 49.40 \\
                                              & Mustache & 0.5971 & 36.33 \\
                        \midrule
     \multicolumn{1}{c}{\multirow{3}{*}{PGD}} & Perceived Age & 0.8065 & 50.19\\
                                              & Perceived Gender & 0.7035 & 42.84 \\
                                              & Mustache & 0.7613 & 25.10 \\
     \bottomrule
   \end{tabular}
   %\end{adjustbox}
   \caption{The comparison of counterfactuals generated with stochastic gradient descent (SGD) and projected gradient descent (PGD) method. We can observe that \ourmodel-PGD image quality is finer under SSIM (Structured Similarity Indexing Method) \cite{SSIM} metrics, while \ourmodel-SGD has a higher flip rate. }
   \label{tab:appendix_pgd}
\end{table}

Our empirical observation during the experiment is that \ourmodel-PGD frequently oscillates around a local minima of edit weights and fails to reach an optimal counterfactual. We hypothesize that the reason of lower flip rates from the signed method is that the edit weight search is constrained on nodes of a grid space (the grid unit length is step-size $\eta$), which loses precision and underperforms during counterfactual search.

% Appendix B

\begin{figure*}[h]
    \centering
    \begin{subfigure}[b]{0.48\linewidth}
        \label{fig:appendix_A_visual_an_Angry_face}
        \centering
        \includegraphics[width=\linewidth]{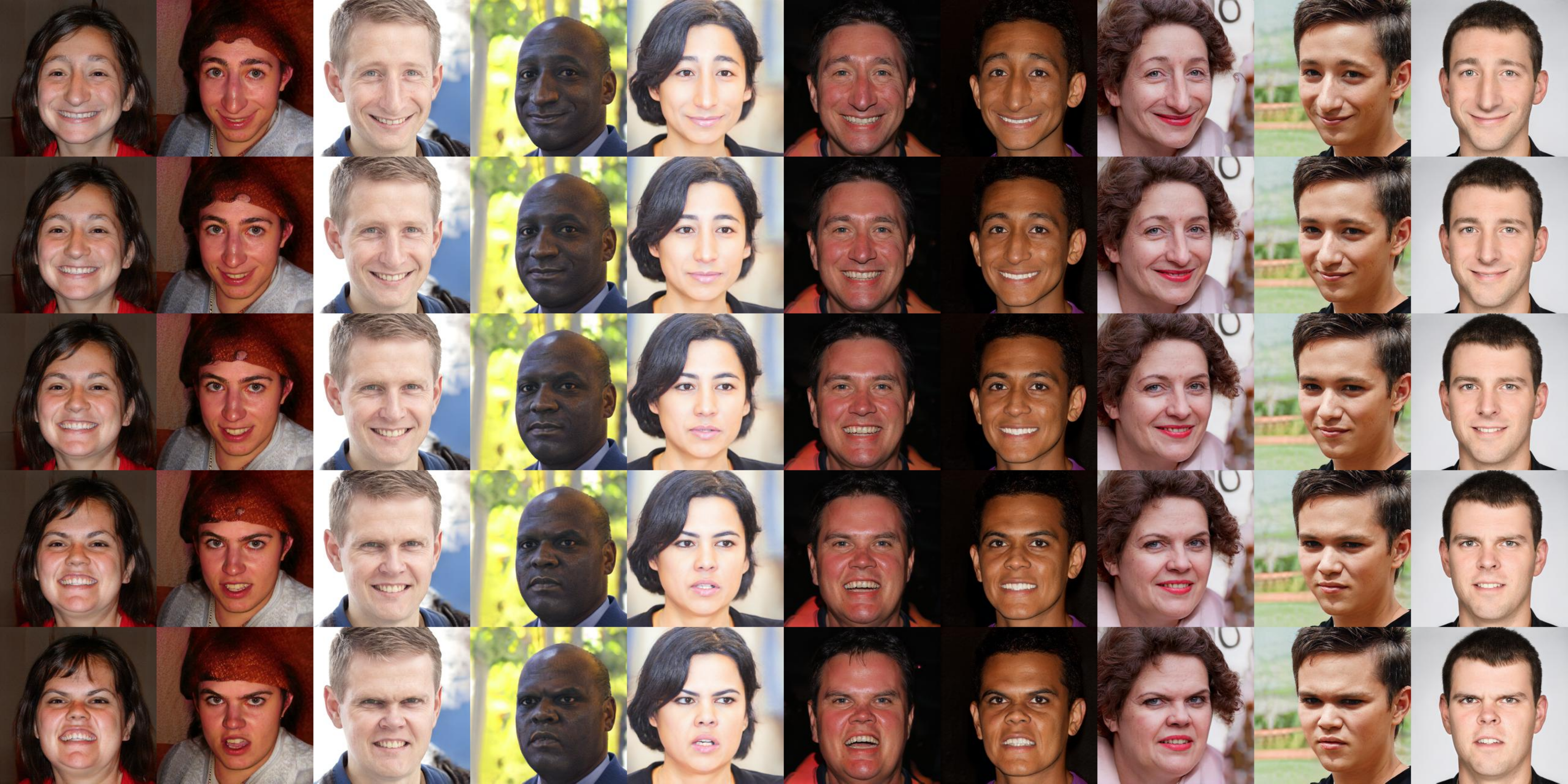}
        \caption{Attribute editing: an angry face.}
    \end{subfigure}
    \begin{subfigure}[b]{0.48\linewidth}
        \label{fig:appendix_A_visual_a_face_with_Eyeglasses}
        \centering
        \includegraphics[width=\linewidth]{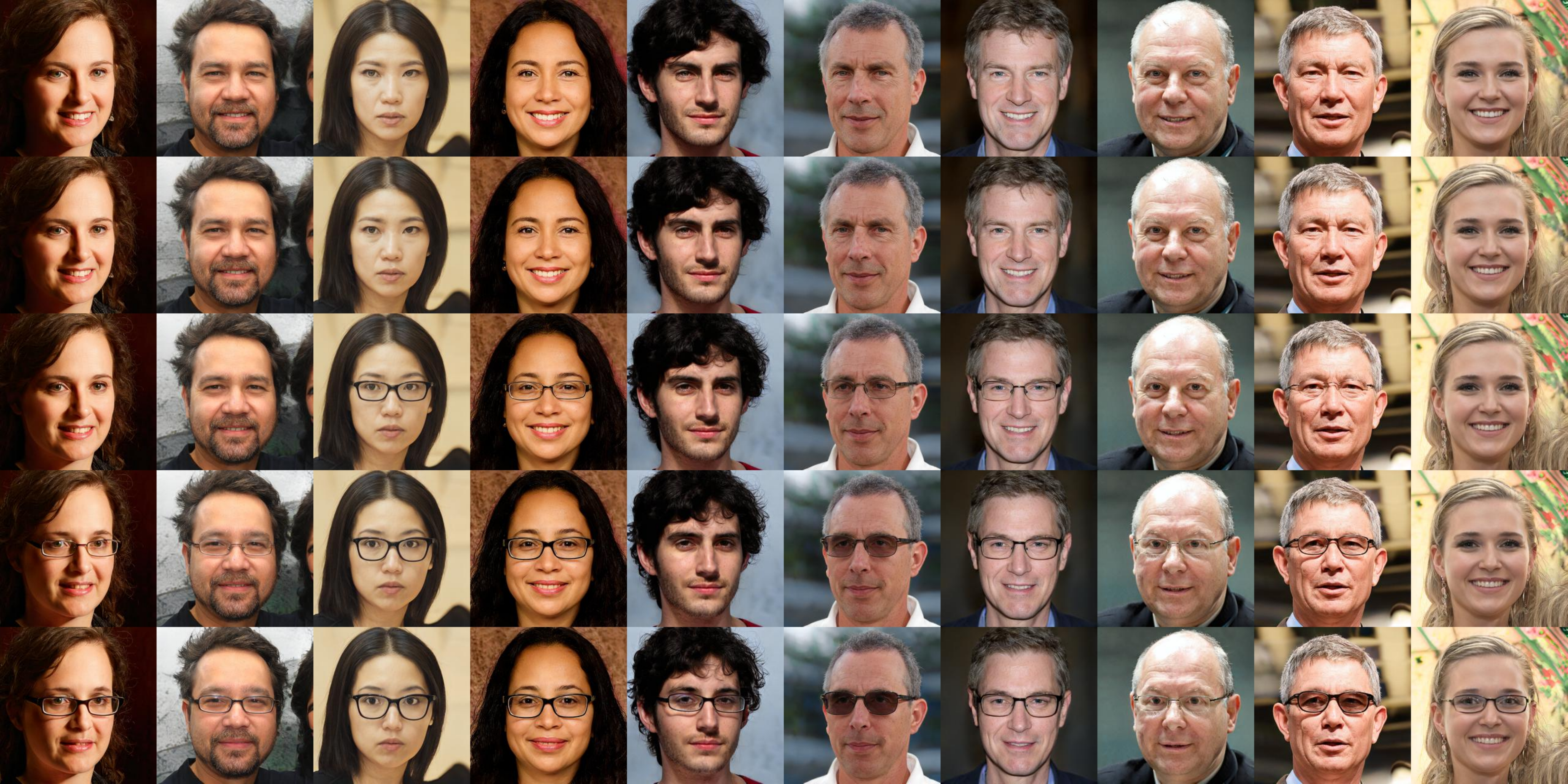}
        \caption{Attribute editing: a face with eyeglasses.}
    \end{subfigure}
    \begin{subfigure}[b]{0.48\linewidth}
        \label{fig:appendix_A_visual_a_Cute_face}
        \centering
        \includegraphics[width=\linewidth]{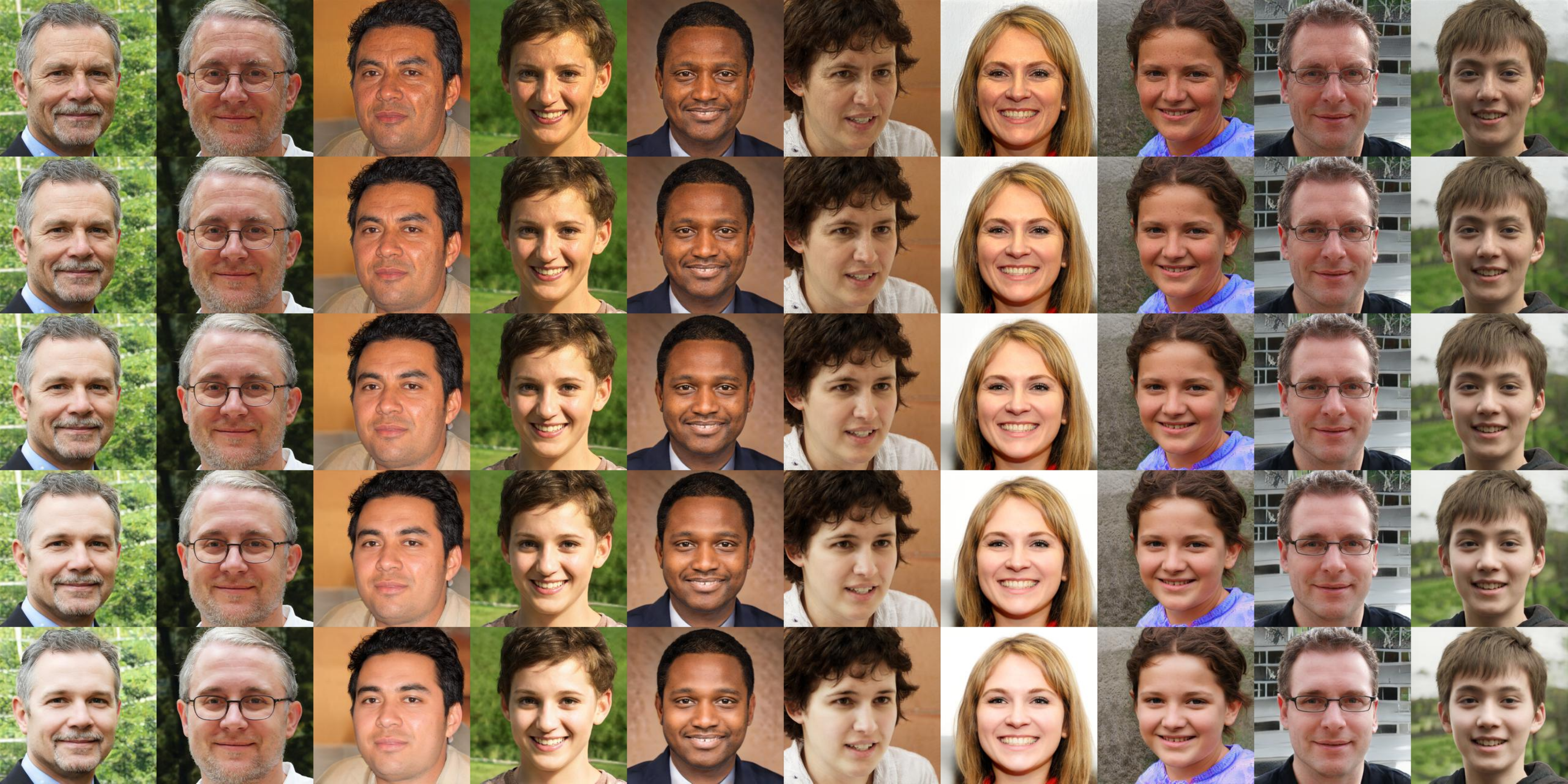}
        \caption{Attribute editing: a cute face.}
    \end{subfigure}
    \begin{subfigure}[b]{0.48\linewidth}
        \label{fig:appendix_A_visual_a_face_with_Blond_Hair}
        \centering
        \includegraphics[width=\linewidth]{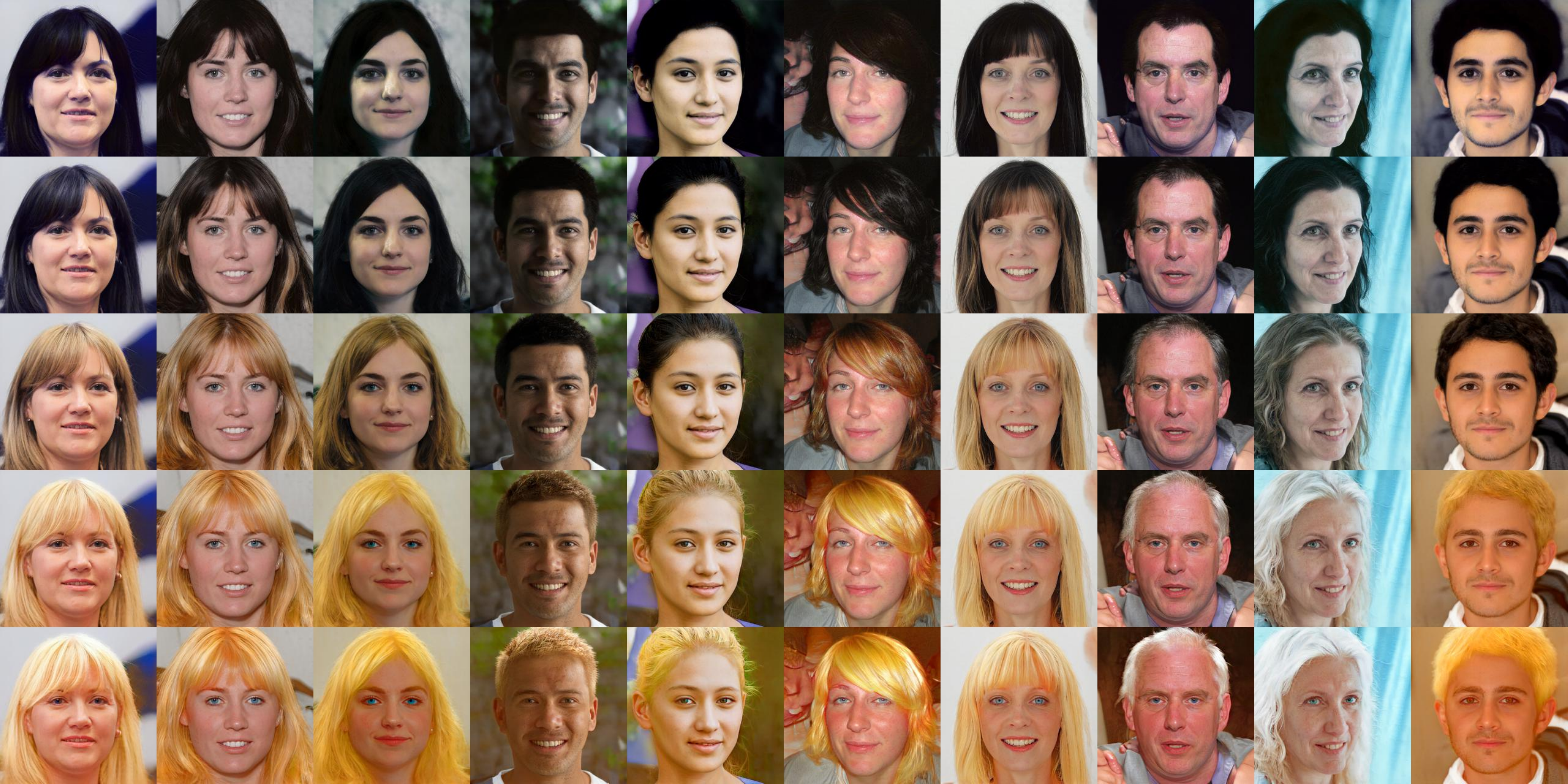}
        \caption{Attribute editing: a face with blond hair.}
    \end{subfigure}\\
    \begin{subfigure}[b]{0.48\linewidth}
        \label{fig:appendix_A_visual_a_face_with_Bangs}
        \centering
        \includegraphics[width=\linewidth]{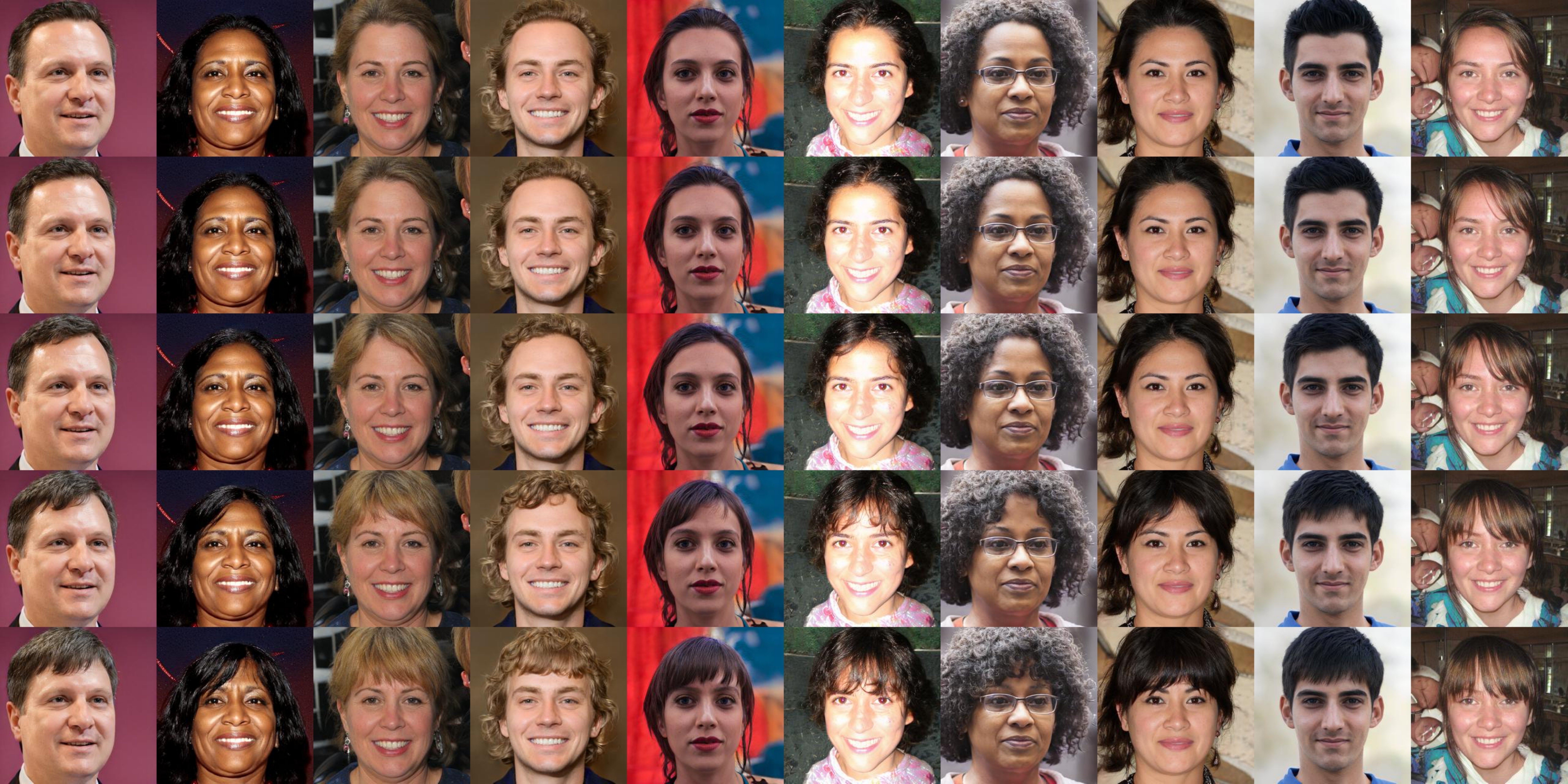}
        \caption{Attribute editing: a face with bangs.}
    \end{subfigure}
    \begin{subfigure}[b]{0.48\linewidth}
        \label{fig:appendix_A_visual_a_Smiling_face}
        \centering
        \includegraphics[width=\linewidth]{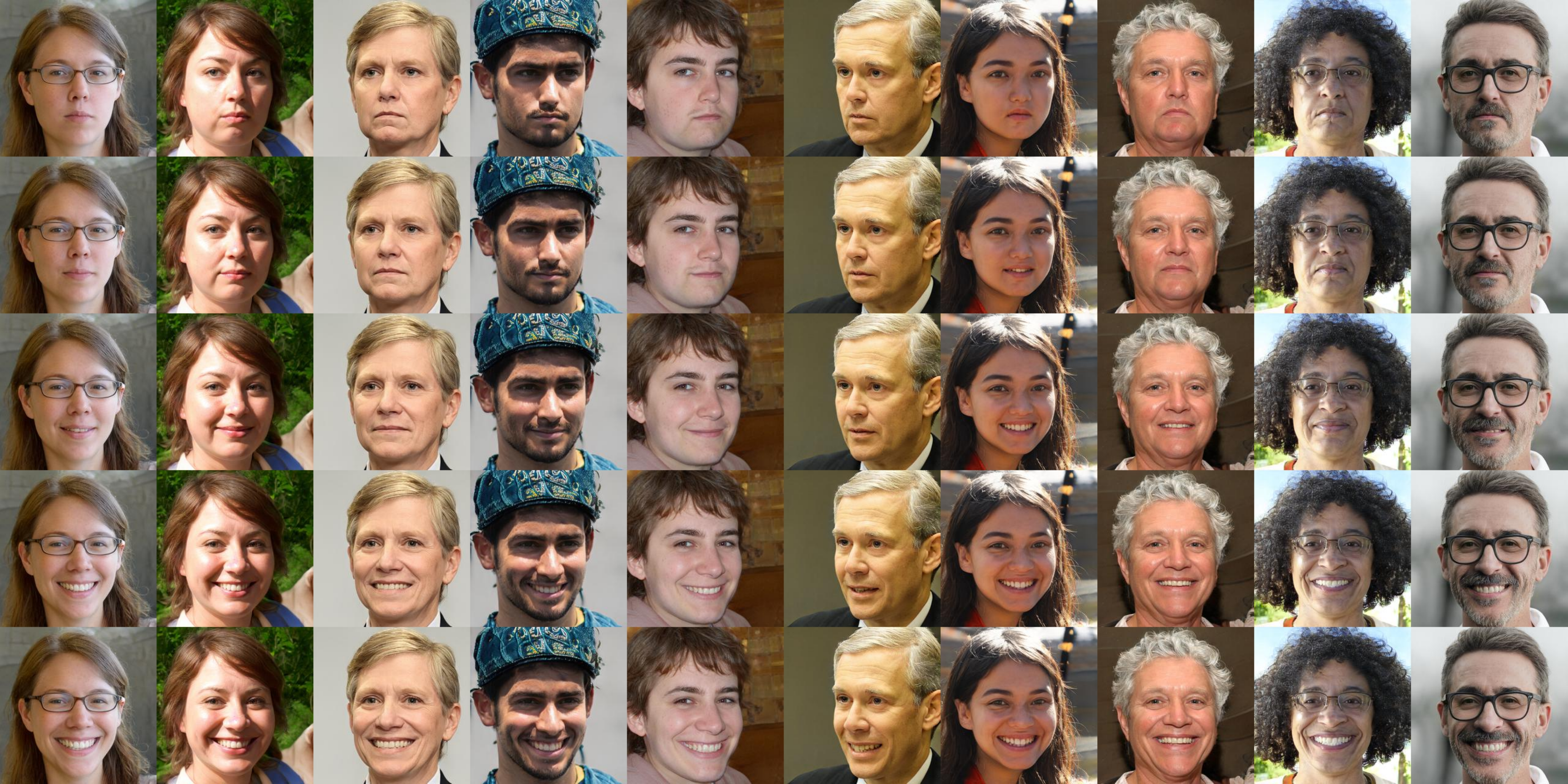}
        \caption{Attribute editing: a smiling face.}
    \end{subfigure}\\
    \begin{subfigure}[b]{0.48\linewidth}
        \label{fig:appendix_A_visual_a_Happy_face}
        \centering
        \includegraphics[width=\linewidth]{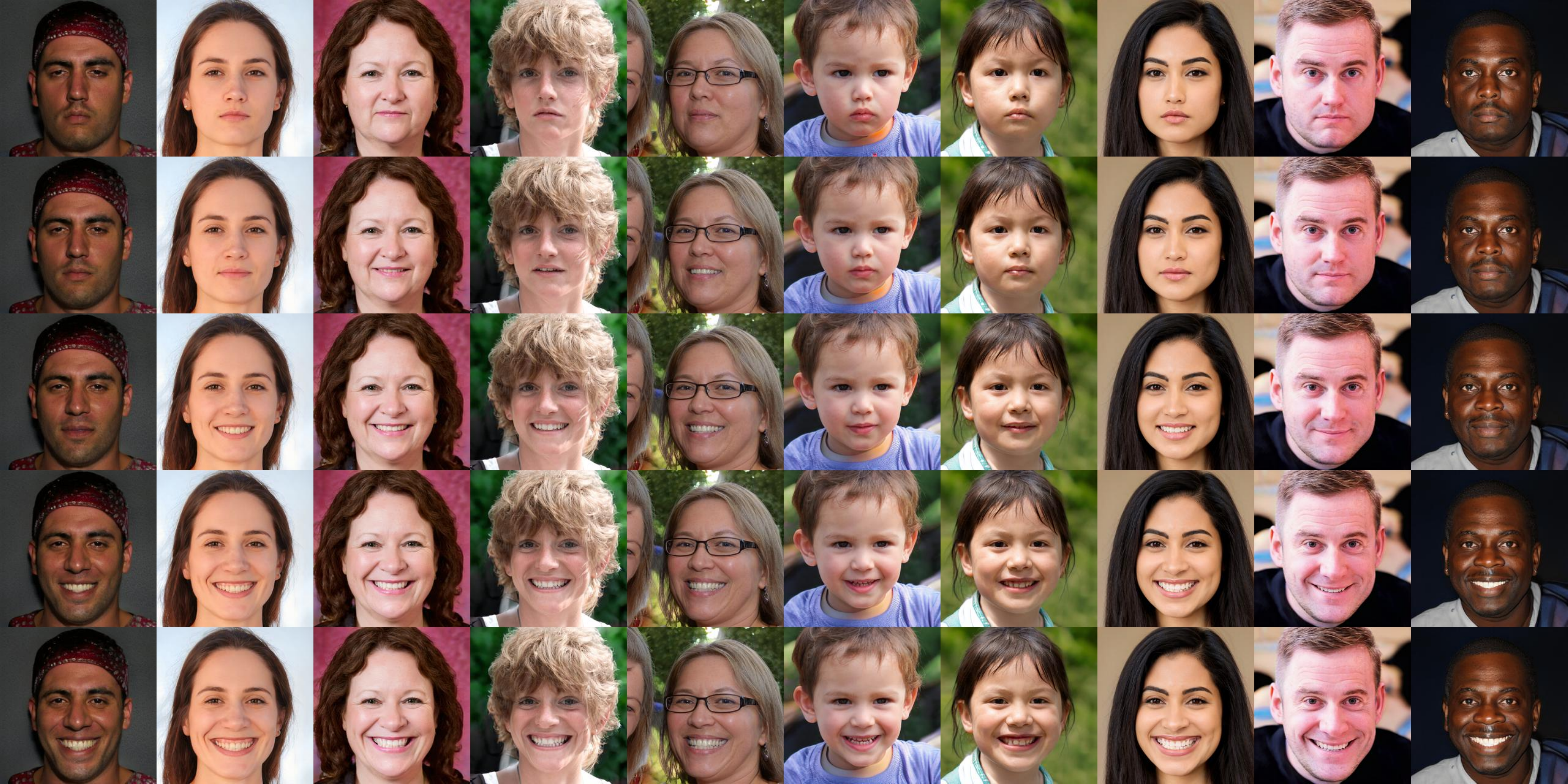}
        \caption{Attribute editing: a happy face.}
    \end{subfigure}
    \begin{subfigure}[b]{0.48\linewidth}
        \label{fig:appendix_A_visual_a_face_with_Curly_Hair}
        \centering
        \includegraphics[width=\linewidth]{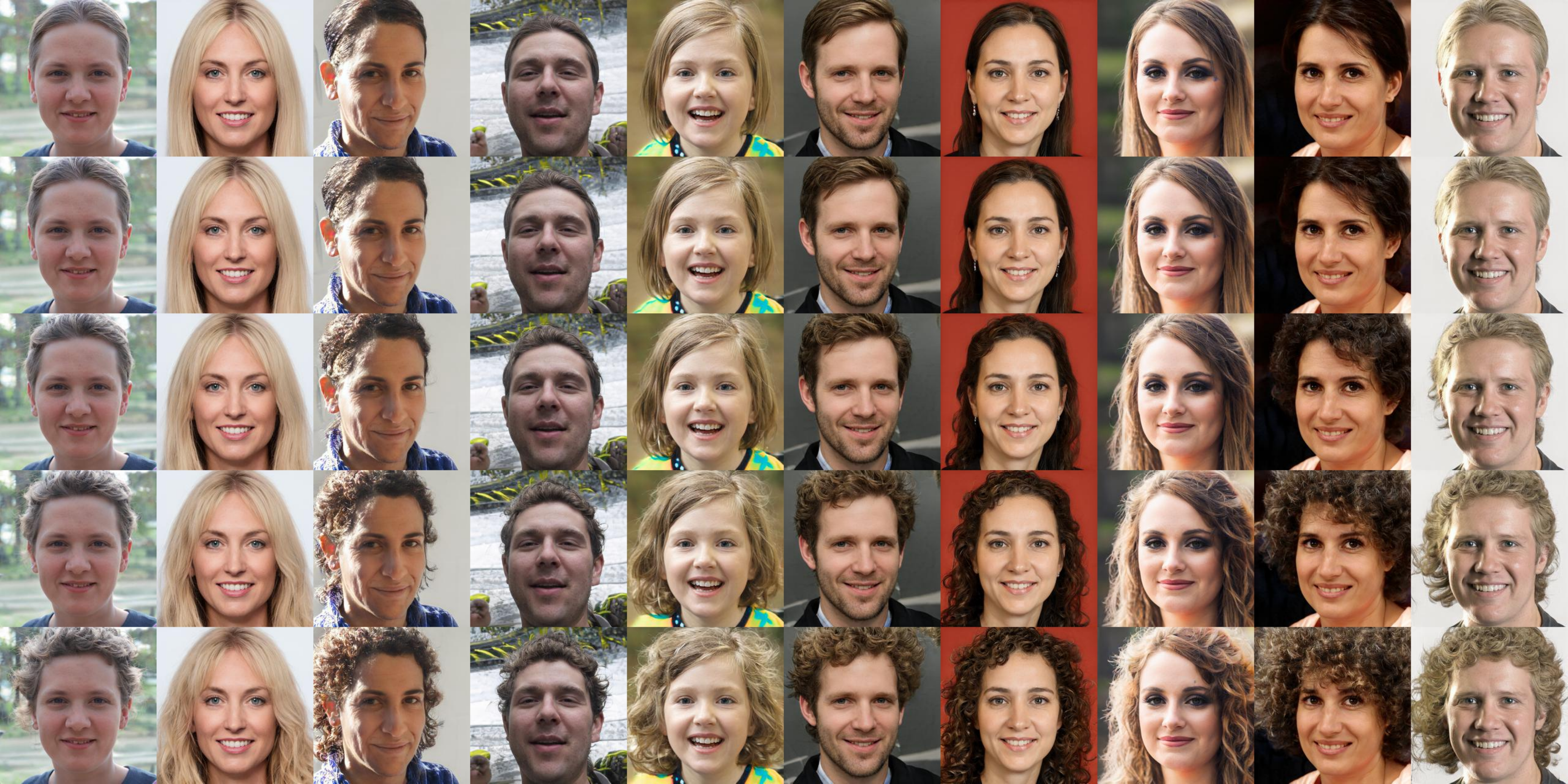}
        \caption{Attribute editing: a face with curly hair.}
    \end{subfigure}
    \end{figure*}

    \begin{figure*}[h]
    \centering
    \ContinuedFloat 
    \begin{subfigure}[b]{0.48\linewidth}
        \label{fig:appendix_A_visual_a_face_with_Beard}
        \centering
        \includegraphics[width=\linewidth]{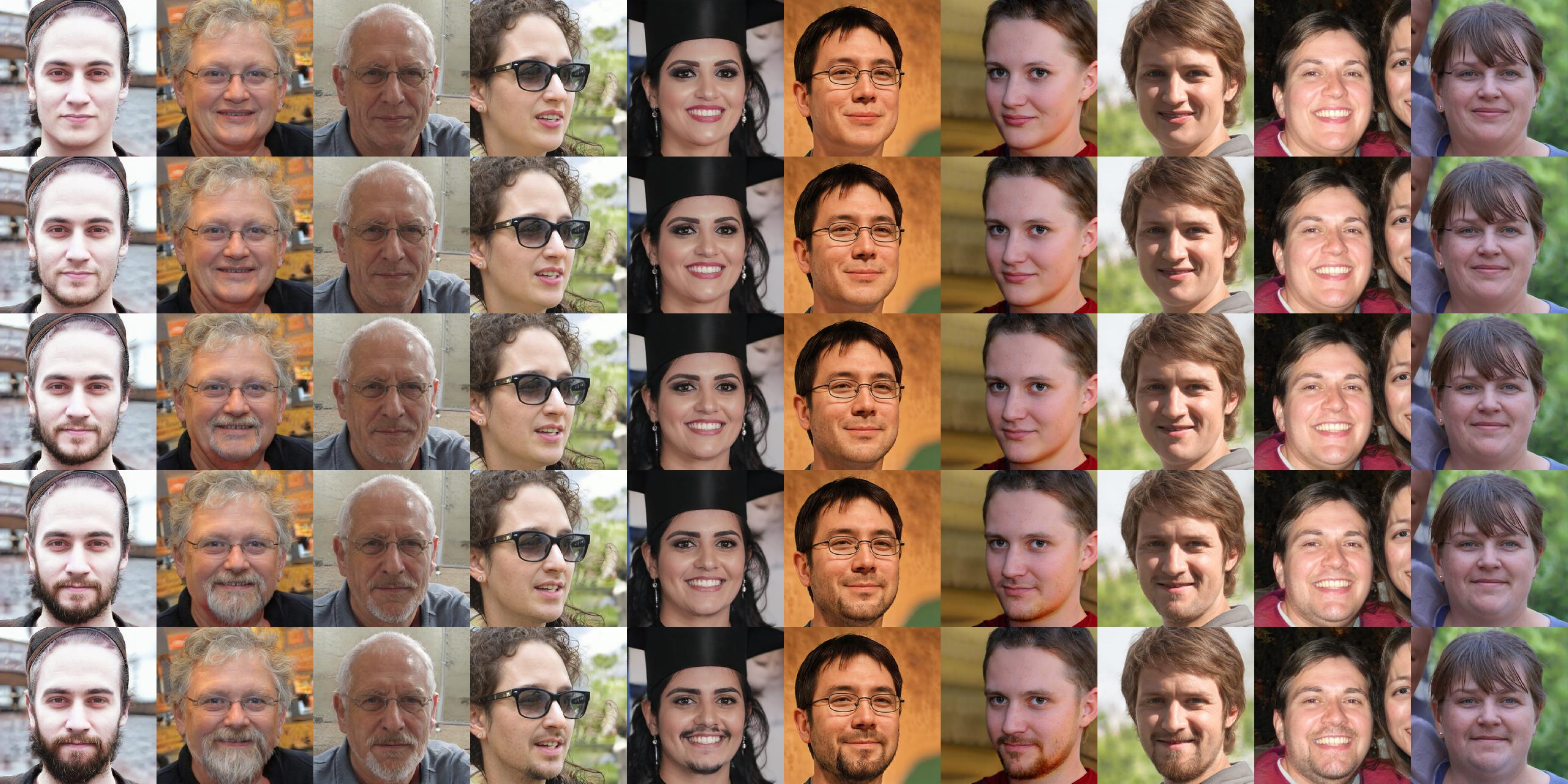}
        \caption{Attribute editing: a face with beard.}
    \end{subfigure}
    \begin{subfigure}[b]{0.48\linewidth}
        \label{fig:appendix_A_visual_a_face_with_Lipstick}
        \centering
        \includegraphics[width=\linewidth]{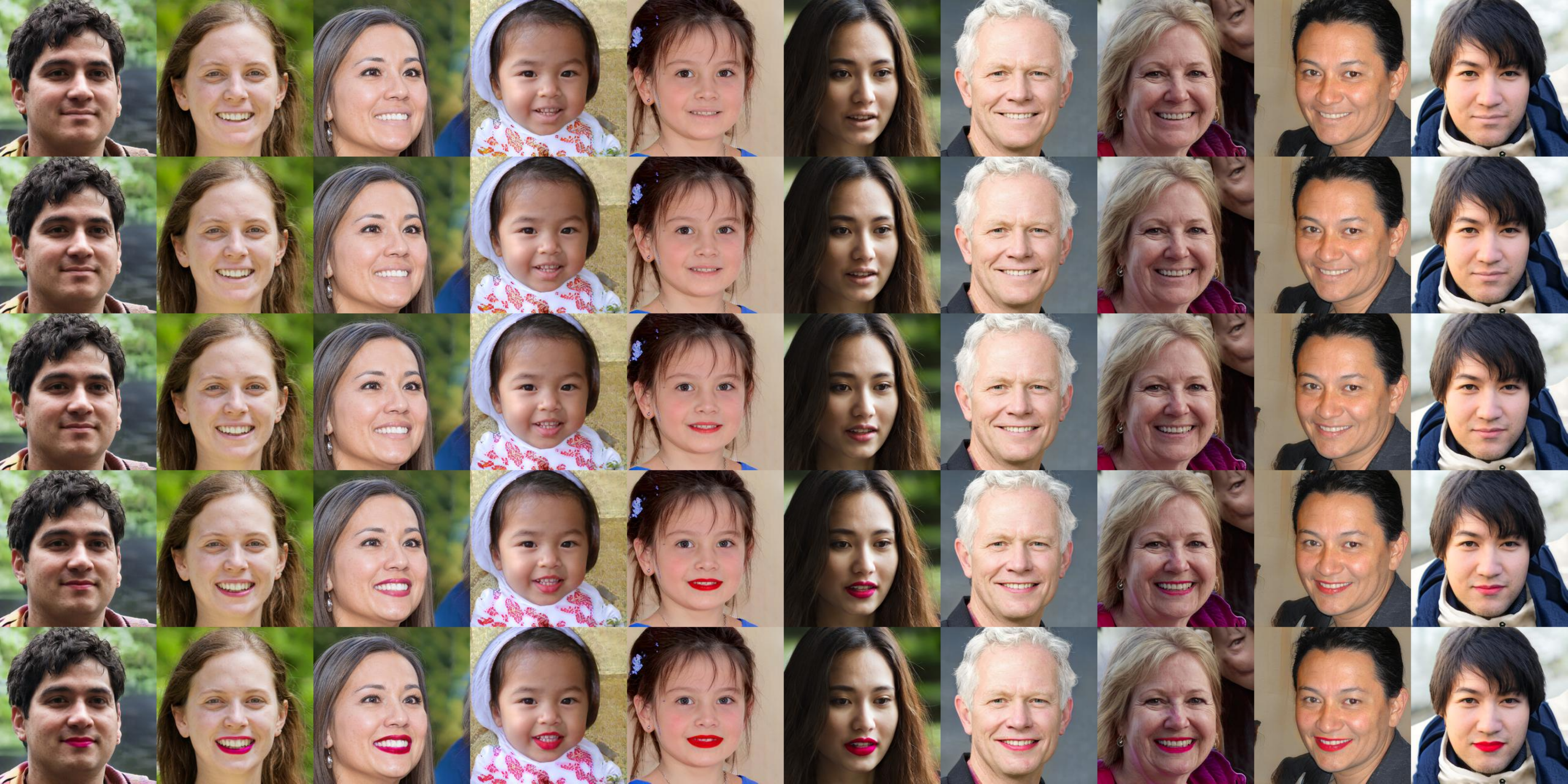}
        \caption{Attribute editing: a face with lipstick.}
    \end{subfigure}\\
    \begin{subfigure}[b]{0.48\linewidth}
        \label{fig:appendix_A_visual_a_Male_face}
        \centering
        \includegraphics[width=\linewidth]{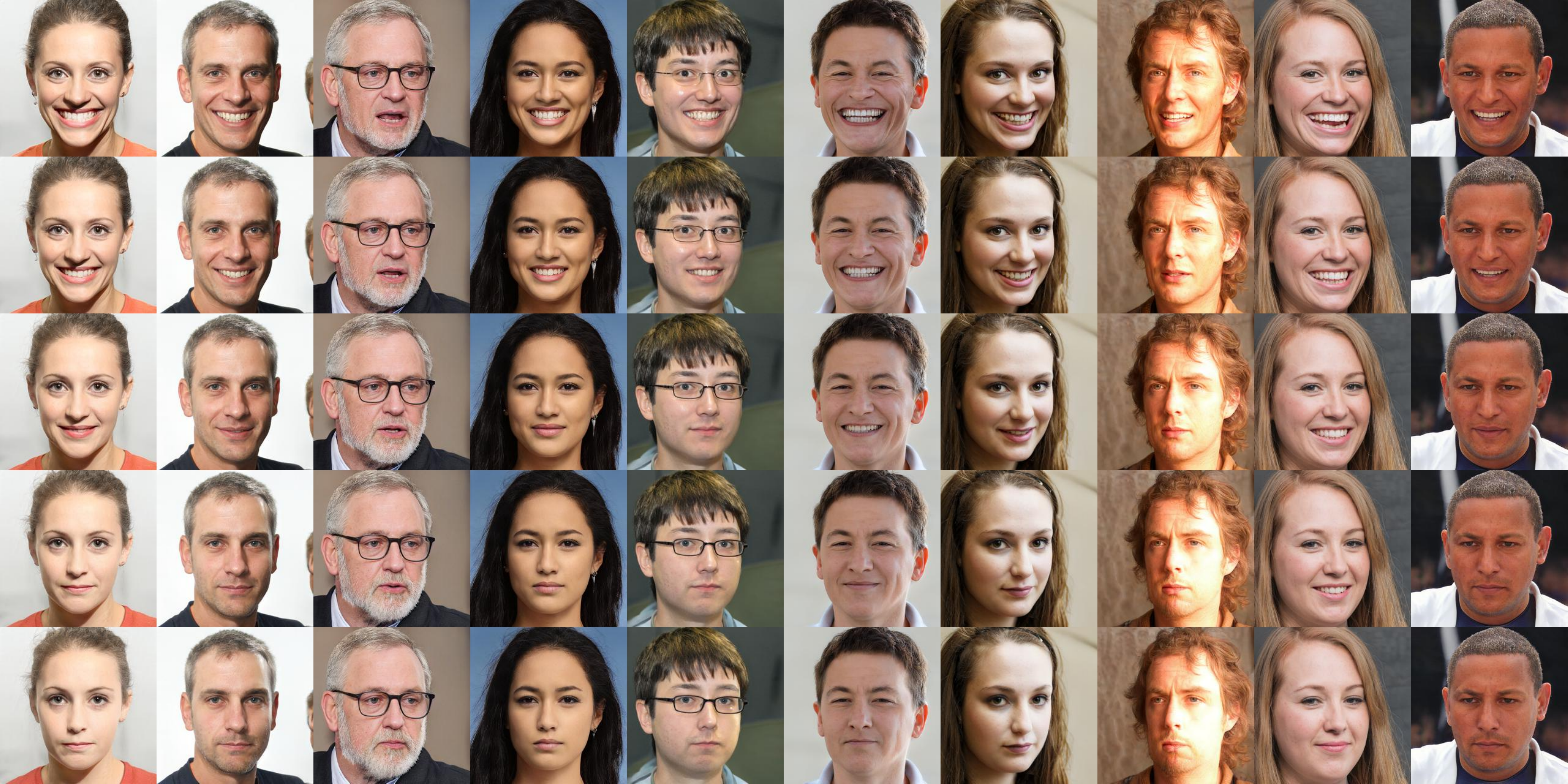}
        \caption{Attribute editing: a tired face.}
    \end{subfigure}
    \begin{subfigure}[b]{0.48\linewidth}
        \label{fig:appendix_A_visual_a_Surprised_face}
        \centering
        \includegraphics[width=\linewidth]{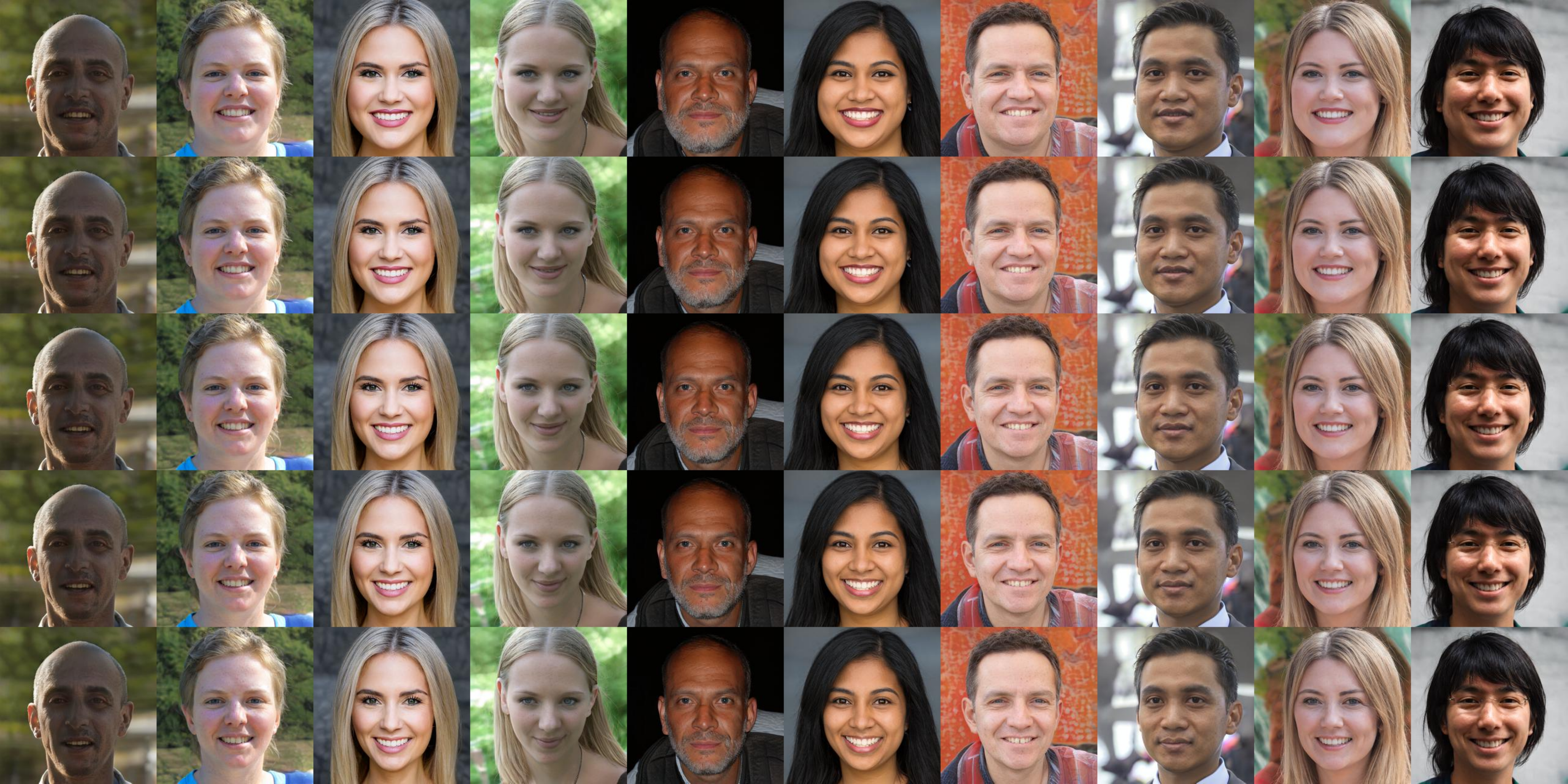}
        \caption{Attribute editing: a skinny face.}
    \end{subfigure}\\
    \begin{subfigure}[b]{0.48\linewidth}
        \label{fig:appendix_A_visual_a_Male_face}
        \centering
        \includegraphics[width=\linewidth]{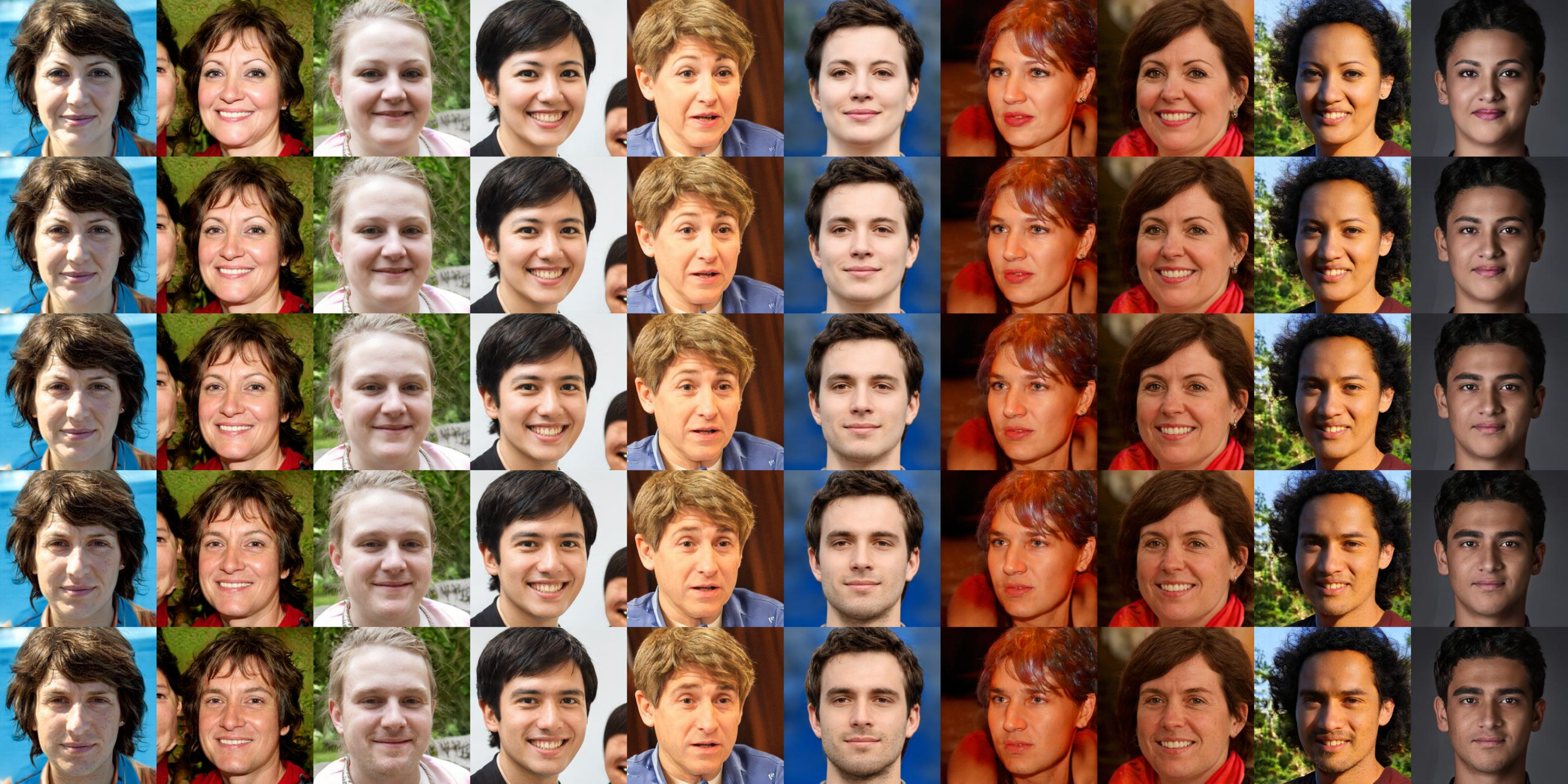}
        \caption{Attribute editing: a male face.}
    \end{subfigure}
    \begin{subfigure}[b]{0.48\linewidth}
        \label{fig:appendix_A_visual_a_Surprised_face}
        \centering
        \includegraphics[width=\linewidth]{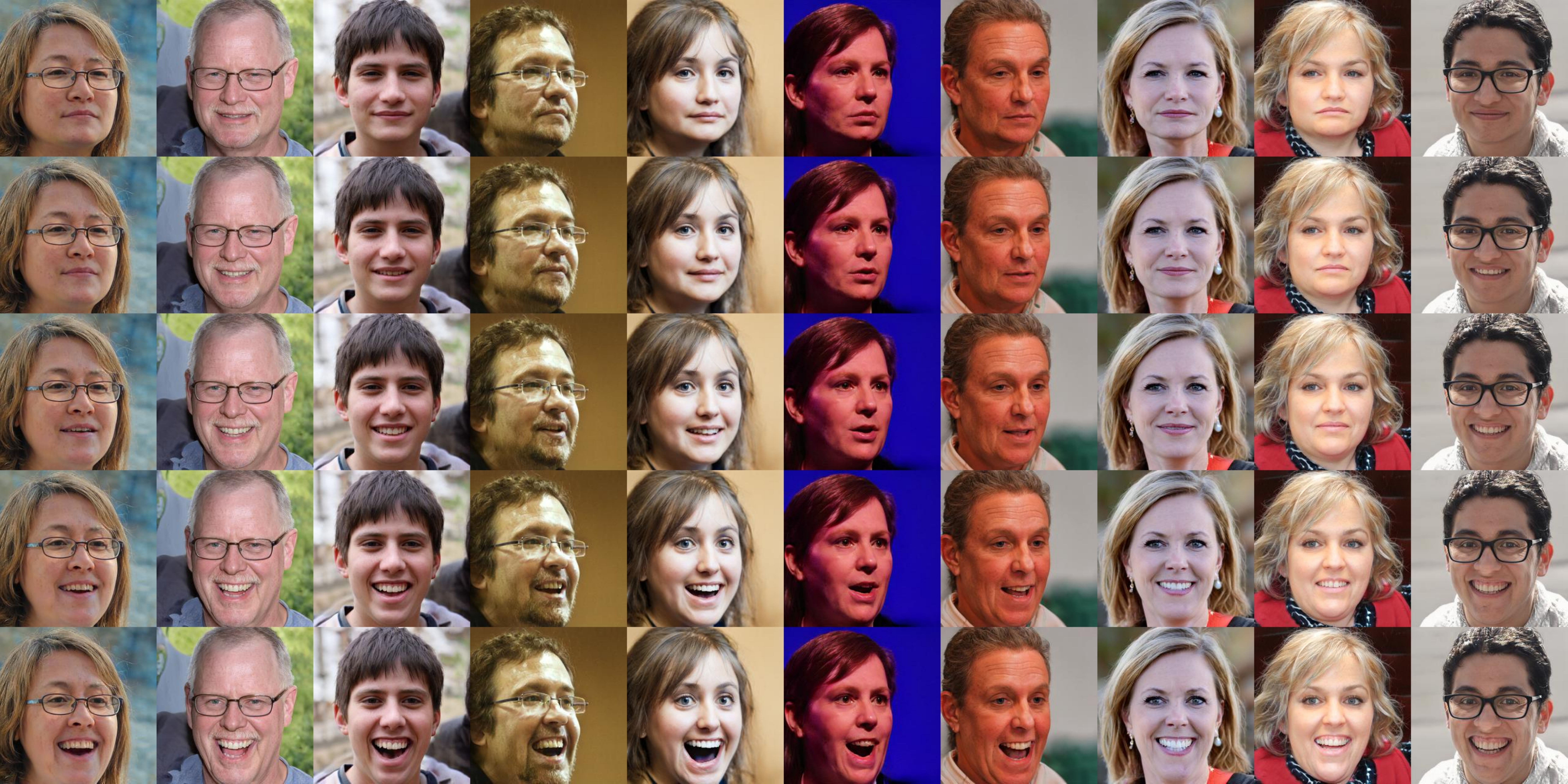}
        \caption{Attribute editing: a surprised face.}
    \end{subfigure}\\
    \begin{subfigure}[b]{0.48\linewidth}
        \label{fig:appendix_A_visual_a_face_with_Long_Hair}
        \centering
        \includegraphics[width=\linewidth]{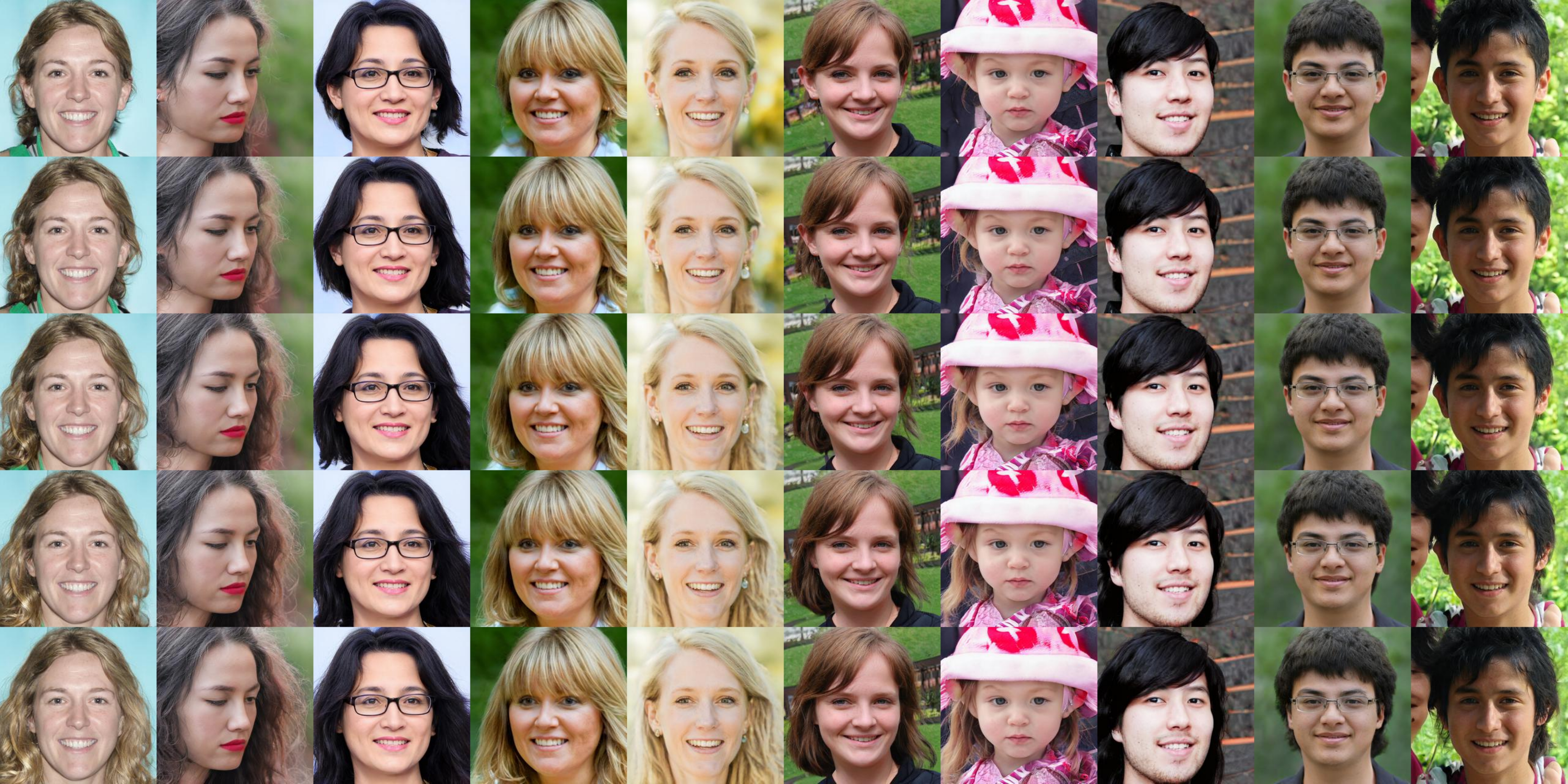}
        \caption{Attribute editing: a face with long hair.}
    \end{subfigure}
    \begin{subfigure}[b]{0.48\linewidth}
        \label{fig:appendix_A_visual_a_face_with_Pale_Skin}
        \centering
        \includegraphics[width=\linewidth]{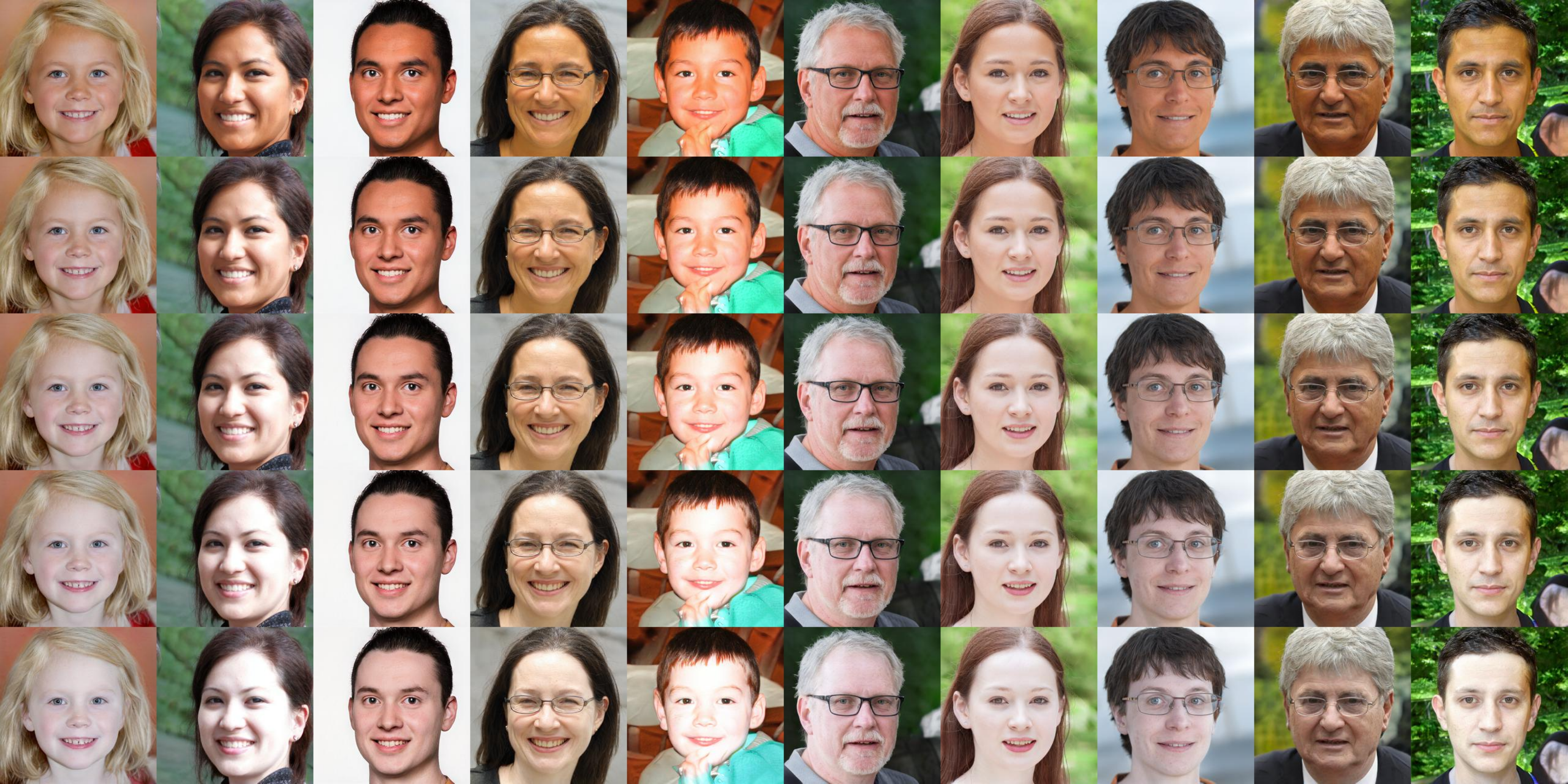}
        \caption{Attribute editing: a face with pale skin.}
    \end{subfigure}
    \caption{Visualization of global edit directions by utilizing the StyleCLIP channel relevance matrix. Images are sampled from the FFHQ domain using StyleGAN2-ADA. Every column demonstrates an edited image from edit weight $w = -30$ to $w = 30$. Weights of five images are linearly interpolated as $\{-30, -15, 0, 15, 30\}$. We can see that global edit directions are generalizable on multiple images.}
    \label{fig:appendix_attribute_editing_ffhq}
    
\end{figure*}

\end{document}